\documentclass{article}
\usepackage{arxiv}
\usepackage{hyperref}
\usepackage{xcolor}
\usepackage{graphicx}
\usepackage{longtable}
\usepackage{booktabs}
\usepackage{multirow}
\usepackage{fontawesome5}
\usepackage{floatpag}
\usepackage{subcaption}
\usepackage[multiple]{footmisc}
\usepackage{amsmath}
\usepackage{latexsym}
\usepackage{url}
\usepackage{caption}
\usepackage{microtype}

\definecolor{darkblue}{rgb}{0, 0, 0.5}
\hypersetup{colorlinks=true,citecolor=darkblue, linkcolor=darkblue, urlcolor=darkblue}

\usepackage{authblk}
\usepackage[titletoc,title]{appendix}

\usepackage[sorting=none]{biblatex} 
\usepackage{newfloat}
\DeclareFloatingEnvironment[name={Supplementary Figure},fileext=lsf,listname={List of Supplementary Figures}]{suppfigure}

\usepackage{microtype}

\makeatletter
\def\namedlabel#1#2{\begingroup
    #2%
    \def\@currentlabel{#2}%
    \phantomsection\label{#1}\endgroup
}
\makeatother

\title{Transformers and the representation of biomedical background knowledge}

\author[ ,1,2]{Oskar Wysocki\thanks{Corresponding author: oskar.wysocki@manchester.ac.uk}}
\author[1,2]{Zili Zhou}
\author[2]{Paul O'Regan}
\author[1]{Deborah Ferreira}
\author[2]{\\Magdalena Wysocka}
\author[2]{D\'onal Landers} 
\author[1,2,3]{Andr\'e Freitas}
\affil[1]{Department of Computer Science, The University of Manchester, Manchester, UK}
\affil[2]{digital Experimental Cancer Medicine Team, Cancer Biomarker Centre,\authorcr CRUK Manchester Institute, University of Manchester, Manchester, UK}
\affil[3]{Idiap Research Institute, Martigny, Switzerland}

\begin{document}

\date{}
\maketitle

\begin{abstract}

Specialised transformers-based models (such as BioBERT and BioMegatron) are adapted for the biomedical domain based on publicly available biomedical corpora. As such, they have the potential to encode large-scale biological knowledge. We investigate the encoding and representation of biological knowledge in these models, and its potential utility to support inference in cancer precision medicine - namely, the interpretation of the clinical significance of genomic alterations. We compare the performance of different transformer baselines; we use probing to determine the consistency of encodings for distinct entities; and we use clustering methods to compare and contrast the internal properties of the embeddings for genes, variants, drugs and diseases. We show that these models do indeed encode biological knowledge, although some of this is lost in fine-tuning for specific tasks. Finally, we analyse how the models behave with regard to biases and imbalances in the dataset. 
\end{abstract}

\section{Introduction}



Transformers are deep learning models which are able to capture linguistic patterns at scale. By using unsupervised learning tasks which can be defined over large-scale textual corpora, these models are able to capture both linguistic and domain knowledge, which can be later specialised for specific inference tasks. The representation produced by the model is a high-dimensional linguistic space which represents words, terms and sentences as vector projections. In Natural Language Processing, transformers are used to support natural language inference and classification tasks. The assumption is that the models can encode syntactic, semantic, commonsense and domain-specific knowledge and use their internal representation for complex textual interpretation. While these models provided measurable improvements in many different tasks, the limited interpretability of their internal representation challenges their application in areas such as biomedicine. 



In this work we elucidate a set of the internal properties of transformers in the context of a well-defined cancer precision medicine inference task,in which the domain knowledge is expressed within the biomedical literature.  We focus on systematically determining the ability of these models to capture fundamental entities (gene, gene variant, drug and disease), their relations and supporting facts, which are fundamental for supporting inference in the context of molecular cancer medicine. For example, we aim to answer the question whether these models capture biological knowledge such as the following: 

\begin{itemize}
\item \textit{``T790M is a gene variant"}
\item \textit{``T790M is a variant of the EGFR gene"}
\item \textit{``The T790M variant of the EGFR gene in lung cancer is associated with resistance to Erlotinib"} - well supported statement (Level A - Validated association, Confidence rating: 5 stars).
\item \textit{``The T790M variant of the EGFR gene in pancreatic cancer is associated with resistance to Osimertinib"} - less supported statement (Level C - Case study, Confidence rating: 2 stars).
\end{itemize}

In the example above, the first two facts capture basic definitional knowledge (mapped respectively to an unary and binary predicate-argument relation), while the third and fourth facts capture a full scientific statement which can be mapped to a complex n-ary relation, which are supported by different levels of evidence in the literature. The establishment of the truth condition of facts of these types in the context of a biomedical natural language inference task, is a desirable property for these models. With this motivation in mind, this work provides a critical exploration of the internal representation properties of these models, using probing and clustering methods. In summary, we aim to answer following research questions (RQs): 

\begin{enumerate}
\item[\namedlabel{itm:RQ1}{RQ1}] Do transformer-based models encode fundamental biomedical domain knowledge at entity level (e.g. gene, gene variant, disease drug) and at a relational level?
\item[\namedlabel{itm:RQ2}{RQ2}] Do these models encode complex  biomedical facts/n-ary relations?
\item[\namedlabel{itm:RQ3}{RQ3}] Are there significant differences on how different model configurations encode domain knowledge?
\item[\namedlabel{itm:RQ4}{RQ4}] How these models cope with  evidence biases in the literature (e.g. are facts more frequently expressed in the literature, elicited in the models)?

\end{enumerate}

In this analysis, we used state-of-the-art transformers  specialised for the biomedical domain: BioBERT \cite{lee2020biobert} and BioMegatron \cite{Shin2020BioMegatronLB}. Both models are pre-trained over large biomedical text corpora (PubMed\footnote{www.ncbi.nlm.nih.gov/pubmed}). These models have demonstrated  on an extrinsic setting to address complex domain-specific tasks   \cite{wangPretrainedLanguageModels2021}, such as answering biomedical questions \cite{Shin2020BioMegatronLB}. Yet, the internal representation properties of these models are not fully  characterised, a requirement for their safe and controlled application in biomedical setting.  

This paper focuses on the following contributions:

\begin{itemize}

\item  A systematic evaluation of the ability of biomedical fine-tuned transformers (BioBERT and BioMegatron) to capture entities, complex relations and level of evidence support for biomedical facts within a specific domain of inference (cancer clinical trials). Instead of focusing only on extrinsic performance (in the context of a classification task), we elicit some of the internal properties of these models with the support of clustering and probing methods.

\item  To the best of our knowledge, this is the first work which systematically links the evidence from high quality, expert-curated knowledge base with the representation of biomedical knowledge in transformers, i.e. n-ary relations and entity types.

\item We used probing methods to inspect the consistency of entities and associated types (i.e. genes, variants, drugs, diseases) contrasting pre-trained and fine-tuned models. This allowed for the evaluation whether the model captures the fundamental biomedical/semantic categories to support interpretation. We quantified how much semantic structure is lost in fine-tuning. 

\item  To the best of our knowledge, this is the first work that quantifies the relation of classification error to entities distribution in the dataset and evidence items in literature, emphasising the risk of and demonstrating examples of significant errors in the cancer precision medicine inference task. We showed, that despite the soundness and strength of the evidence in the biomedical literature, some well-known clinical relations can be misclassified. 

\item Lastly, we provided a qualitative analysis of the significant clustering patterns of the embeddings, using dimensionality reduction and unsupervised clustering methods to identify qualitative patterns expressed in the representations. This approach allowed for identification of biologically meaningful representations, e.g. groups with genes from the same pathways. Additionally,  by measuring homogeneity of clusters, we quantified the associations between the representations and the entity type and target labels. 	

\end{itemize}

\begin{figure}[htb!]
\centering
\begin{subfigure}{.55\textwidth}
  \centering
  \includegraphics[width= .99\textwidth]{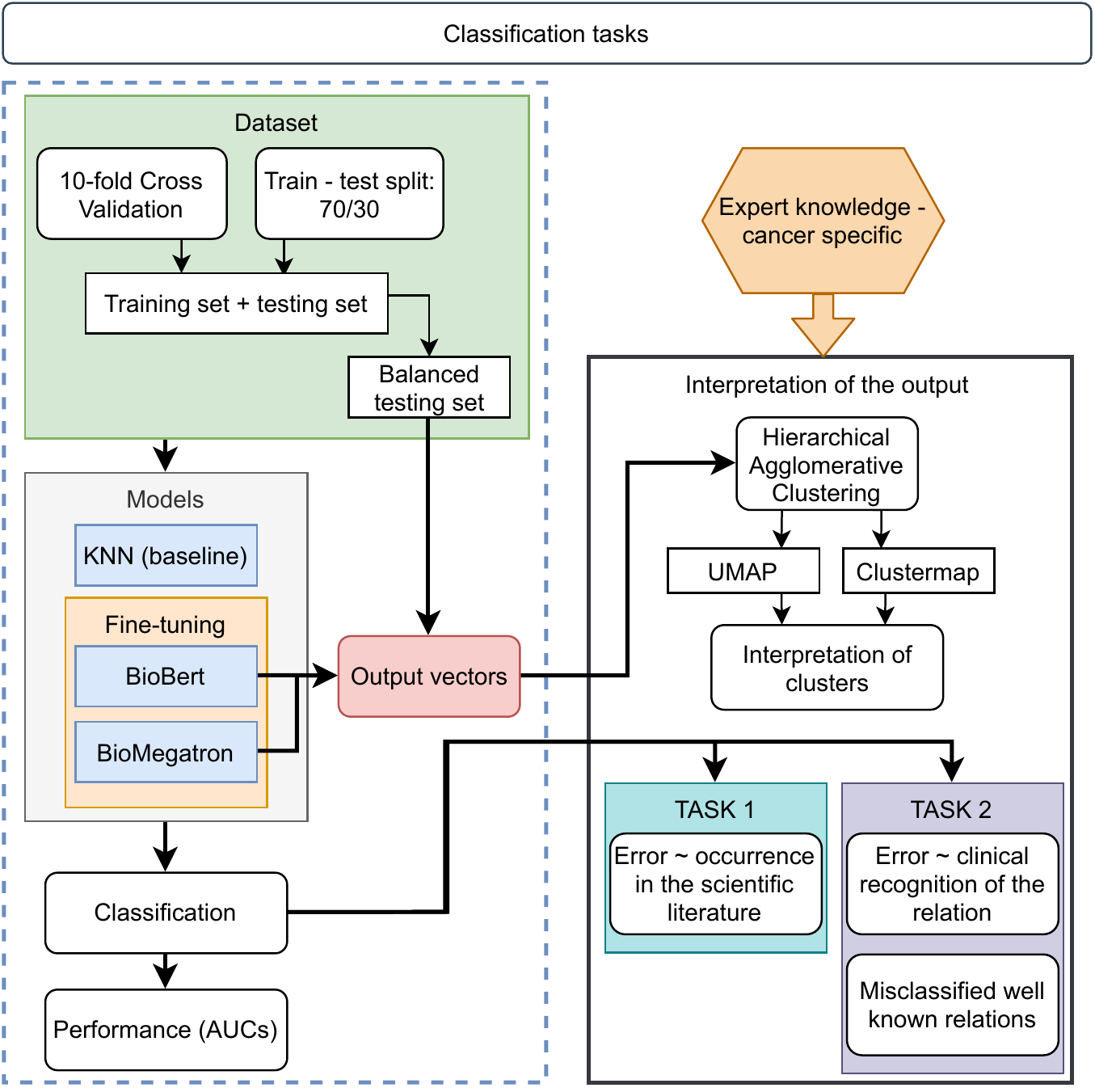}
\caption{}
\label{fig:workflow_tasks}
\end{subfigure}%
\begin{subfigure}{.45\textwidth}
  \centering
 \includegraphics[width= .99\textwidth]{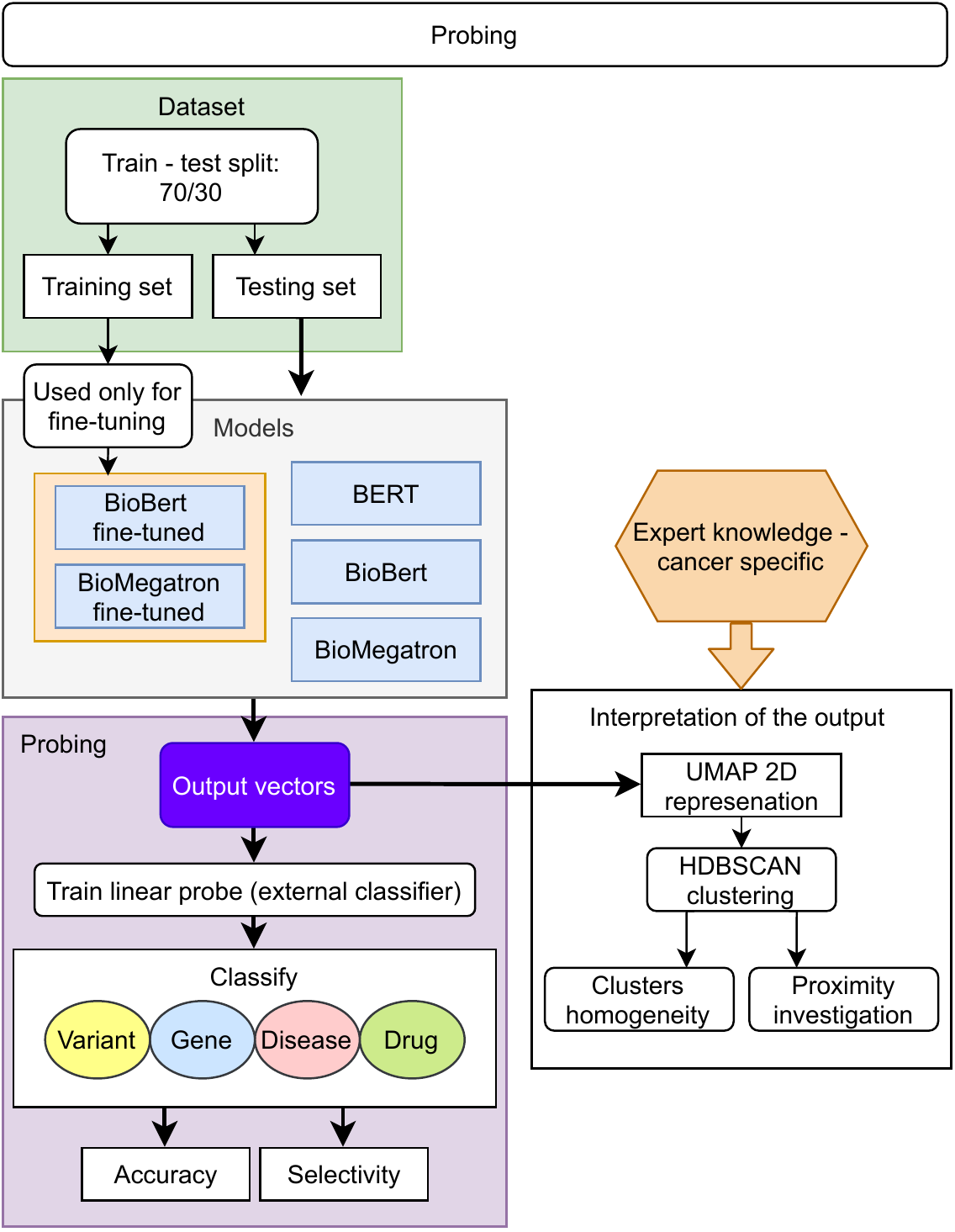}
\caption{}
\label{fig:workflow_probing}
\end{subfigure}

\caption{The workflow of the performed analysis.}
\label{fig:workflow}
\end{figure}

The workflow of the analysis is summarized in Fig.\ref{fig:workflow}.

\begin{figure}[htb!]
\centering
\includegraphics[width= 0.8\textwidth]{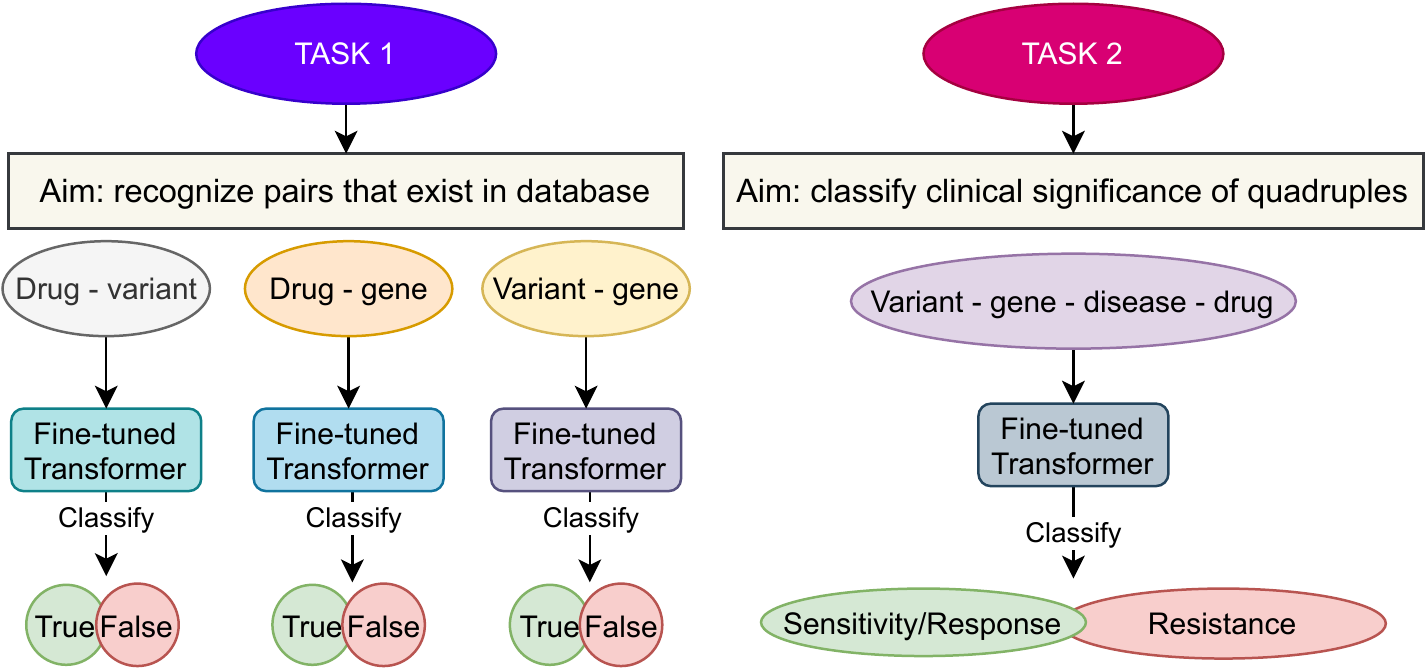}
\caption{An overview of classification task 1 and 2. Each Transformer block represents a separate model which was fine-tuned separately for each classification. Two Transformers were used: BioBERT and BioMegatron.}
\label{fig:tasks_overview}
\end{figure}

\section{Methods}

\subsection{Motivational scenario: natural language inference in cancer clinical research}

Cancer precision medicine, which is the selection of a treatment for a patient based on molecular characterisation of their tumour, has the potential to improve patient outcomes. For example, activating mutations in the epidermal growth factor receptor gene (EGFR) predict response to gefitinib, and amplification or overexpression of ERBB2 predicts response to anti-ERBB2 therapies such as lapatinib. Tests for these markers that guide therapy decisions are now part of the standard of care in non-small-cell lung cancer (NSCLC) and breast cancer  \cite{Good2014}. 

Routine molecular characterisation of patients’ tumours has become feasible because of improved turnaround times and reduced costs of molecular diagnostics \cite{doi:10.1200/PO.18.00098}. In England, the NHS England genomic medicine service aims to offer whole genome sequencing as part of routine care. The aim is to match people to the most effective interventions, in order to increase survival and reduce the likelihood of adverse drug reactions \footnote{https://www.england.nhs.uk/genomics/nhs-genomic-med-service/}.

Even considering only licensed treatments, the number of alternative treatments available may be very large. For example, in the USA, there are over 70 drugs approved by the FDA for the treatment of NSCLC \footnote{\url{https://www.cancer.gov/about-cancer/treatment/drugs/lung}}. If experimental treatments are included in the decision making process, the number of alternative treatments available is substantially increased. 

Furthermore, as the breadth of molecular testing increases, so too does the volume of information available for each patient and thus the complexity of the treatment decision. Interpretation of the clinical and functional significance of the resulting data presents a substantial and growing challenge to the implementation of precision medicine in the clinical setting. 

This creates a need for tools to support clinicians in the evaluation of the clinical significance of genomic alterations in order to be able to implement precision medicine. However, much of the information available to support clinicians in making treatment decisions is in the form of unstructured text, such as published literature, conference proceedings and drug prescribing information. Natural language processing methods has the potential to scale-up the interpretation of this evidence space which could be integrated into decision support tools.  The utility of a decision support tool is expressed in providing support for individual recommendations. Despite acknowledging inherent imperfectness of the model's overall performance, the trustworthiness and safety of such tool would require the correct interpretation of biological facts and emerging evidence. This work validates an approach of applying fine-tuned transformers to two simple NLI tasks, investigating encoded knowledge within the models together with aforementioned individual well-established clinical relations. This work contributes for the first time with two concrete cancer precision medicine inference tasks based on a high quality, manually-curated dataset. For general evaluation of transformers in biomedical applications please refer to \cite{wangPretrainedLanguageModels2021,alghanmiProbingPreTrainedLanguage2021,jinProbingBiomedicalEmbeddings2019}, where the models are tested in multiple downstream tasks.

\subsection{ Reference Clinical Knwoledge Base (KB)}

CIViC\footnote{\url{https://civicdb.org/home}} (Clinical Interpretation of Variants in Cancer) is a community-edited knowledge base of associations between genetic variations (or other alterations), drugs and outcomes in cancer \cite{96a1c27041c0460091a2ed7a0b07b421}. The goal of CIViC is to support the implementation of personalized medicine in cancer. Data is freely available and licensed under a Creative Commons Public Domain Dedication (CC0 1.0 Universal). The knowledge base includes a detailed curation of evidence obtained from peer-reviewed publications and meeting abstracts. The CIViC database supports the development of computational tools for the functional prediction and interpretation of the clinical significance of cancer variants.  Together with OncoKB \cite{doi:10.1200/PO.17.00011} and My Cancer Genome\footnote{\url{https://www.mycancergenome.org/}}
, it is one of the most commonly used knowledge bases for this purpose \cite{10.1093/bib/bbab134}.

An evidence statement is a brief description of the clinical relevance of a variant that has been determined by an experiment, trial, or study from a published literature source. It captures a variant’s impact on clinical action, which can be predictive of therapy, correlated with prognostic outcome, inform disease diagnosis (i.e. cancer type or subtype), predict predisposition to cancer in the first place, or relate to the functional impact of the variant. For each item of evidence, additional attributes are captured, including:

\begin{itemize}  
\item \textit{Type} - the type of clinical (or biological) association described (e.g. Predictive, Prognostic, Functional etc.).
\item \textit{Direction} - whether the evidence supports or refutes the clinical significance of an event.
\item \textit{Level} - a measure of the robustness of the associated study, where \textit{A - Validated association} is the strongest evidence, and \textit{E - Inferential association} is the weakest evidence.  
\item \textit{Rating} - a score (1-5 stars) reflecting the database curator’s confidence in the quality of the summarized evidence. 
\item \textit{Clinical Significance} - describes how the variant is related to a specific, clinically-relevant property (e.g. drug sensitivity or resistance). 
\end{itemize}

CIViC is programmatically accessible via API and  as a full dataset and is integrated into various recent annotation tools and it follows an ontology driven conceptual model. It allows users to transparently generate current and accurate variant interpretations because it receives monthly updates. As of 5 January 2022, the database holds 8,441 interpretations of clinical relevance for 2,969 variants among 460 genes associated with 322 diseases and 479 drugs. Its accessibility and tabular format of the data allows for easy integration into  Machine Learning (ML) pipelines, both as input data and domain knowledge incorporated in the model.

\subsection{Data preprocessing \& Set-up}
The process of  pre-processing the CIViC data for the purpose of this study is detailed in Supp Methods \ref{sec:download_data}.

As we were interested in identifying gene variants that predict response to one or more drugs, we retained only those evidence items where \textit{Evidence Direction} contains the value \textit{Supports} and \textit{Evidence type} has the value \textit{Predictive}. 
  
  
\subsubsection{Task 1 - generation of true/false entity pairs}
The first classification task (Fig.\ref{fig:tasks_overview}) was to determine whether a transformer model, pre-trained on the existing biomedical corpus and fine-tuned for the task could correctly classify associations between pairs of entities \textit{entity1-entity2} as true or false based on knowledge embedded from the biomedical corpus. For example,  the correct classification of T790M as a variant of the EGFR gene, but not of the KRAS gene. 

Three types of  binary relations were considered:
\begin{itemize}
\item drug - gene
\item drug - variant
\item variant - gene
\end{itemize}

Pairs of entities with genuine associations (‘true pairs’) were generated from the CIViC knowledge base; pairs of entities with no such association (‘false pairs’) were generated by randomly selecting entities from CIViC, and excluding those that already exist (i.e., negative sampling). The dataset includes equal  number of false and true pairs. Of note, a pair can occur in multiple evidence items, i.e. be duplicated in the database, but our datasets of pairs consisted of unique pairs.

\subsubsection{Task 2 - generation of variant-gene-disease-drug quadruples}
 The second classification task (Fig.\ref{fig:tasks_overview}) was to infer the clinical significance (CS) of a gene variant for drug treatment in a given cancer type. For example, considering examples of resistance mutations from the CIViC dataset, can the model correctly classify that the T790M variant of the EGFR gene in lung cancer confers resistance to gefitinib?
  
 Sentences describing genuine relationships were generated using quadruples of entities extracted from CIViC, following the pattern: 

``\texttt{[variant entity]} of \texttt{[gene entity]} identified in \texttt{[disease entity]} is associated with \texttt{[drug entity]}''

 An evidence item in the  KB contains a variant, gene, disease, drugs and CS, so a quadruple can be extracted directly from the  KB, and there are no false quadruples. Only unique quadruples were used to create the dataset. In the case of combination or substitution of multiple drugs in the evidence item, we replaced \texttt{[drug entity]} with multiple entities joined with the conjunction \textit{and} (e.g. \texttt{[drug entity1]} and \texttt{[drug entity2]} and \texttt{[drug entity3]}).
 
After the filtering in the pre-processing stage, 4 values for CS remained: \textit{Resistant}, \textit{Sensitivity/Response}, \textit{Reduced Sensitivity} and \textit{Adverse Response}. Due to a negligible number of quadruples we excluded the \textit{Adverse Response} class. The class \textit{Reduced Sensitivity}  was joined with \textit{Sensitivity/Response}.
  
Multiple evidence items in CIViC can represent one quadruple. For the purpose of Task 2, only the quadruples with uniform clinical significance were selected (98\% of total), i.e. all evidence items for a unique quadruple describe the same relation.
 


\subsubsection{Balancing the test set}

In order to reduce the bias that some pairs/quadruples containing specific entities are almost always true$\mid$false or sensitive$\mid$resistant, we applied a balancing procedure (Supp Methods \ref{sec:balancing_testset}). We excluded the imbalanced pairs/quadruples from the \textit{test set} in creating a \textit{balanced test set}. Reducing the bias allows us to compare the test results more fairly.

\subsection{Model building}
\subsubsection{Baseline model}

In this paper, we used a naive classification model (Nearest Neighbors Classification model  \cite{Fix1989DiscriminatoryA}) as a baseline.  The intent behind this baseline was to contrast a transformer-based model with a simple, non pre-trained model (K-Nearest Neighbour (KNN)). This is to control for the role of the pre-training (i.e. transformer models would show better performance as a result of knowledge embedded in the model, and not due to the relations expressed in the training set). The KNN baseline is used as a control to assess the performance achieved solely due to the distribution of entities in the dataset, as KNN does not embed any distributional knowledge.

Briefly, each entity was represented as a sparse, one-hot encoded vector such that, e.g. for genes, the length of the vector was equal to the total number of genes, and the element corresponding to the given gene was set to 1, whilst all other elements were set to 0. The model was trained and validated for each task based on subsets of the CIViC data as described below. 

For task 1, each pair of vectors (representing each pair of entities) was concatenated as an input; for task 2, sets of 4 vectors, representing \textit{variant}, \textit{gene}, \textit{disease} and \textit{drug} entities were concatenated. Note that vectors for \textit{drug} entities may contain multiple 1-values because some sentences may mention more than one drug.

\subsubsection{Transformers}

In this work, we transfer pairs and evidence sentences into text sequences as input data of both BioBERT and BioMegatron; aggregate the outputs of transformers into one vector representation for each input sequence; and stack classification layers on top of this vector representation for our defined pairs/sentences classification tasks.

Specifically, in Task 1 when predicting the relation between a gene entity and a drug entity, we can input the following sequence into the model:

$seq_{drug\_gene}$=``[CLS] \texttt{[drug entity]} is associated with \texttt{[gene entity]} [SEP]''

Similarly, for the relationship between a variant entity and a drug entity:

$seq_{drug\_variant}$=``[CLS] \texttt{[drug entity]} is associated with \texttt{[variant entity]} [SEP]''

And for a pair of gene and variant entities:

$seq_{variant\_gene}$=``[CLS] \texttt{[variant entity]} is associated with \texttt{[gene entity]}  [SEP]''

In Task 2, for a sentence representing a clinical significance, we define the input sequence as:

$seq_{sentence}$=``[CLS] \texttt{[variant entity]}  of  \texttt{[gene entity]}  identified in \texttt{[disease entity]}
is associated with \texttt{[drug entities]}[SEP]''

Pre-trained BioBERT and BioMegatron were fine tuned: for pairs (gene-variant, gene-drug, variant-drug true/false) classification, 5 epochs 3e-5 learning rate; for quadruple classification, 5 epochs, 1e-4 learning rate.
For more details please refer to Supp Methods \ref{supp_transformers}.

\subsection{Probing}

This section describes the semantic probing methodology implemented in order to shed light on the obtained representations from Task 1 and Task 2. All probing experiments have been performed using the Probe-Ably \footnote{\url{https://github.com/ai-systems/Probe-Ably/}} framework, with default configurations.

Probing is the training of an external classifier model (also called a “probe”) to determine the extent to which a set of auxiliary target feature labels can be predicted from the internal model representations \cite{ferreira-etal-2021-representation, hewitt2019structural, pimentel2020information}. Probing is often performed as a \textit{post hoc} analysis, taking a pre-trained or fine-tuned model and analysing the obtained embeddings. For example, previous probing studies \cite{rives2021biological} have found that training language models across amino acid sequences can create embeddings that encode biological structure at multiple levels, including proteins and evolutionary homology. Knowledge of intrinsic biological properties emerges without supervision, i.e., with no explicit training to capture such property.
 
 As previously highlighted, Task 1 has three different subtasks: classifying the existence of three different pairs of entities in the dataset (drug-gene, drug-variant and variant-gene). For each task, we obtain a fine-tuned version of BioBERT and BioMegatron. For Task 2, only one fine-tuned version is produced for each model. One crucial question is: \textit{Do such models retain the meaning of those entities when fine-tuning the models?} One way of examining such properties is by testing if such representations can still correctly map the entities to their type (e.g., taking the representation of the word Tamoxifen and correctly classifying it as a drug).
 
  Intending to answer this question, we implement the following probing steps:
  
      1. Generate the representations (embeddings) obtained by the fine-tuned (for Task 1 and Task 2) and non fine-tuned models (BioBERT and BioMegatron) for each entity: drug, variant, gene and disease, for each sentence in the test set. We also include BERT-base to the analysis in order to assess the performance of a more general model. Even though most of the entities are composed of a single word, these models depend on the WordPiece tokenizer, often breaking a word into separate pieces. For example, the word Tamoxifen is tokenized as four pieces: \texttt{[Tam, \#\#ox, \#\#ife, \#\#n]} using the BioBERT tokenizer. To obtain a single vector for each entity, we compute the average of all the token representations composing that word. For instance, the word Tamoxifen is represented as a vector containing the average of the vectors representing each of its four pieces.
      
      2. The goal of probing is merely to find what information is already stored in the new model, not to train a new task. Thus, following standard probing guidelines \cite{ferreira-etal-2021-representation}, we split the representations into training, validation and test set, using a 20/40/40 scheme. By such split, we want to limit the number of instances seen during training and avoid overfitting over a large part of the dataset, since part of the dataset was already observed during the first task training, and the information is partly stored in the generated vectors. The model overfitting is also prevented with the use of a linear model.
      Each model is trained for 5 epochs, with the validation set being used to selected the best performing model (in terms of accuracy). 
      
      3. After obtaining all representations for each model and respective entity types, we train a total of 50 linear probes to classify each representation into the correct entity label. The number 50 is a default configuration and recommended value from the Probe-Ably framework. These different 50 models are contrasted using a measure of complexity. When using models containing a large number of parameters, there is a possibility that the probing training will reshape the representation to fit the new task, leading to inconclusive results, therefore, we opt for a simpler linear model, to avoid this phenomena.  We follow previous research in probing \cite{pimentel2020information}, measuring the complexity of a linear model $ \hat{y} = W \mathbf{x} + \mathbf{b} $ by using the nuclear norm of the weight matrix $W$, computed as:
    $$
         ||\mathbf{W}||_{*} = \sum_{i=1}^{min(|\mathcal{T}|, d)}\sigma_i(\mathbf{W}).
    $$
    where $\sigma_i(W)$ is the i-th singular value of $W$, $\mathcal{|T|}$ is the number of targets (e.g., number of possible entities) and $d$ is the number of dimensions in the representation (e.g., 768 dimensions for BERT-base).
    
    The nuclear norm is then included in the loss (weighted by a parameter $\lambda$) 
    $$ 
    -\sum_{i=1}^{n}\log p(t^{(i)} \mid \mathbf{h}^{(i)}) + \lambda \cdot ||\mathbf{W}||_{*}  $$
    and is thus regulated in the training loop , where \textit{t }is a single value of $\mathcal{T}$. In order to obtain 50 different models, we randomly initialize the dropout and $\lambda$ parameter.  As suggested in \cite{pimentel2020information}, we show the results across all the different initializations in Figures \ref{fig:probing_results:task1},\ref{fig:probing_results:task2}.
    Having models with different complexity allows us to see if the results are consistent across different complexities, with the best performance usually being obtained by the more complex models.
    
    
    4. For each trained probe, we also train an equivalent control probe. The control probe is a model trained for the same task as the original probe, however, the training is performed using random labels, instead of the correct ones. Having a control task can been seen as an analogy to having a study with placebo medication. When the performance on the probing task is better than the control task, it is known that the probe model is capturing more than random noise.
    
    
    5. The performance of the probes is measured in terms of \textit{Accuracy} and \textit{Selectivity} for the test set. The selectivity score, namely the difference in accuracy between the representational probe and a control probing task with randomised labels, indicates that the probe architectures used are not expressive enough to "memorise" unstructured labels. Ensuring that there is no drop-off in selectivity increases the confidence that we are not falsely attributing strong accuracy scores to the representational structure where over-parameterised probes (i.e, probes that contain several learnable parameters) could have explained them.

\subsection{Clustering}
In addition to the evaluation of models’ performance  in a probing setting, we investigated with the support of clustering methods  whether the output vectors can identify potential relationships between entity pairs and/or quadruples. 

For clustering the output in Task 1 and 2 we used hierarchical agglomerative clustering (HAC) with Ward variance minimization algorithm (ward linkage) and euclidean distance as distance metric on both the rows (output dimensions) and the columns (vector representations of true pairs). Then we identified clusters using a distance threshold defined pragmatically after visual investigation of the clustermap and dendrogram. For clustering the output used in Probing, we used HDBSCAN \cite{mcinnes2017hdbscan,mcinnes2017accelerated}, with parameter min cluster $\text{size}=120$,  while the remaining parameters kept their default values.

We applied Uniform Manifold Approximation and Projection for Dimension Reduction (UMAP) \cite{mcinnes2018umap-software} to compare patterns observable after dimensionality reduction into 2 dimensions with clusters obtained via HAC.
UMAP parameters: default (n components$=2$, n neighbors$=15$)


The UMAP representation constitutes multiple distinct groups that contain various entity types or target labels. To quantify that, the HDBSCAN algorithm was used, which identifies clusters of densely distributed points. We used homogeneity metric as a measure of proportion of various labels in one cluster. It can be defined as ratio of the count of most common label in the cluster and the total count in the cluster, e.g. if a cluster contains 40 drugs and 10 genes, homogeneity equals 0.8. Ideally, all clusters would score 1.


\section{Results}

\subsection{Can transformers recognize existing  relations/associations? - Task 1}

\subsubsection{Distribution of entities in pairs}

A total of 8,032 entity pairs were included in this analysis - 5,320 (66\%) in the training set, 2,412 in the imbalanced test set and 1,090 in the balanced test set (Table \ref{tab:pairs_counts_summary}).


\begin{table}
\centering
\caption{ Statistics about the datasets used in Task 1: number of unique pairs and entities.}
\label{tab:pairs_counts_summary}
 
\resizebox{\textwidth}{!}{
\begin{tabular}{ccccccccccc} 
\midrule
               & \multicolumn{4}{c}{Pairs (both True and   False) (n)}                                                       & \multicolumn{3}{c}{Unique (n)} & \multicolumn{3}{c}{\begin{tabular}[c]{@{}c@{}}Unique in balanced\\ test set (n)\end{tabular}}  \\
               & Total & Train set & Test set & \begin{tabular}[c]{@{}c@{}}Balanced test set\\ (\% of test set)\end{tabular} & Genes & Variants & Drugs       & Genes & Variants & Drugs                                                                      \\ 
\midrule
drug - variant & 3676  & 2272      & 1104     & 418 (38\%)                                                                   & -     & 897      & 242         & -     & 321      & 134                                                                        \\
drug - gene    & 2480  & 1736      & 744      & 396 (53\%)                                                                   & 302   & -        & 432         & 235   & -        & 193                                                                        \\
variant - gene & 1876  & 1312      & 564      & 276 (49\%)                                                                   & 125   & 910      & -           & 72    & 235      & -                                                                          \\
\bottomrule
\end{tabular}}
\end{table}

Entities in the dataset were distributed non-uniformly, resembling a Pareto distribution. For drug-gene pairs, the majority of pairs involving the most common genes and drugs were true (Supp Fig.\ref{fig:top_50_pairs}a). A similar pattern was observed for drug-variant pairs (Supp Fig.\ref{fig:top_50_pairs}b). In contrast, for variant-gene pairs, the majority of pairs involving the most common variant entities were false (Supp Fig.\ref{fig:top_50_pairs}c). 

\subsubsection{Performance}
We evaluated the classification performance both on the test set and balanced test set using area under the Receiver Operator Characteristic curve (AUC, Table \ref{tab:performance_task1}). 

\begin{table}[hbt!]
\centering
\caption{AUC in classification task 1.}
\label{tab:performance_task1}
\begin{tabular}{@{}lllll@{}}
\toprule
\multicolumn{1}{c}{\multirow{2}{*}{\textbf{Pairs + Model}}} &
  \multicolumn{2}{c}{\textbf{Imbalanced}} &
  \multicolumn{2}{c}{\textbf{Balanced}} \\
\multicolumn{1}{c}{} &
  \multicolumn{1}{c}{\textbf{Test set}} &
  \multicolumn{1}{c}{\textbf{10fold CV (sd)}} &
  \multicolumn{1}{c}{\textbf{Test set}} &
  \multicolumn{1}{c}{\textbf{10fold CV (sd)}} \\ \midrule
\multicolumn{5}{l}{{ \textbf{Drug-Variant}}}             \\
KNN (baseline) & 0.771 & .821 (.023)  & 0.486 & .444 (.044) \\
BioBERT        & 0.834 & .856 (.027)  & 0.59  & .569 (.033) \\
BioMegatron    & 0.847 & .850 (.022)  & 0.642 & .580 (.070) \\ \midrule
\multicolumn{5}{l}{{ \textbf{Drug-Gene}}}                \\
KNN (baseline) & 0.705 & .770 (.025)  & 0.492 & .425 (.037) \\
BioBERT        & 0.743 & .762 (.024)  & 0.544 & .506 (.048) \\
BioMegatron    & 0.722 & .755 (.045)  & 0.572 & .512 (.055) \\ \midrule
\multicolumn{5}{l}{{ \textbf{Variant-Gene}}}             \\
KNN (baseline) & 0.683 & .778 (0.022) & 0.434 & .413 (.056) \\
BioBERT        & 0.826 & .855 (.033)  & 0.677 & .669 (0.62) \\
BioMegatron    & 0.828 & .813 (.078)  & 0.671 & .627 (.104) \\ \bottomrule
\end{tabular}
\end{table}

In all cases, performance was superior for the imbalanced dataset compared with the balanced dataset.  As the usage of balanced test set is to adjust the analysis for frequent pairs with consistent labels (almost all true or all false), the drop in the performance suggests that the fine-tuned models are sensitive to the distribution bias in the training set and learn statistical regularities. They favor more frequent pairs and disfavor less frequent ones, which aligns with previous research \cite{nadeemStereoSetMeasuringStereotypical2021, gehmanRealToxicityPromptsEvaluatingNeural2020, mccoyRightWrongReasons2019,zhongFactualProbingMASK2021b,gururanganAnnotationArtifactsNatural2018,minSyntacticDataAugmentation2020}. 

Performance of the transformers was superior to the  baseline model in all cases, except for drug-gene classification against the imbalanced dataset.  For the drug-gene scenario, the AUC is close to 0.5 which means that classification resembles a random guessing and very limited, if any biological knowledge is utilized (\ref{itm:RQ1}). Considering only the performance in Task 1, there is no significant difference between BioBERT and BioMegatron, establishing an equivalence of both representations in the context of this task (\ref{itm:RQ3}).

\subsubsection{The impact of imbalance on model's error}
As we observed significant differences between performance on the imbalanced and balanced test sets, we investigated further the specifics of this phenomenon, i.e. classification error for individual pairs. One or more evidence items can represent each pair, i.e. each pair can be found in one or more scientific papers. Similarly to entities distribution, there is an imbalance in the  number of evidence items related to pairs. For example, 73\% of variant-drug pairs are supported only by one, 2\% by $>$ 2, and 1.4\% by$ >=$10 evidence items. Details for all 3 types of pairs are shown in the Table \ref{tab:evidence_imbalance_in_pairs}.
\begin{table}[hbt!]
\centering
\caption{Number of evidence items related to the type of pair in the dataset.}
\label{tab:evidence_imbalance_in_pairs}
\begin{tabular}{@{}llllll@{}}
\toprule
\multicolumn{1}{c}{\multirow{2}{*}{\textbf{Pair}}} & \multicolumn{5}{c}{\textbf{Number of evidence items}}                  \\
\multicolumn{1}{c}{} &
  \multicolumn{1}{c}{\textbf{\textbf{1}}} &
  \multicolumn{1}{c}{\textbf{\textbf{\textgreater{}1}}} &
  \multicolumn{1}{c}{\textbf{\textbf{\textgreater{}2}}} &
  \multicolumn{1}{c}{\textbf{\textbf{\textgreater{}=10}}} &
  \multicolumn{1}{c}{\textbf{\textbf{\textgreater{}=20}}} \\ \midrule
gene-drug ($n=1240$)                                & 792 (64.1\%)  & 442 (32.9\%) & 127 (12.7\%) & 82 (6.9\%) & 12 (0.97\%) \\
variant-gene ($n=938$)                              & 296 (63.2\%)  & 342 (36.2\%) & 117 (12.2\%) & 49 (2.2\%) & 2 (0.2\%)   \\
variant-drug ($n=1838$)                             & 1347 (73.3\%) & 491 (26.7\%) & 91 (2\%)     & 22 (1.4\%) & 1 (0.02\%)  \\ \bottomrule
\end{tabular}
\end{table}

\begin{figure}[htb!]
\centering
\begin{subfigure}{.85\textwidth}
  \centering
  \includegraphics[width=\linewidth]{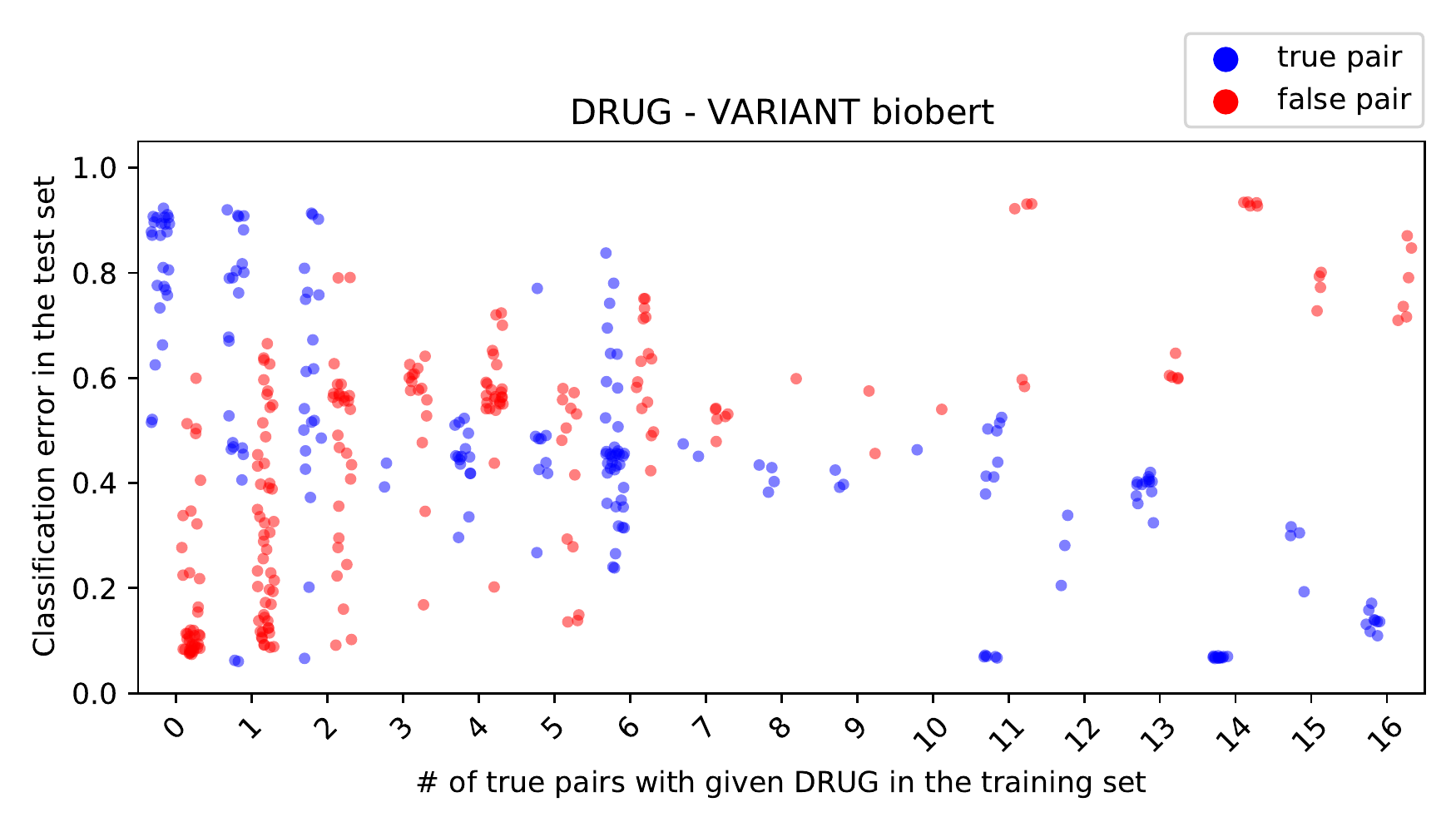}
  \caption{Classification error in relation to number of true pairs in the training set containing the entity.}
  \label{fig:error_vs_nb_of_true_pairs}
\end{subfigure}%
\\
\begin{subfigure}{.85\textwidth}
  \centering
  \includegraphics[width=\linewidth]{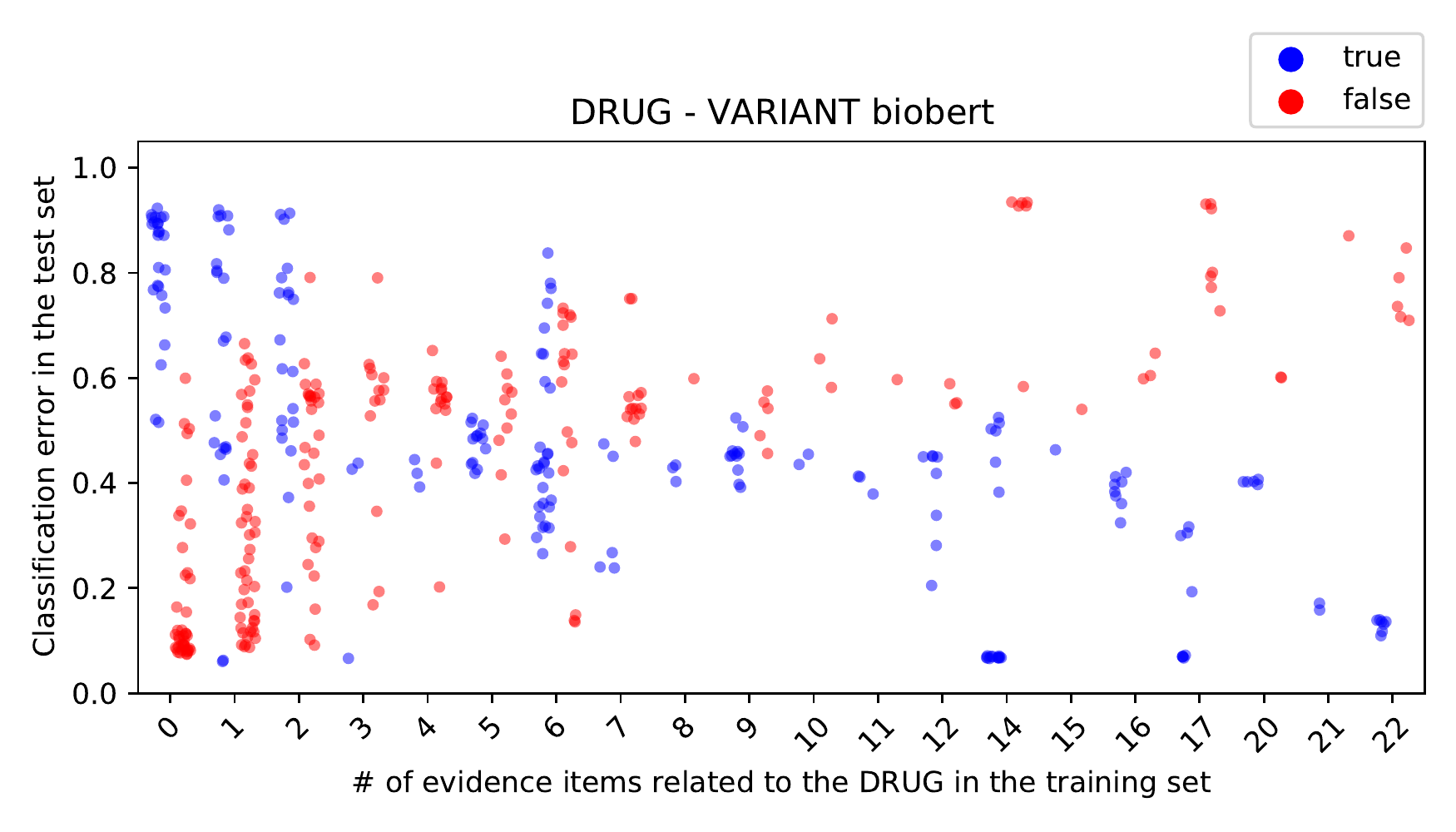}
  \caption{Classification error in relation to  number of evidence items (i.e. scientific papers) describing the entity.}
  \label{fig:error_vs_nb_evid_items}
\end{subfigure}
\caption{ Evaluation of the impact of the dataset imbalance on model's performance: The more true pairs in the training set containing a DRUG entity (a), or the more evidence items related to a DRUG entity in the knowledge base (b), the higher change for a pair (containing the DRUG entity) of being classified as true.}
\end{figure}

Classification error on the balanced test set varied according to the frequency of true pairs in the dataset - for drugs that occurred frequently in the training set (Fig.\ref{fig:error_vs_nb_of_true_pairs}) or in the knowledge base (Fig.\ref{fig:error_vs_nb_evid_items}), true drug-variant pairs were typically classified correctly, whilst false drug-variant pairs were typically misclassified.

 The analysis of error quantifies the impact of the imbalance in the dataset on the performance (\ref{itm:RQ4}). It shows that if an entity occurs in many true pairs in the training set, an unseen pair containing the entity from the test set is likely to be classified as true, regardless of biological meaning. Fine-tuned transformers are highly influenced by learnt statistical regularities. For instance, pairs with drugs that occur in 15 true pairs in the training set obtain error $<$0.1 for true pairs and error $>$0.7 for false pairs (Fig.\ref{fig:error_vs_nb_of_true_pairs}) as to all of them the model assigns a high probability of being true. This applies to the drug (significant Spearman correlation, $p < 0.001$), gene ($p < 0.001$) and variant entities ($p < 0.05$). All correlations are summarized in Supp Table \ref{tab:correlations_error_nb_pairs}.

 Similar correlation is observed regarding the error and the number of evidence items in the knowledge base. The more evidence items related to an entity, the higher change for a pair (containing this entity) of being classified as true. For instance, if a pair contains a drug which is supported by only one evidence item, the pair is more likely to be labeled as false (Fig.\ref{fig:error_vs_nb_evid_items}). 

 This can be a major concern in a applications in cancer precision medicine. There is little value of  being accurate for well-known relations and facts. The true potential is for the less obvious queries, which the experts are less familiar with. However, as shown above, biomedical transformers suffer from reduced performance for underrepresented cases in the dataset (\ref{itm:RQ4}).

\subsection{Can transformers recognize clinical significance of a relation? - Task 2}
 
\subsubsection{Distribution of entities in quadruples}
A total of 2,989 quadruples were included in this analysis, 897 in the test set. As a result of balancing the test set, 207 quadruples are left for further investigation of the output vectors. It comprised 147 unique variants, 67 genes, 43 diseases and 89 drugs (see Table \ref{tab:number_of_unique_entities_quads}).

Similar to the observed distribution of entity pairs, the distribution of entities among the quadruples was also non-uniform, with a Pareto distribution - the most common variant entity was \textit{MUTATION}, the most common gene entity was \textit{EGFR}, the most common disease was \textit{Lung Non-small Cell Carcinoma} and the most common drug was \textit{Erlotinib} (see Supp Fig.\ref{fig:top_30_quads}).  

In most cases (64\%), the clinical significance of quadruples in the dataset was \textit{Sensitivity/Response}. The imbalance between \textit{Sensitivity/Response} and \textit{Resistance} was most evident for the most common variants (\textit{MUTATION, OVEREXPRESSION, AMPLIFICATION, EXPRESSION, V600E, LOSS, FUSION, LOSS-OF-FUNCTION and UNDEREXPRESSION}), where approximately 80\% of quadruples related to drug sensitivity.     

\begin{table}
\centering
\caption{ Statistics about the datasets used in Task 2: number of unique quadruples and entities.}
\label{tab:number_of_unique_entities_quads}
 
\begin{tabular}{cccccc} 
\toprule
\multirow{2}{*}{\textbf{Dataset}} & \multicolumn{5}{c}{\textbf{Unique (n)}}                                                    \\
                                  & \textbf{Quadruples} & \textbf{Variant} & \textbf{Gene} & \textbf{Disease} & \textbf{Drug}  \\ 
\midrule
\textbf{Total}                    & 2989                & 1015             & 302           & 215              & 733            \\
\textbf{Training set}             & 2092                & 803              & 258           & 186              & 579            \\
\textbf{Test set}                 & 897                 & 432              & 165           & 135              & 339            \\
\textbf{Balanced test set}             & 207                 & 147              & 67            & 43               & 89             \\
\bottomrule
\end{tabular}
\end{table}

\subsubsection{Performance}
We evaluated the performance of the models in predicting clinical significance of quadruples using AUC. In all cases, performance of the transformer models was superior to that of the  KNN (non pre-trained) baseline. Similar to the results for classification of entity pairs, performance was superior for the imbalanced dataset compared with the balanced dataset. Nevertheless, both BioBERT and BioMegatron achieved high accuracy (AUC greater than 0.8) on the balanced dataset (Table \ref{tab:performance_task2}).  No significant difference between BioBERT and BioMegatron was observed (\ref{itm:RQ3}).
Compared to the performance in Task 1, we observe a smaller drop in AUCs between the imbalanced and balanced test set, while the difference between transformers and KNN is significantly higher. This suggests that in the more complex Task 2, fine-tuned BioBERT and BioMegatron exploit some of the biological knowledge encoded within the architecture (\ref{itm:RQ1}). This accentuated difference between pre-trained and transformer-based baselines (when contrasted to the previous task), demonstrates that the benefit of the pre-training component of transformers can be better observed in the context of complex n-ary relations (\ref{itm:RQ2}).

\begin{longtable}[c]{@{}ccccc@{}}
\caption{AUC in classification task 2  for imbalanced and balanced test set. CV - Cross Validation, sd - standard deviation.}
\label{tab:performance_task2}\\
\toprule
\multirow{3}{*}{AUC} & \multicolumn{4}{c}{ Binary classification of quadruples} \\* \cmidrule(l){2-5} 
 & \multicolumn{2}{c}{Imbalanced} & \multicolumn{2}{c}{Balanced} \\
 & Test set & 10fold CV (sd) & Test set & 10fold CV (sd) \\* \cmidrule(r){1-1}
\endfirsthead
\multicolumn{5}{c}%
{{\bfseries Table \thetable\ continued from previous page}} \\
\toprule
\multirow{3}{*}{AUC} & \multicolumn{4}{c}{Quadruples} \\* \cmidrule(l){2-5} 
 & \multicolumn{2}{c}{Imbalanced} & \multicolumn{2}{c}{Balanced} \\
 & Test set & 10fold CV (sd) & Test set & 10fold CV (sd) \\* \cmidrule(r){1-1}
\endhead
\bottomrule
\endfoot
\endlastfoot
KNN   (baseline) & 0.878 & .864 (.023) & 0.753 & .655 (.065) \\
BioBERT & 0.898 & .904 (.024) & 0.806 & .835 (.060) \\
BioMegatron & 0.905 & .910 (.022) & 0.826 & .833 (.037) \\* \bottomrule
\end{longtable}

\subsubsection{Model's error vs strength of biomedical evidence}
High confidence associations (\textit{Evidence rating = 5}) were rare - most quadruples in the balanced test set were either unrated or evidence level 3 (\textit{Evidence is convincing, but not supported by a breadth of experiments}).  

\begin{figure*}[tb]
\centering
\includegraphics[width= .9\textwidth]{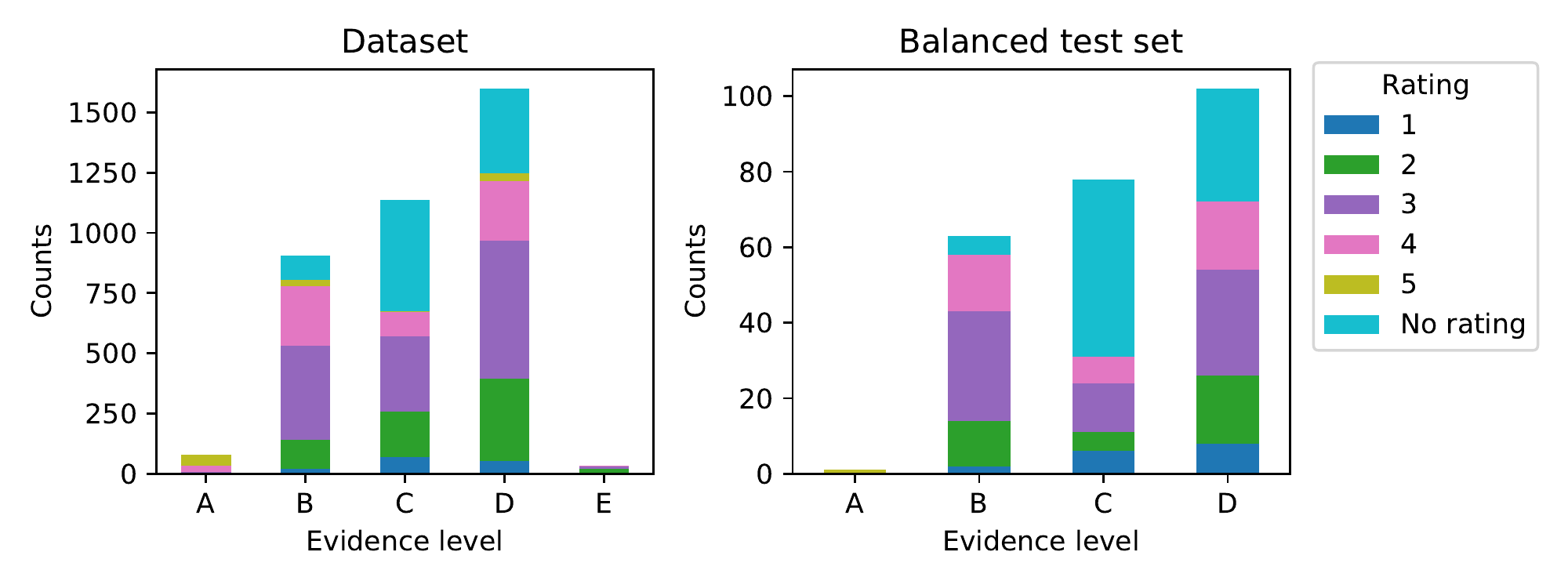}
\caption{Number of evidence items in the datasets stratified by evidence level and evidence rating.}
\label{fig:evidence_level_vs_counts}
\end{figure*}

The most common type of evidence (denoted by the \textit{Evidence level} attribute) described by quadruples in the dataset was \textit{D - Preclinical evidence}; validated associations (\textit{Evidence level = A}) were rare - only a single example remained in the test set after balancing. No inferential associations (\textit{Evidence level = E}) remained in the balanced test set (Fig.\ref{fig:evidence_level_vs_counts}). 

In the balanced test set, considering all levels of evidence, there was no correlation between level of evidence and model performance (p$>$0.05, Spearman correlation).  Thus, we do observe that transformers are not better in classifying relations, that are supported by strong evidence in the knowledge base. Quite the opposite, AUCs for evidence level B were lower (.683 and .703) than for C and D (BioBert: .900 and .812; BioMegatron: .939 and .816, see Supp.Table\ref{tab:auc_brier_by_evidence_level}).  Considering pre-clinical evidence only (\textit{Evidence level D}), the KNN model had significantly higher error compared with BioBERT ( Mann-Whitney U test: p=0.014) and BioMegatron (p=0.007). This finding was supported by AUC and Brier scores (Supp Table \ref{tab:auc_brier_by_evidence_level}).

\newpage

\subsubsection{Misclassified well known relations}
A total of 16 well-known relations, defined as Evidence level A (\textit{Validated association}) or B (\textit{Clinical evidence}) and Evidence rating 5 (\textit{Strong, well supported evidence from a lab or journal with respected academic standing}) or 4 (\textit{Strong, well supported evidence}) were identified in the balanced test set (Table \ref{tab:well_known_relations}).

Despite the higher confidence assigned to these quadruples, the models did not perform better against these relations compared with the overall balanced test set - AUC for these quadruples was 0.75, 0.78 and 0.75 for BioBERT, BioMegatron and KNN, respectively. For example, high classification error rates ($\geq$ .6) were observed for transformer models for the following quadruples: 

\begin{itemize}
\item \textit{EXPRESSION - HSPA5 - Colorectal Cancer - Fluorouracil}
\item \textit{EXPRESSION - PDCD4 - Lung Cancer - Paclitaxel}
\item \textit{V600E - BRAF - Colorectal Cancer - Cetuximab and Encorafenib and Binimetinib} (BioMegatron only)
\end{itemize}

From a cancer precision medicine perspective, these are significant misclassifications elicit the safety limitations of these models when considering clinical applications. In previous paragraphs we show that high error is expected for underrepresented relations, while here we demonstrate that transformers can fail even for well-known, strong evidence relations (\ref{itm:RQ1}).

\begin{table}[b!]
\centering
\caption{List of 16 well known relations and corresponding classification error. R stands for \textit{Resistance} and S/R is for \textit{Sensitivity/Response}}
\label{tab:well_known_relations}
\resizebox{\textwidth}{!}{%
\begin{tabular}{@{}llllllllll@{}}
\toprule
\multicolumn{1}{c}{\textbf{Variant}} &
  \multicolumn{1}{c}{\textbf{Gene}} &
  \multicolumn{1}{c}{\textbf{Diseases}} &
  \multicolumn{1}{c}{\textbf{Drugs}} &
  \multicolumn{1}{c}{\textbf{\begin{tabular}[c]{@{}c@{}}Clinical\\  significance\end{tabular}}} &
  \multicolumn{1}{c}{\textbf{\begin{tabular}[c]{@{}c@{}}BioBERT\\ error\end{tabular}}} &
  \multicolumn{1}{c}{\textbf{\begin{tabular}[c]{@{}c@{}}BioMegatron\\  error\end{tabular}}} &
  \multicolumn{1}{c}{\textbf{\begin{tabular}[c]{@{}c@{}}KNN\\  rror\end{tabular}}} &
  \multicolumn{1}{c}{\textbf{\begin{tabular}[c]{@{}c@{}}Evidence\\  level\end{tabular}}} &
  \multicolumn{1}{c}{\textbf{Rating}} \\ \midrule
EXON 2 MUTATION                      & KRAS                              & Pancreatic Cancer                                                       & \begin{tabular}[c]{@{}l@{}}Erlotinib and\\Gemcitabine\end{tabular}                  & R                                                                                                     & 0.895                                                                                         & 0.270                                                                                             & 0.2                                                                                       & B                                                                                              & 4                                    \\
EXPRESSION                           & EGFR                              & Colorectal Cancer                                                       & Cetuximab                                                                           & S/R                                                                                                   & 0.280                                                                                         & 0.296                                                                                             & 0.4                                                                                       & B                                                                                              & 4                                    \\
EXPRESSION                           & FOXP3                             & Breast Cancer                                                           & Epirubicin                                                                          & S/R                                                                                                   & 0.153                                                                                         & 0.776                                                                                             & 0.6                                                                                       & B                                                                                              & 4                                    \\
EXPRESSION                           & HSPA5                             & Colorectal Cancer                                                       & Fluorouracil                                                                        & S/R                                                                                                   & 0.845                                                                                         & 0.608                                                                                             & 0.4                                                                                       & B                                                                                              & 4                                    \\
EXPRESSION                           & PDCD4                             & Lung Cancer                                                             & Paclitaxel                                                                          & S/R                                                                                                   & 0.954                                                                                         & 0.939                                                                                             & 0.4                                                                                       & B                                                                                              & 4                                    \\
EXPRESSION                           & AREG                              & Colorectal Cancer                                                       & Panitumumab                                                                         & S/R                                                                                                   & 0.434                                                                                         & 0.120                                                                                             & 0.4                                                                                       & B                                                                                              & 4                                    \\
EXPRESSION                           & EREG                              & Colorectal Cancer                                                       & Panitumumab                                                                         & S/R                                                                                                   & 0.345                                                                                         & 0.202                                                                                             & 0.6                                                                                       & B                                                                                              & 4                                    \\
ITD                                  & FLT3                              & \begin{tabular}[c]{@{}l@{}}Acute Myeloid\\ Leukemia\end{tabular}        & Sorafenib                                                                           & S/R                                                                                                   & 0.418                                                                                         & 0.355                                                                                             & 0.6                                                                                       & B                                                                                              & 4                                    \\
K751Q                                & ERCC2                             & Osteosarcoma                                                            & Cisplatin                                                                           & R                                                                                                     & 0.285                                                                                         & 0.827                                                                                             & 0.2                                                                                       & B                                                                                              & 4                                    \\
LOSS-OF-FUNCTION                     & VHL                               & \begin{tabular}[c]{@{}l@{}}Renal Cell \\ Carcinoma\end{tabular}         & \begin{tabular}[c]{@{}l@{}}Anti-VEGF Monoclonal \\ Antibody\end{tabular}            & R                                                                                                     & 0.074                                                                                         & 0.360                                                                                             & 0.8                                                                                       & B                                                                                              & 4                                    \\
MUTATION                             & KRAS                              & Colorectal Cancer                                                       & \begin{tabular}[c]{@{}l@{}}Cetuximab and\\Chemotherapy\end{tabular}                 & R                                                                                                     & 0.067                                                                                         & 0.021                                                                                             & 0                                                                                         & B                                                                                              & 4                                    \\
MUTATION                             & SMO                               & \begin{tabular}[c]{@{}l@{}}Basal Cell\\ Carcinoma\end{tabular}          & Vismodegib                                                                          & R                                                                                                     & 0.062                                                                                         & 0.039                                                                                             & 0                                                                                         & B                                                                                              & 4                                    \\
OVEREXPRESSION                       & IGF2                              & \begin{tabular}[c]{@{}l@{}}Pancreatic\\ Adenocarcinoma\end{tabular}     & \begin{tabular}[c]{@{}l@{}}Gemcitabine and\\Ganitumab\end{tabular}                  & S/R                                                                                                   & 0.068                                                                                         & 0.100                                                                                             & 0.6                                                                                       & B                                                                                              & 4                                    \\
OVEREXPRESSION                       & ERBB3                             & Breast Cancer                                                           & Patritumab Deruxtecan                                                               & S/R                                                                                                   & 0.006                                                                                         & 0.028                                                                                             & 0.2                                                                                       & B                                                                                              & 4                                    \\
PML-RARA A216V                       & PML                               & \begin{tabular}[c]{@{}l@{}}Acute\\Promyelocytic\\ Leukemia\end{tabular} & Arsenic Trioxide                                                                    & R                                                                                                     & 0.161                                                                                         & 0.015                                                                                             & 0.4                                                                                       & B                                                                                              & 4                                    \\
V600E                                & BRAF                              & Colorectal Cancer                                                       & \begin{tabular}[c]{@{}l@{}}Cetuximab and\\Encorafenib and\\Binimetinib\end{tabular} & S/R                                                                                                   & 0.264                                                                                         & 0.761                                                                                             & 0.4                                                                                       & A                                                                                              & 5                                    \\
\bottomrule
\end{tabular}%
}
\end{table}

\subsection{ Does the fine-tuning corrupt the representation of pre-trained models?}

\subsubsection{Recognising entity types from representations of pairs}

\begin{figure}[hbt!]
\centering
\begin{subfigure}{.5\textwidth}
  \centering
  \includegraphics[width=\linewidth]{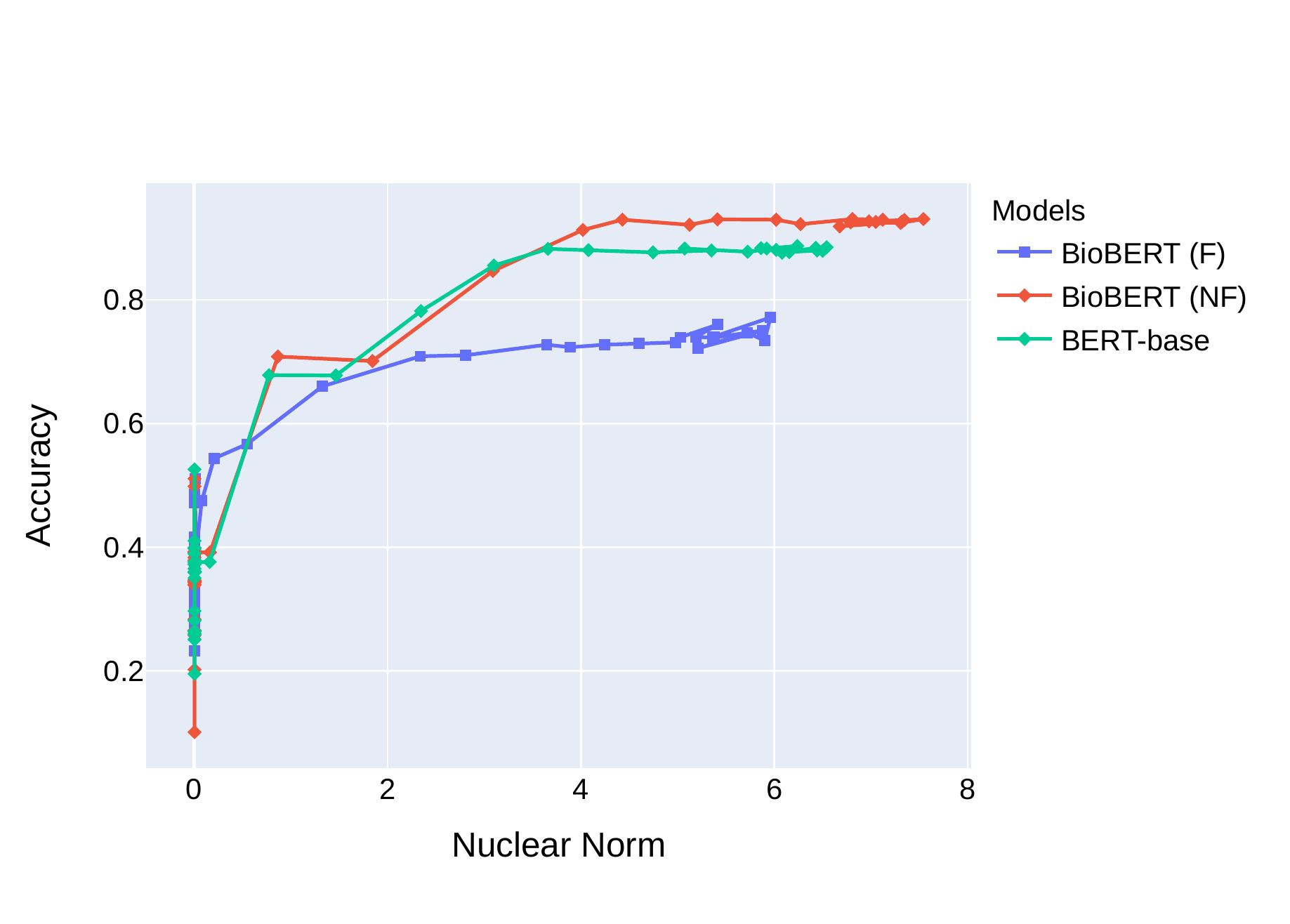}
  \caption{Accuracy VS Nuclear Norm (BioBERT)}
  \label{fig:probing_results:task1:acc_biobert}
\end{subfigure}%
\begin{subfigure}{.5\textwidth}
  \centering
  \includegraphics[width=\linewidth]{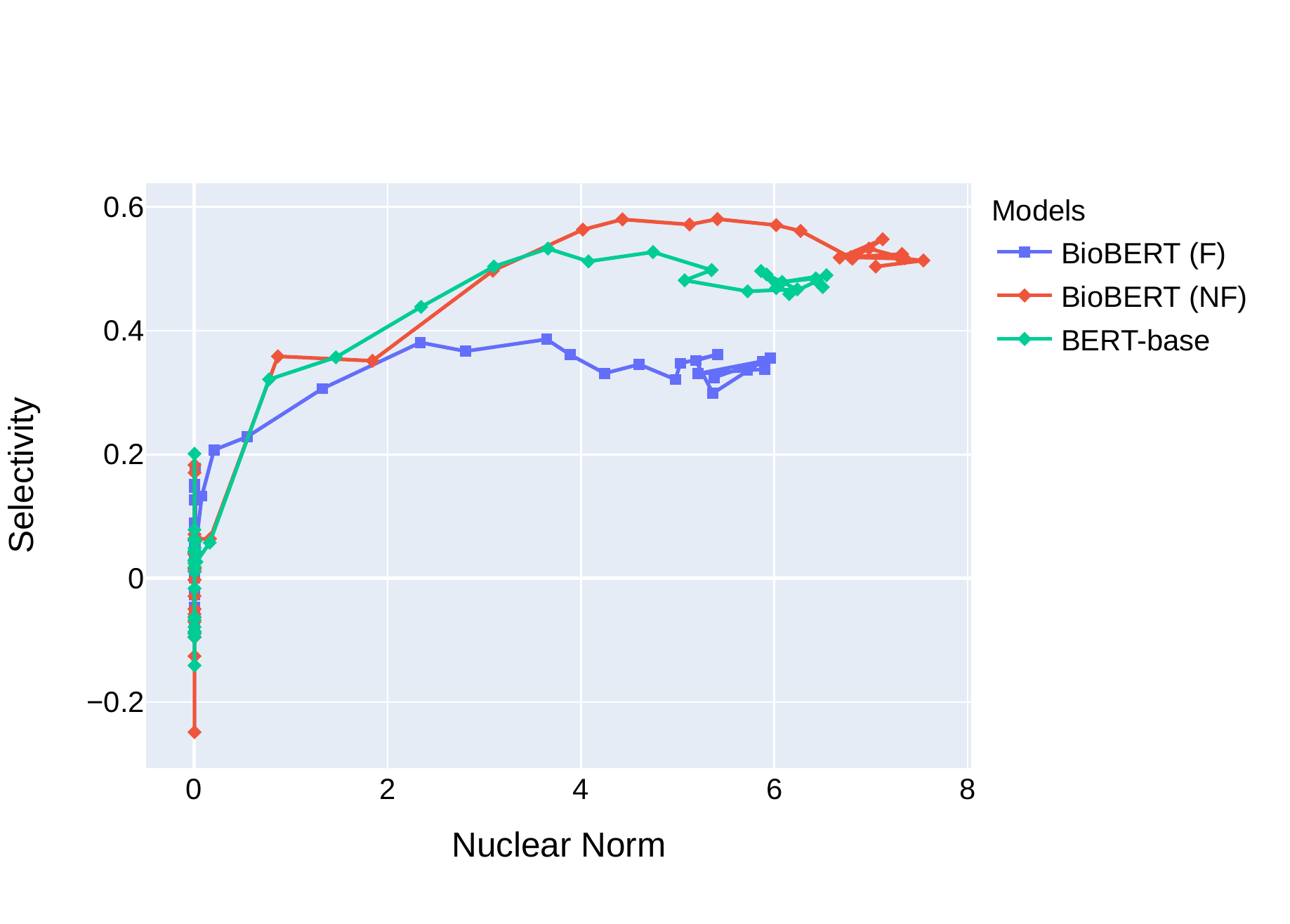}
  \caption{Selectivity VS Nuclear Norm (BioBERT)}
  \label{fig:probing_results:task1:sel_biobert}
\end{subfigure}
\\
\begin{subfigure}{.5\textwidth}
  \centering
  \includegraphics[width=\linewidth]{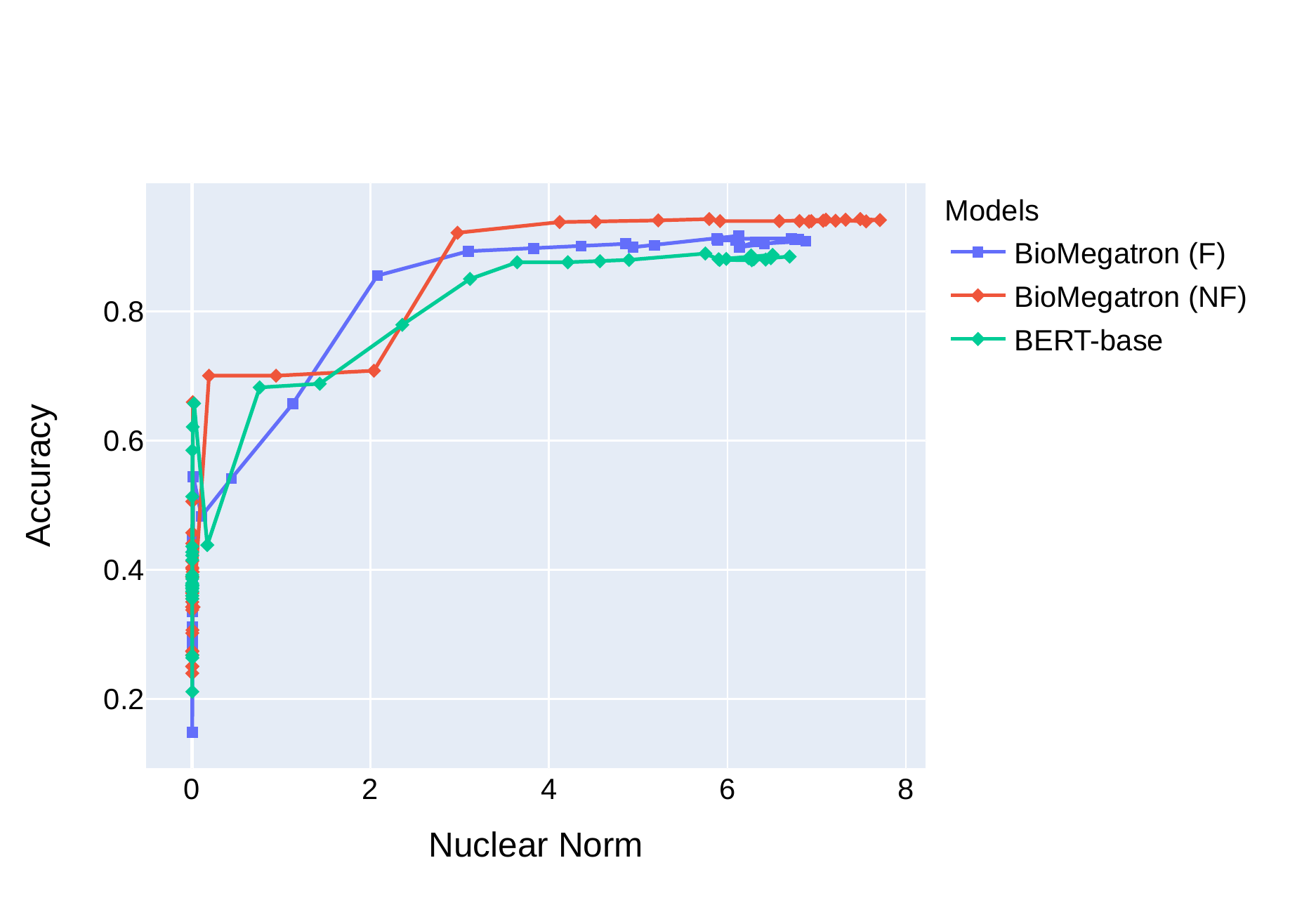}
  \caption{Accuracy VS Nuclear Norm (BioMegatron)}
  \label{fig:probing_results:task1:acc_biomegatron}
\end{subfigure}%
\begin{subfigure}{.5\textwidth}
  \centering
  \includegraphics[width=\linewidth]{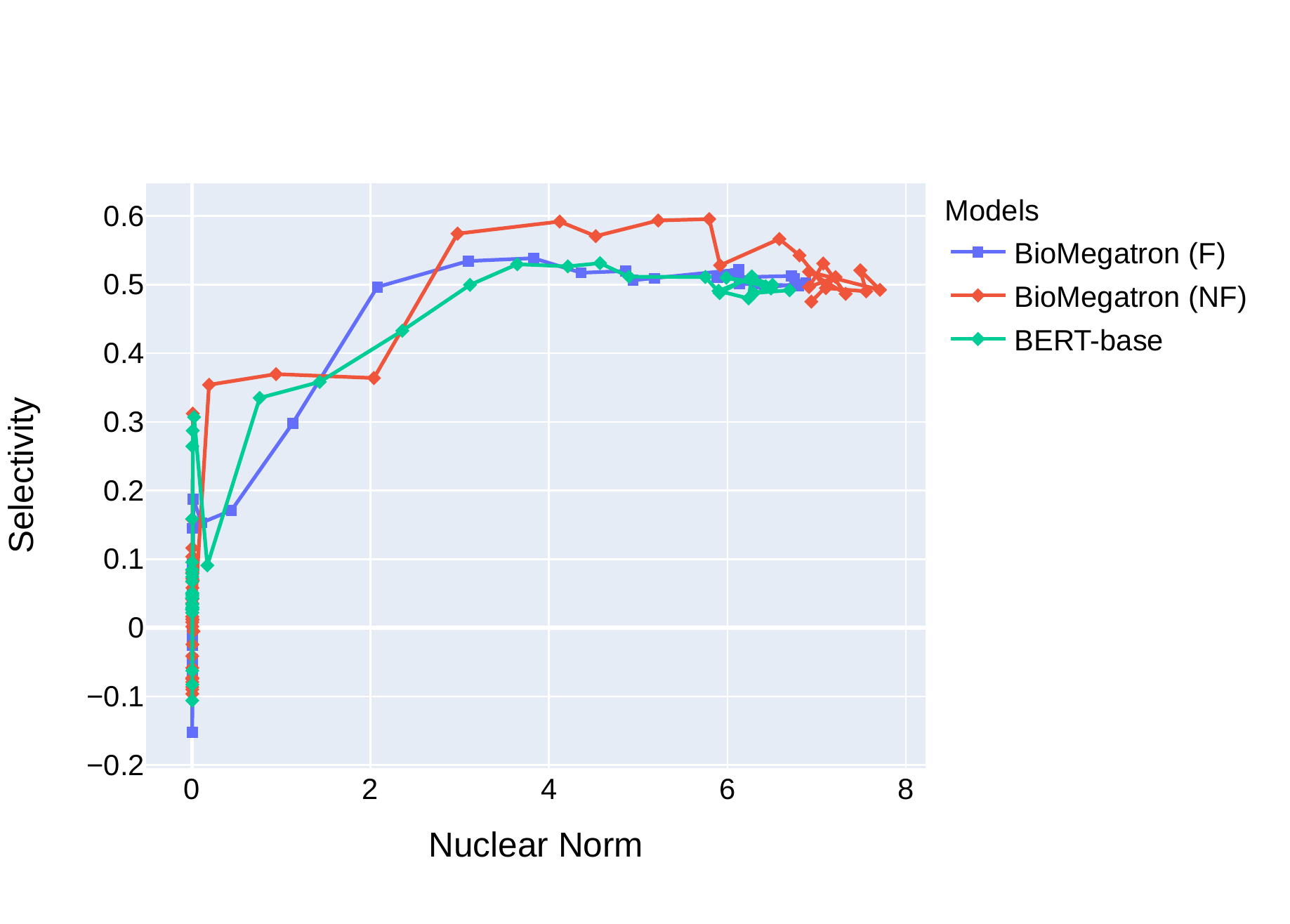}
  \caption{Selectivity VS Nuclear Norm (BioMegatron)}
  \label{fig:probing_results:task1:sel_biomegatron}
\end{subfigure}
\caption{Probing results for models fine-tuned (F) on Task 1, together with the original (non fine-tuned) models (NF). }
\label{fig:probing_results:task1}
\end{figure}

Figure~\ref{fig:probing_results:task1} presents the probing results for Task 1, with the left column containing the Accuracy results and the right column containing the Selectivity results. Selectivity was greater than zero for a control task containing random labels. For BioBERT, both accuracy and selectivity were higher for the non-fine-tuned models compared with the fine-tuned model. In fact, performance of the BERT (base) model was greater than that of the fine-tuned model for this task.  This suggest that BioBERT loses some of the accuracy of background knowledge as a result of fine-tuning. This finding aligns with other works \cite{durraniHowTransferLearning2021,merchantWhatHappensBERT2020,rajaeeHowDoesFinetuning2021a}.
For BioMegatron, performance of the fine-tuned model was slightly worse than non-fine-tuned one, suggesting a similar behaviour for BioMegatron, but in lower magnitude (\ref{itm:RQ3}).

\subsubsection{Recognising entity types from representations of quadruples}

\begin{figure}[hbt!]
\centering
\begin{subfigure}{.5\textwidth}
  \centering
  \includegraphics[width=\linewidth]{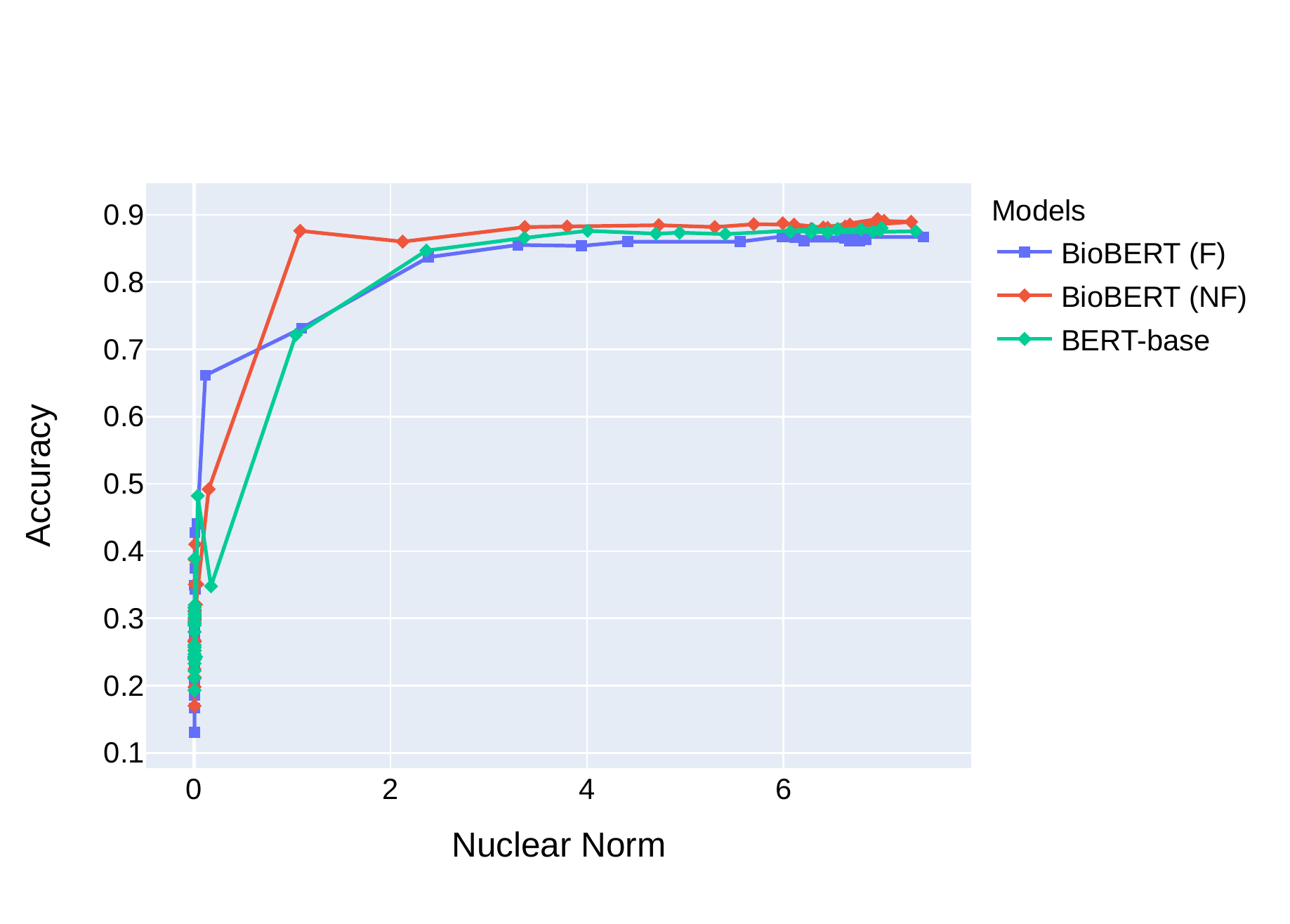}
  \caption{Accuracy VS Nuclear Norm (BioBERT)}
  \label{fig:probing_results:task2:acc_biobert}
\end{subfigure}%
\begin{subfigure}{.5\textwidth}
  \centering
  \includegraphics[width=\linewidth]{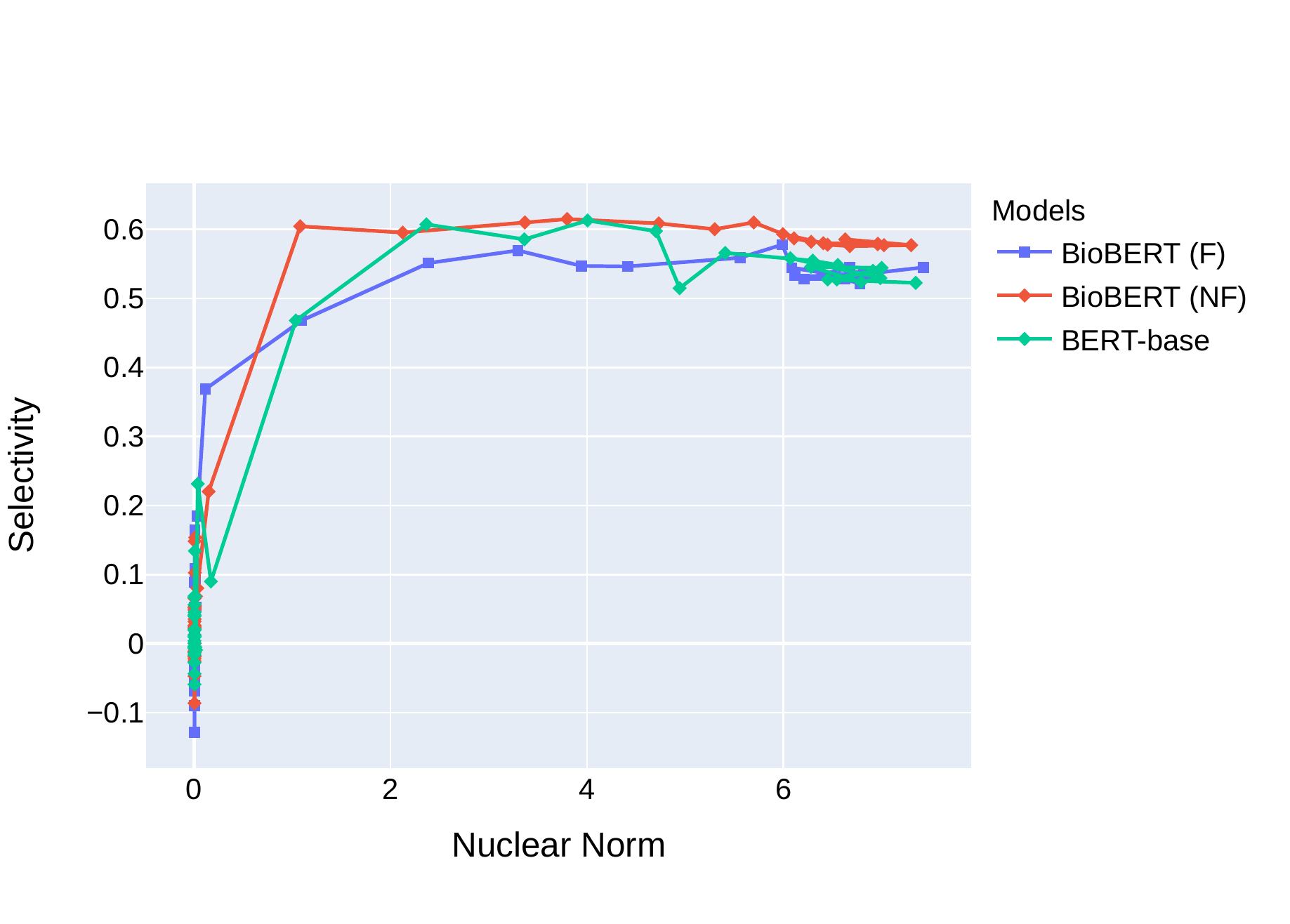}
  \caption{Selectivity VS Nuclear Norm (BioBERT)}
  \label{fig:probing_results:task2:sel_biobert}
\end{subfigure}
\\
\begin{subfigure}{.5\textwidth}
  \centering
  \includegraphics[width=\linewidth]{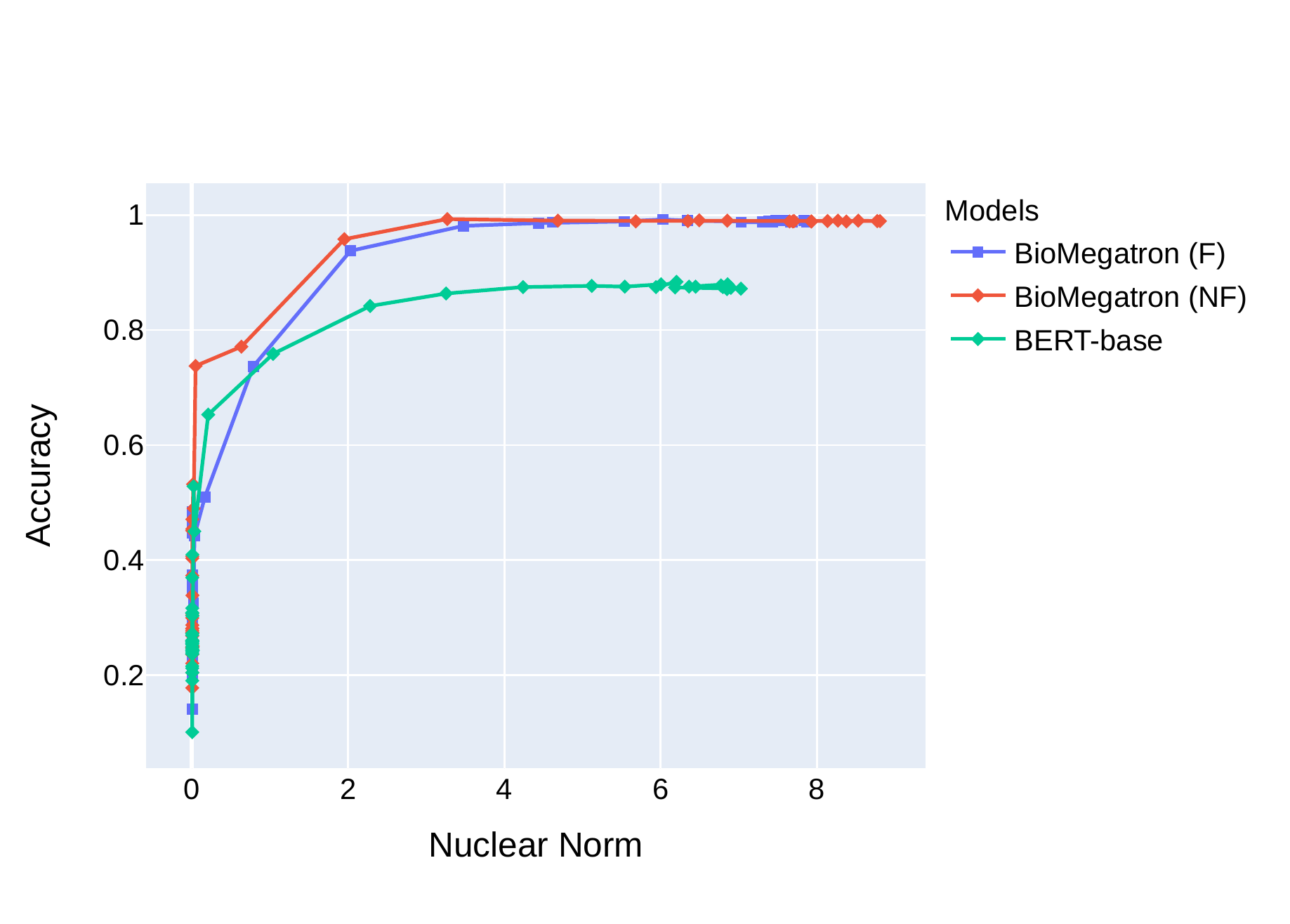}
  \caption{Accuracy VS Nuclear Norm (BioMegatron)}
  \label{fig:probing_results:task2:acc_biomegatron}
\end{subfigure}%
\begin{subfigure}{.5\textwidth}
  \centering
  \includegraphics[width=\linewidth]{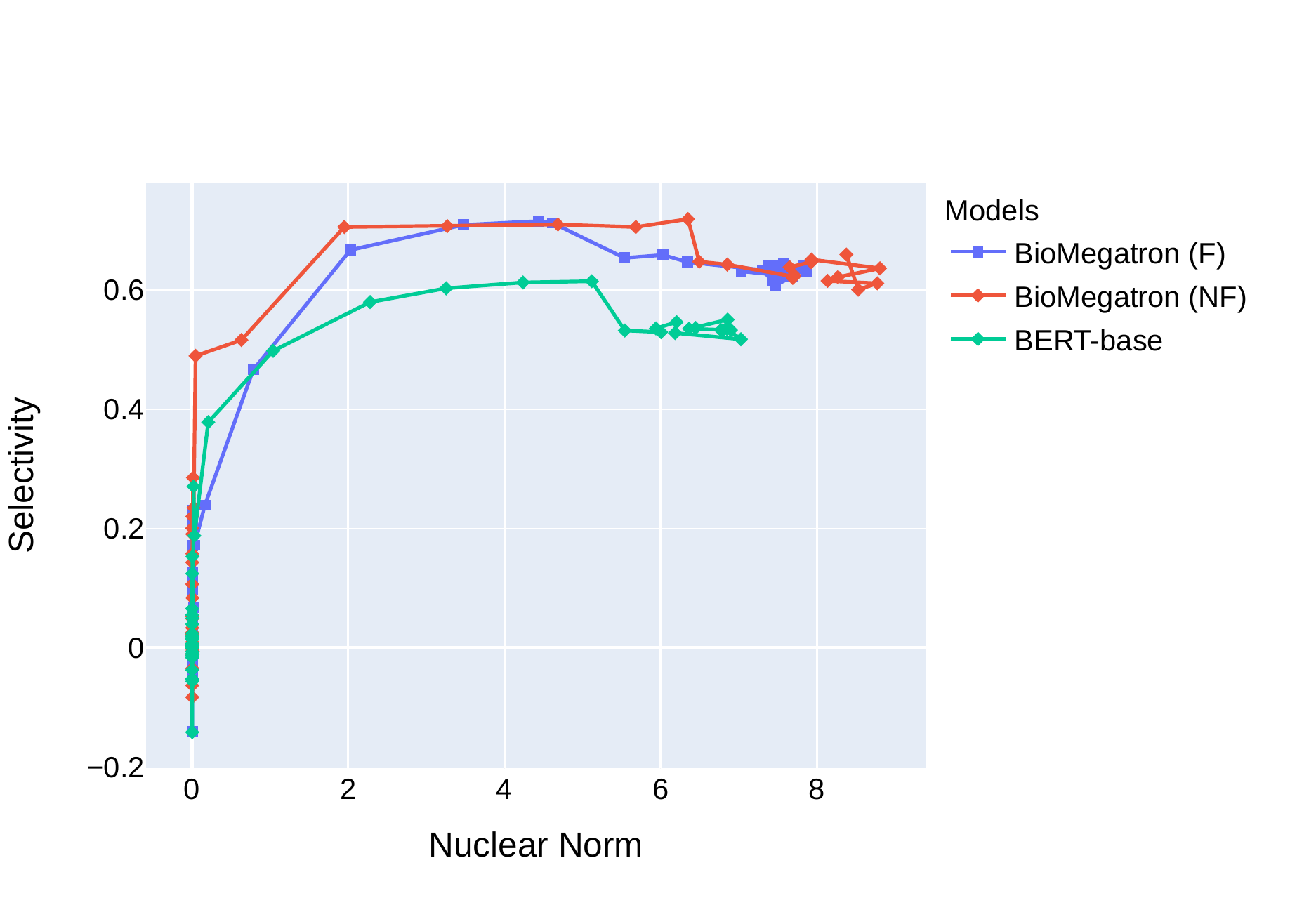}
  \caption{Selectivity VS Nuclear Norm (BioMegatron)}
  \label{fig:probing_results:task2:sel_biomegatron}
\end{subfigure}
\caption{Probing results for models fine-tuned on Task 2, following the same experiment design as Task 1. With fine-tuned (F) and non fine-tuned (NF) models}
\label{fig:probing_results:task2}
\end{figure}

Figure~\ref{fig:probing_results:task2} presents the probing results for Task 2, following the same task design as Task 1. Similar to task 1, selectivity was greater than zero for a control task containing random labels, and BERT-base and BioBERT both had higher accuracy compared with fine-tuned BioBERT. For this task, we can  observe minimal difference between the performance of the fine-tuned and non fine-tuned versions of BioMegatron, which outperform BERT and BioBERT models. For probes with a lower value for  their nuclear norm (i.e., less complex probes), the performance of the original model is slightly better. However, the difference is non-existent for more complex probes. 

 Probing results suggest that when fine-tuned for encoding complex n-ary relations (in Task 2) BioMegatron preserves more semantic information about entity type in the top layer that BioBERT (\ref{itm:RQ3}), as the difference in selectivity between fine-tuned (F) and non fine-tuned (NF) versions is smaller (Fig.\ref{fig:probing_results:task2}). Both BioBERT and BioMegatron achieve acceptable selectivity (both F and NF) suggesting that they do encode semantic domain knowledge at entity level (\ref{itm:RQ1}).

\subsection{How much biological knowledge do transformers embed?}

\subsubsection{Biologically relevant clusters in representations of pairs}
\begin{figure}[htp!]
\centering
\begin{subfigure}{.9\textwidth}
  \centering
  \includegraphics[width=\linewidth]{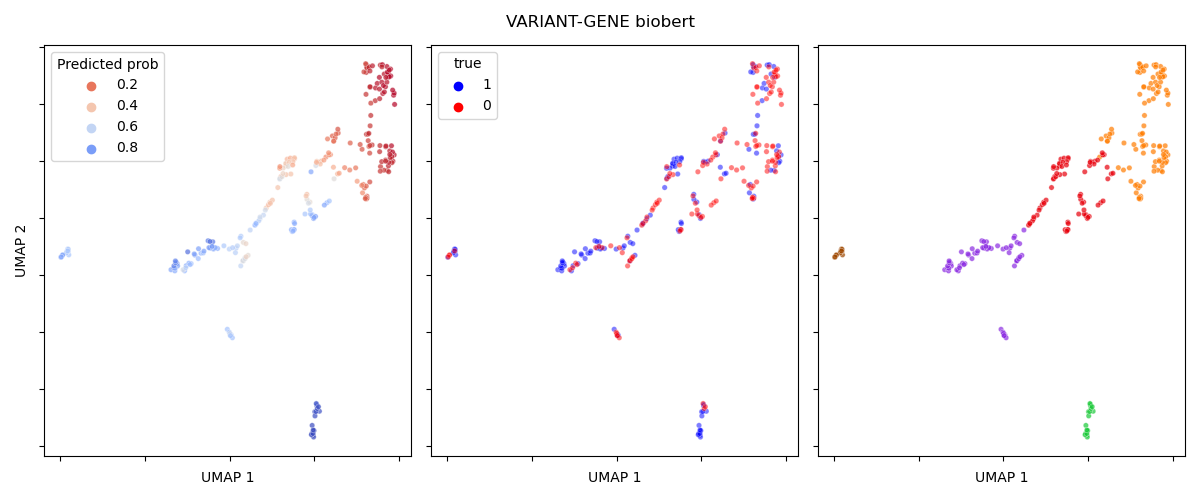}
  \caption{UMAP 2-dimensional.}
  \label{fig:umap_variant_gene_biobert}
\end{subfigure}%
\\
\begin{subfigure}{.9\textwidth}
  \centering
  \includegraphics[width=\linewidth]{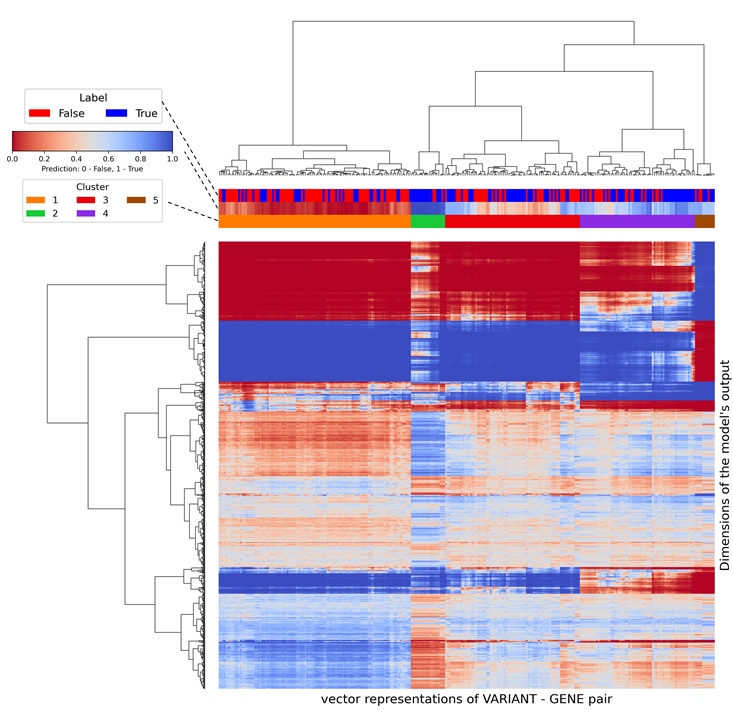}
  \caption{Clustermap based on Hierarchical Agglomerative Clustering.}
  \label{fig:clustermap_variant_gene_biobert}
\end{subfigure}
\caption{Representations of BioBERT output for variant-gene pairs in the balanced test set.}
\end{figure}

Based on clustering of BioBERT representations of variant-gene pairs in the balanced test set, and visual inspection of the clustermap and dendrogram, a cut point was applied that resulted in 5 clusters (Fig.\ref{fig:clustermap_variant_gene_biobert}).

The dendrogram shows that cluster 5 (brown) contained 11 gene-variant pairs and remained separated from the other pairs until late in the merging process. The gene-variant pairs in this cluster involved only the PIK3CA and ERBB3 genes, and these genes did not occur in any other clusters. BioBERT classified all these pairs as true, with probability $>$0.60, although 4/11 pairs were false (Supp Table \ref{tab:biobert_cluster5}). Interestingly, these genes participate in the same signaling pathways, including PI3K/AKT/mTOR.

Cluster 2 (green) contained 19 gene-variant pairs; 14/19 variants in this cluster represented gene fusions, denoted by the notation \textit{gene name - gene name}. All pairs were assigned as true, with probability $>$0.96, although 3/19 pairs were false (Supp Table \ref{tab:gene_variant_pairs_cluster21}).

\begin{figure}[htp!]
\centering
\begin{subfigure}{.9\textwidth}
  \centering
  \includegraphics[width=\linewidth]{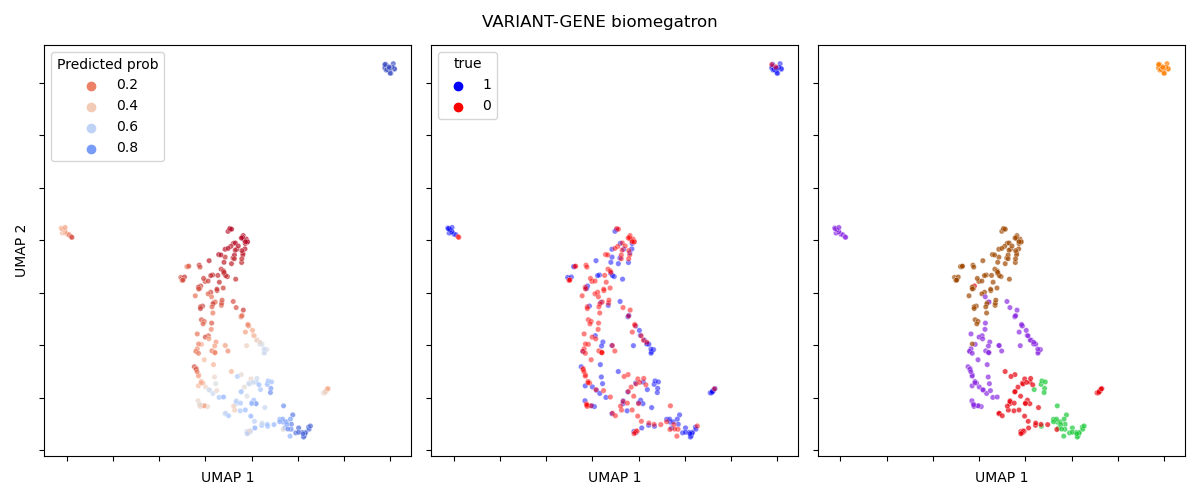}
  \caption{UMAP 2-dimensional.}
  \label{fig:umap_variant_gene_biomegatron}
\end{subfigure}%
\\
\begin{subfigure}{.9\textwidth}
  \centering
  \includegraphics[width=\linewidth]{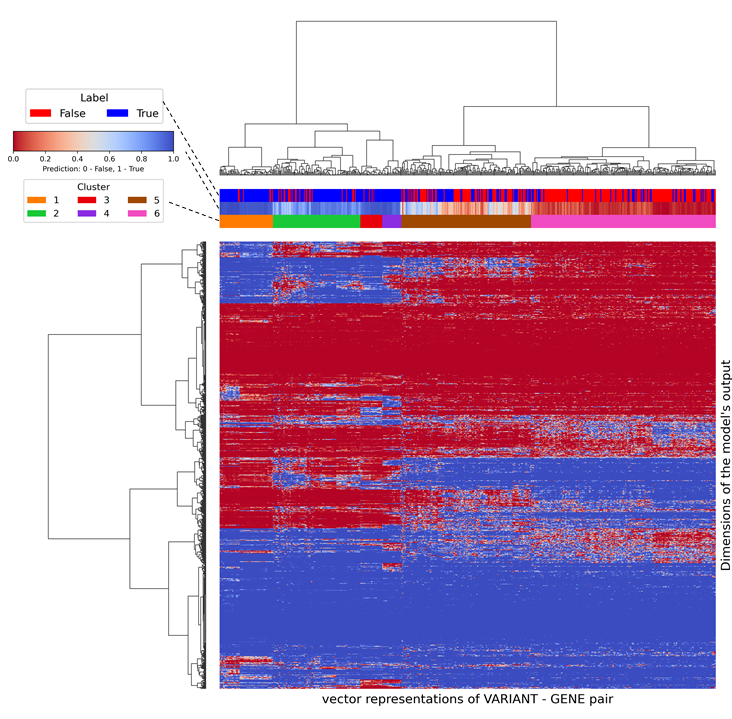}
  \caption{Clustermap based on Hierarchical Agglomerative Clustering.}
  \label{fig:clustermap_variant_gene_biomegatron}
\end{subfigure}
\caption{Representations of BioMegatron output for variant-gene pairs in the balanced test set.}
\label{fig:umap_and_clustermap_variant_gene_biomegatron}
\end{figure}

Following the clustering of BioMegatron representations on variant-gene pairs in the balanced test set, a cut point was applied that resulted in 6 clusters (Fig.\ref{fig:umap_and_clustermap_variant_gene_biomegatron}).

BioMegatron cluster 1 contained 16 of the 19 gene-variant pairs found in BioBERT cluster 2 (Supp Table \ref{tab:gene_variant_pairs_cluster21}) As observed for BioBERT,  BioMegatron  determined all these pairs to be true with high confidence (probability $>$0.96). 
 
Clustering analysis reveals an evident dataset artefact, i.e. gene fusions as \textit{gene name - gene name} which is reflected in the representation. Both models encoded these fusions in a significantly different way compared to other pairs.


\subsubsection{Biologically relevant clusters in representations of clinical relations}

\begin{figure}[htp!]
\centering
\begin{subfigure}{.9\textwidth}
  \centering
  \includegraphics[width=\linewidth]{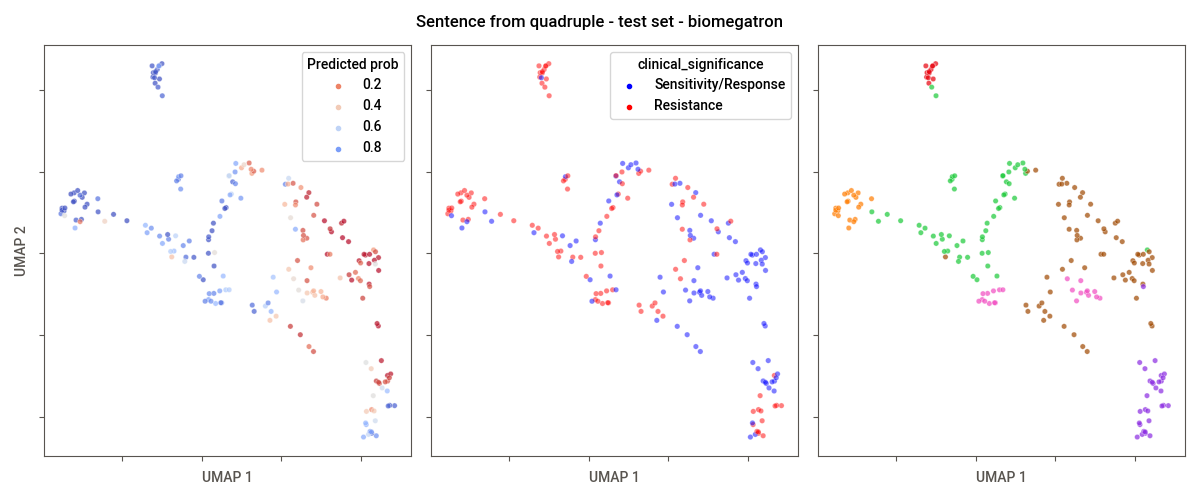}
  \caption{UMAP 2-dimensional.}
  \label{fig:umap_quads_biomegatron}
\end{subfigure}%
\\
\begin{subfigure}{.9\textwidth}
  \centering
  \includegraphics[width=\linewidth]{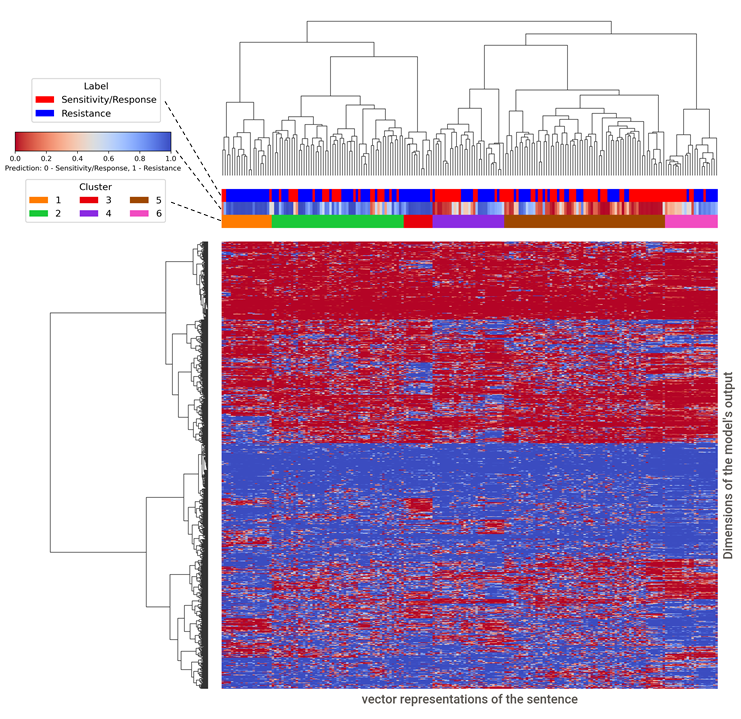}
  \caption{Clustermap based on Hierarchical Agglomerative Clustering.}
  \label{fig:clustermap_quads_biomegatron}
\end{subfigure}
\caption{Representations of BioMegatron output for quadruples in the balanced test set.}
\label{fig:umap_and_clustermap_quads_biomegatron}
\end{figure}

Following clustering of BioMegatron representations of quadruples, a cut-off point was applied that resulted in 6 clusters (Fig.\ref{fig:umap_and_clustermap_quads_biomegatron}).

\begin{itemize}
\item Cluster 1 included 21 quadruples, all of which related to colorectal cancer. Most quadruples involved either BRAF, EGFR or KRAS genes.  

\item Cluster 3 included 11 quadruples, all of which related to the drug vemurafenib. Most (9/11) related to melanoma, and 10/11 were associated with resistance.  

\item Cluster 4 included 30 quadruples, all of which related to the KIT gene, Gastrointestinal Stromal Tumor and either sunatinib or imatinib drugs; KIT was not associated with any other clusters.  

\item Cluster 6 included 22 quadruples, all of which related to the ABL gene and fusions with the BCR gene (denoted by \textit{Variant BRCA-ABL})  
\end{itemize}


\begin{figure}[htp!]
\centering
\begin{subfigure}{.9\textwidth}
  \centering
  \includegraphics[width=\linewidth]{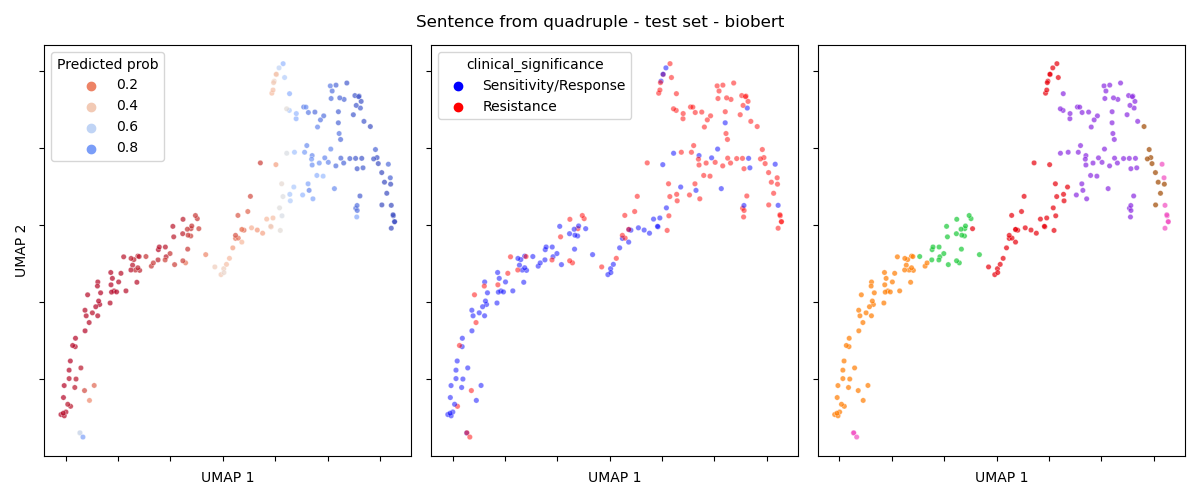}
  \caption{UMAP 2-dimensional.}
  \label{fig:umap_quads_biobert}
\end{subfigure}%
\\
\begin{subfigure}{.9\textwidth}
  \centering
  \includegraphics[width=\linewidth]{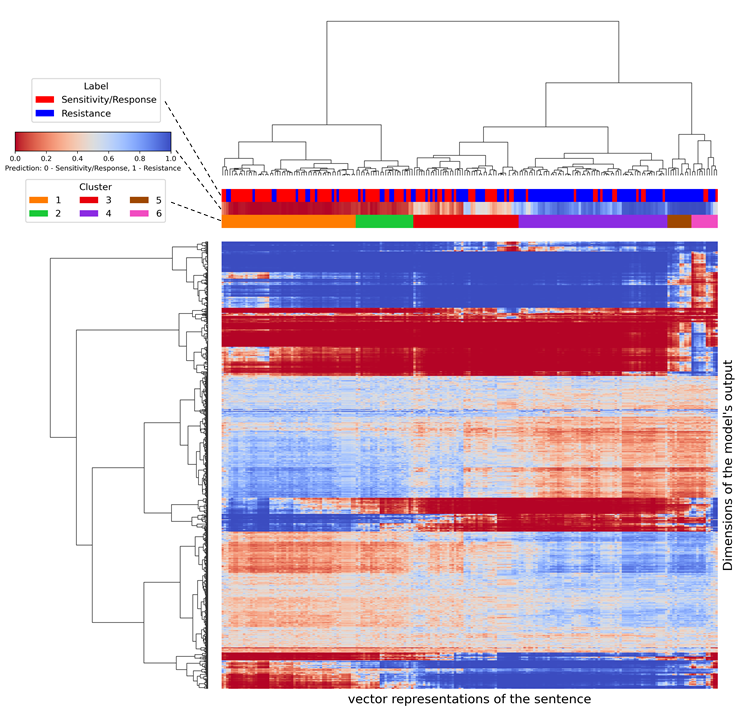}
  \caption{Clustermap based on Hierarchical Agglomerative Clustering.}
  \label{fig:clustermap_quads_biobert}
\end{subfigure}
\caption{Representations of BioBERT output for quadruples in the balanced test set.}
\label{fig:umap_and_clustermap_quads_biobert}
\end{figure}
 
Similarly, 6 clusters were defined based for the BioBERT representations (Fig.\ref{fig:umap_and_clustermap_quads_biobert}). Quadruples in BioBERT clusters were less homogeneous compared with those for the BioMegatron clusters. Two small clusters 5 and 6 are described in Supp Table \ref{tab:biobert_cluster56}. Cluster 5 included 10 quadruples, involving 7 different genes, 7 diseases and 6 drugs; cluster 6 included 11 quadruples, with 4 genes, 5 diseases and 11 drugs; no clear pattern was evident in either cluster. 
 
Clustering analysis reveals that representations encoded by fine-tuned BioMegatron form biologically meaningful clusters, in terms of gene-variant-disease-drug (\ref{itm:RQ2}). For BioBERT, the patterns are less apparent and may require deeper, more granular investigation (\ref{itm:RQ3}).

\subsubsection{Entity types  clusters in fine-tuned models}

In this section, we investigated the  clustering of the latent vectors. These vectors were also used form the probing task. Each vector represents one entity contextualised inside sentences from the test set (From both Task 1 and Task 2, more in Supp Methods \ref{supp_clustering_probing}).

\begin{figure}[htp!]
\centering
\begin{subfigure}{0.85\textwidth}
  \centering
 
\includegraphics[width= 1\textwidth]{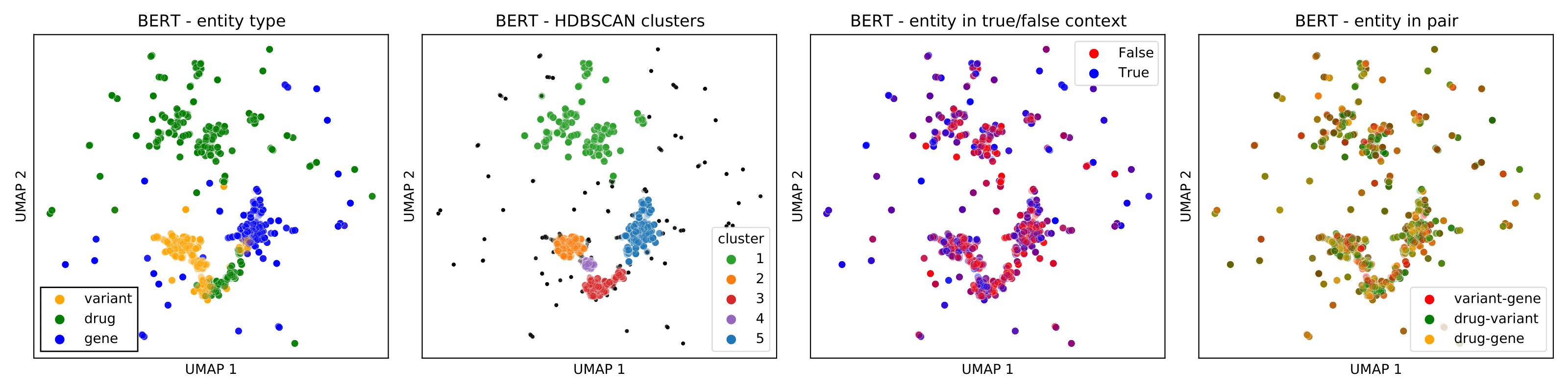}
\caption{BERT}
\label{fig:probing_umap_bert_pairs}
\end{subfigure}

\begin{subfigure}{0.85\textwidth}
\centering
\includegraphics[width= 1\textwidth]{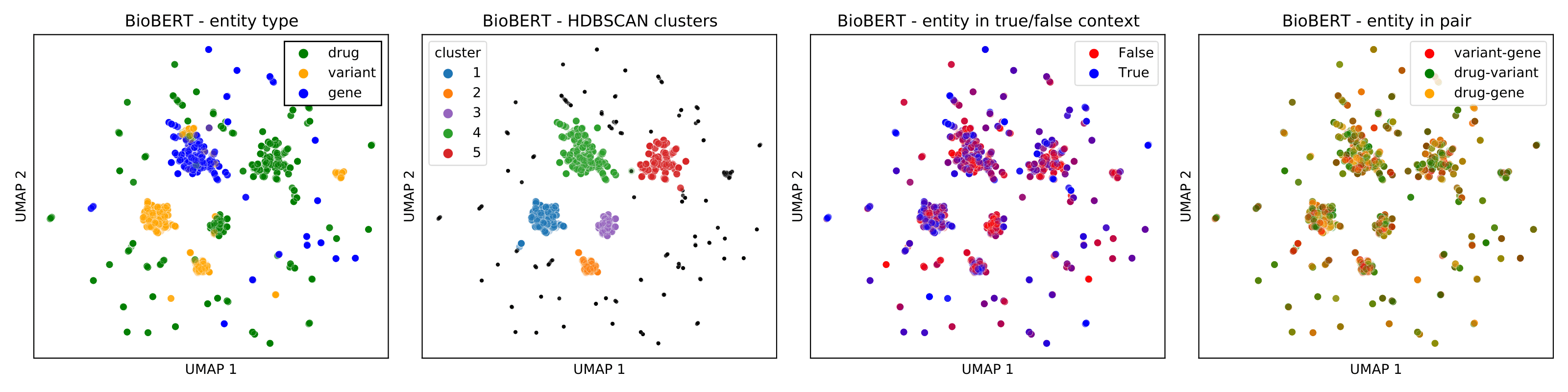}
\caption{BioBERT}
\label{fig:probing_umap_biobert_pairs}
\end{subfigure}

\begin{subfigure}{0.85\textwidth}
\centering
\includegraphics[width= 1\textwidth]{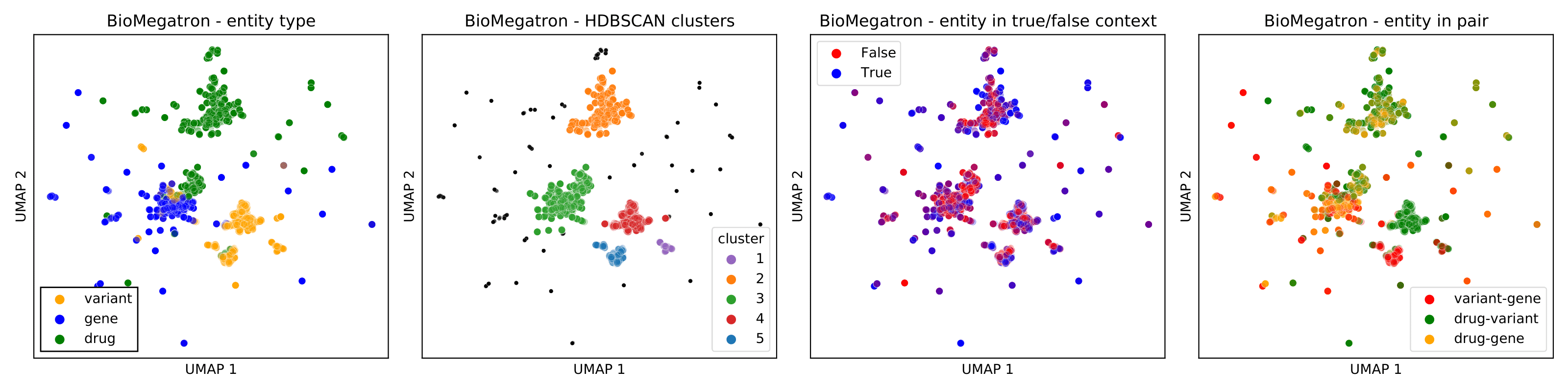}
\caption{BioMegatron}
\label{fig:probing_umap_biomegatron_pairs}
\end{subfigure}

\begin{subfigure}{0.85\textwidth}
\centering
\includegraphics[width= 1\textwidth]{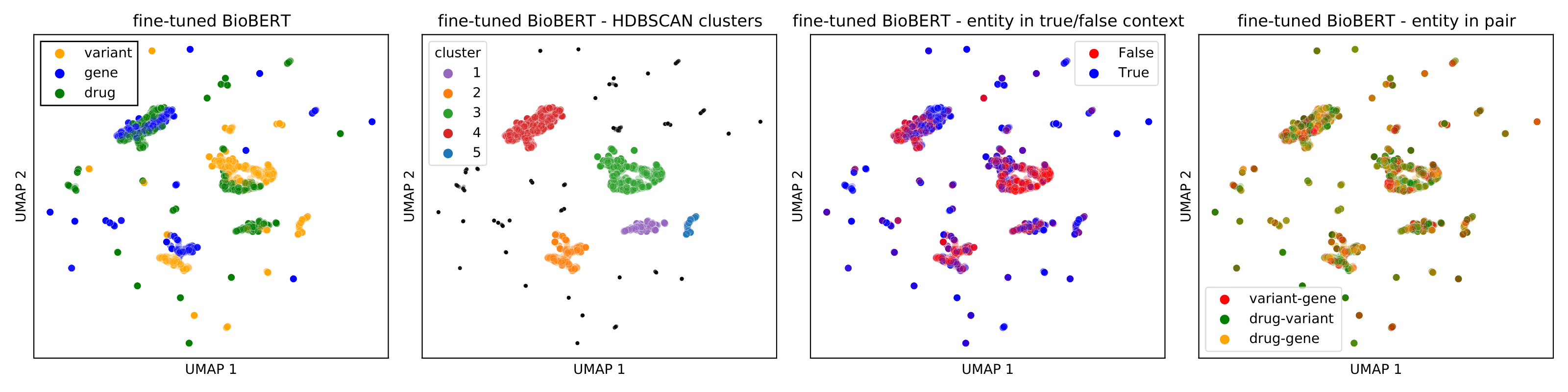}
\caption{fine-tuned BioBERT}
\label{fig:probing_umap_fine_tuned_biobert_pairs}
\end{subfigure}

\begin{subfigure}{0.85\textwidth}
\centering
\includegraphics[width= 1\textwidth]{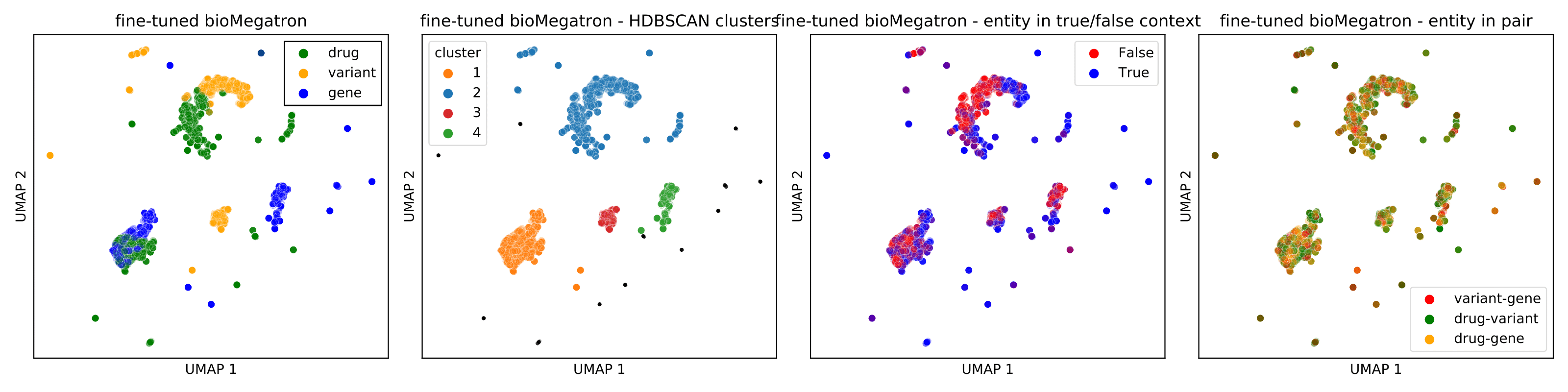}
\caption{fine-tuned BioMegatron}
\label{fig:probing_umap_fine_tuned_biomegatron_pairs}
\end{subfigure}
\caption{UMAP representation of entities from Task 1 used as input to Probing. In BERT, BioBERT and BioMegatron the clusters are homogeneous regarding the entity type (left plots). Fine-tuned models lose this property.}
\label{fig:probing_umap_pairs}
\end{figure}

\begin{figure}[htp!]
\centering
\begin{subfigure}{.68\textwidth}
  \centering

\includegraphics[width= 0.95\textwidth]{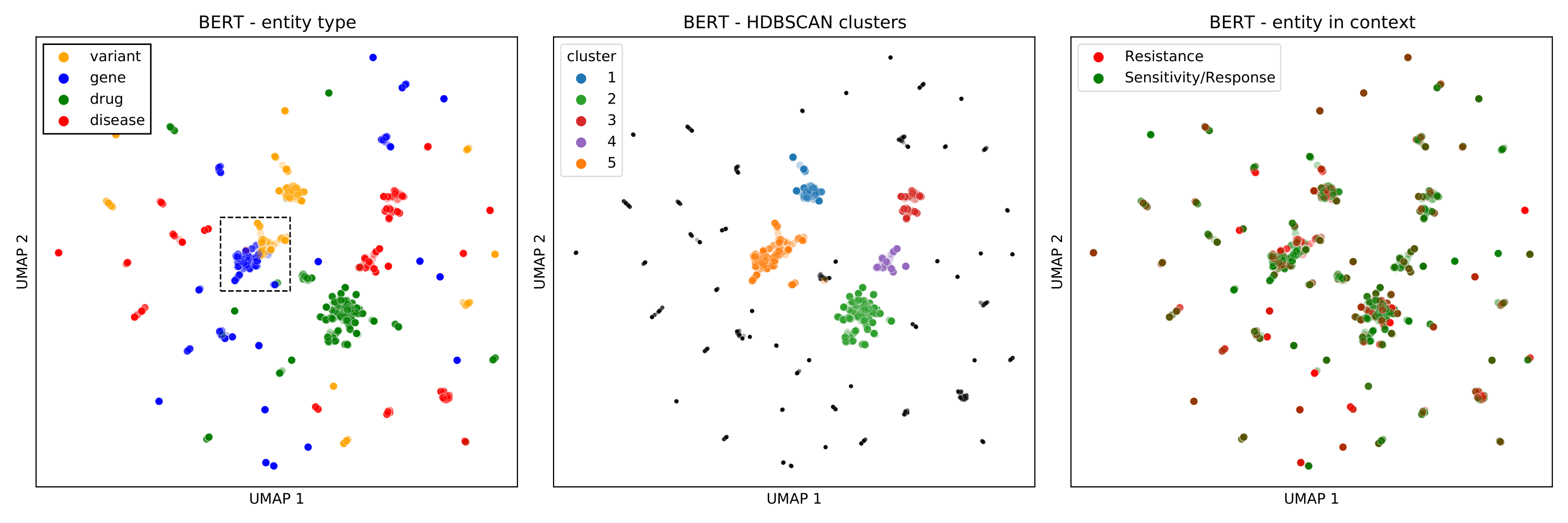}
\caption{BERT}
\label{fig:probing_umap_bert}
\end{subfigure}

\begin{subfigure}{.68\textwidth}
\centering
\includegraphics[width= 0.95\textwidth]{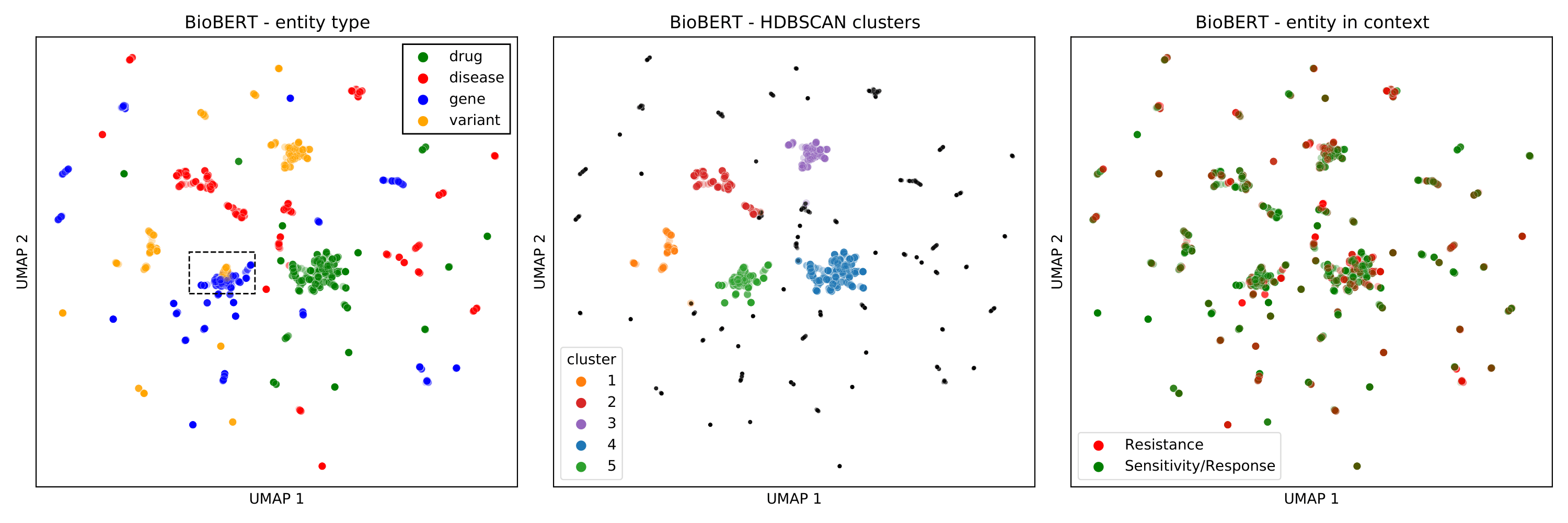}
\caption{BioBERT}
\label{fig:probing_umap_biobert}
\end{subfigure}

\begin{subfigure}{.68\textwidth}
\centering
\includegraphics[width= 0.95\textwidth]{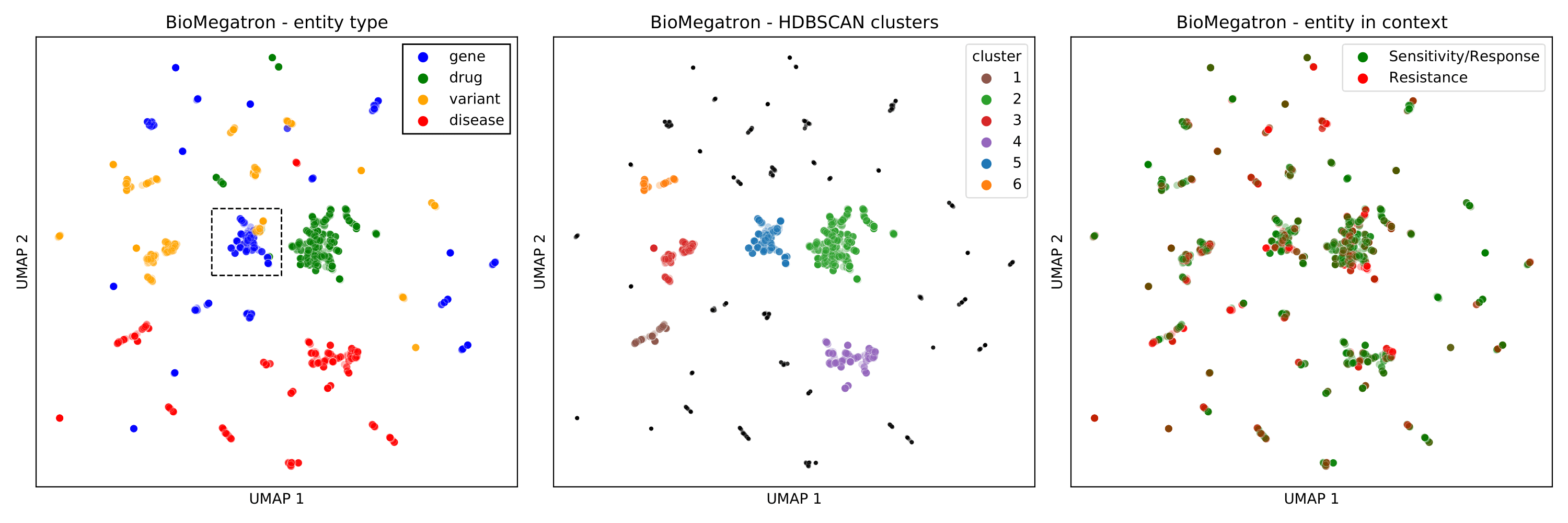}
\caption{BioMegatron}
\label{fig:probing_umap_biomegatron}
\end{subfigure}

\begin{subfigure}{.68\textwidth}
\centering
\includegraphics[width= 0.95\textwidth]{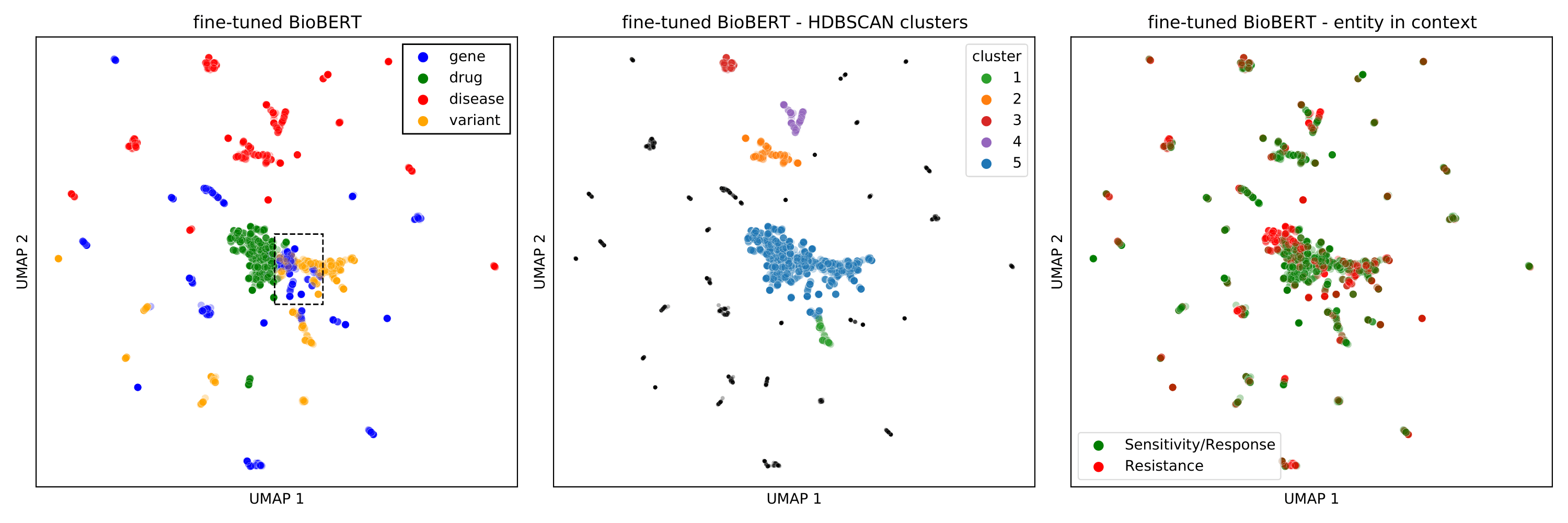}
\caption{fine-tuned BioBERT}
\label{fig:probing_umap_fine_tuned_biobert}
\end{subfigure}

\begin{subfigure}{.68\textwidth}
\centering
\includegraphics[width= 0.95\textwidth]{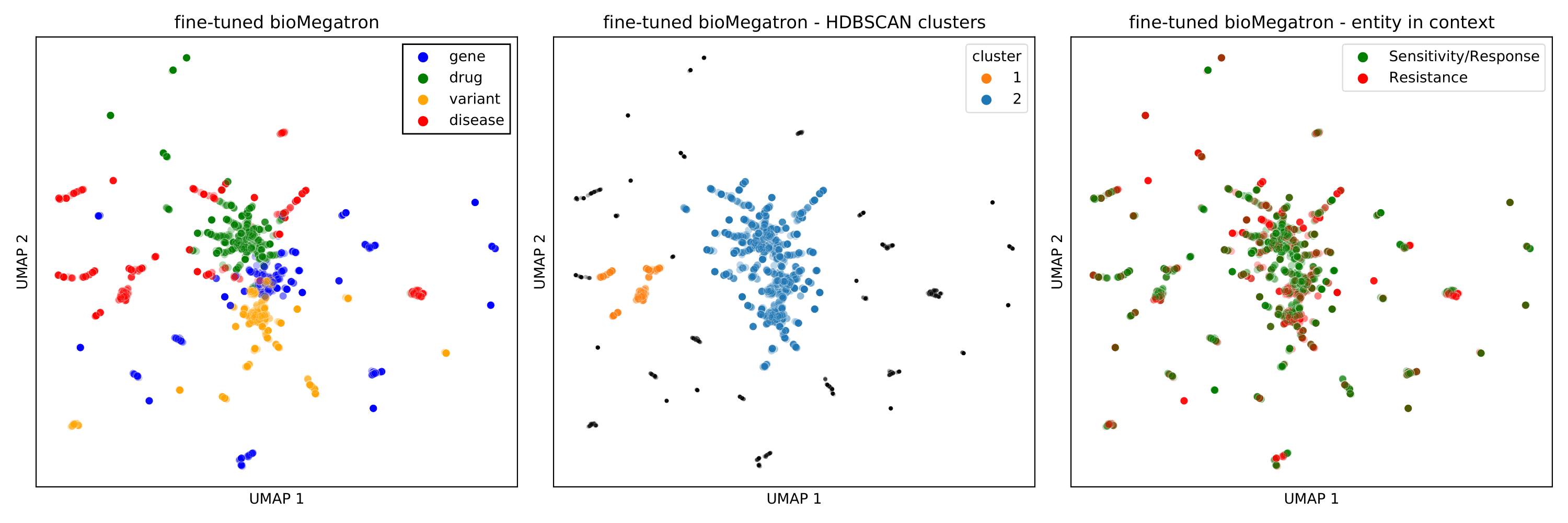}
\caption{fine-tuned BioMegatron}
\label{fig:probing_umap_fine_tuned_biomegatron}
\end{subfigure}
\caption{UMAP representation of entities from Task 2: (left) entity types; (center) clusters from HDBSCAN; (right) target label in classification task. Dashed box corresponds to entities from quadruples, in which variant entity contains the gene entity name. Representations from non fine-tuned models form more distinctive clusters, more homogeneous in terms of entity type.}
\label{fig:umap_task_2_probing}
\end{figure}

Results from HDBSCAN evaluation of UMAP representations are summarized in Table \ref{tab:probing_clusters_homogeneity} and Fig.\ref{fig:probing_umap_biomegatron_pairs},\ref{fig:probing_umap_fine_tuned_biomegatron}.

For Task 1, the non-fine-tuned transformer models clustered entities according to their type - the average homogeneity of clusters was 0.940 for BioBERT, 0.911 for BioMegatron and 0.883 for BERT. In contrast, clusters generated by the fine-tuned transformer models were less homogeneous (0.758 and 0.726 for BioMegatron and BioBERT, respectively) - this was observed across all types of entity-pairs. 

For Task 2, clusters generated by the non-fine-tuned models were almost perfectly homogeneous (homogeneity $>$98.8\%), except for cluster 5 consisting of both gene and variant entities (black dashed box in Fig\ref{fig:umap_task_2_probing}). 

However, for fine-tuned models, the majority of  entities get projected closely under a 2D UMAP projection, similarly to the findings in \cite{rajaeeHowDoesFinetuning2021a, durraniHowTransferLearning2021}. In fine-tuned BioBERT, drugs  are projected to variants and some genes. As a result, a large cluster (5) with mixed entity types emerges.  A similar type of clustering behaviour is observed in the fine-tuned BioMegatron, showing one large cluster (2) containing portions of all types of entities.

\begin{table}[hbt!]
\centering
\caption{Mean homogeneity in clusters.}
\label{tab:probing_clusters_homogeneity}
\resizebox{\textwidth}{!}{%
\begin{tabular}{@{}llccc@{}}
\toprule
\multicolumn{1}{c}{\textbf{Task}} &
  \multicolumn{1}{c}{\textbf{Model}} &
  \textbf{\begin{tabular}[c]{@{}c@{}}Entity type \\ (gene, variant, drug)\end{tabular}} &
  \textbf{\begin{tabular}[c]{@{}c@{}}Target label Pair type\\  (True or False)\end{tabular}} &
  \textbf{\begin{tabular}[c]{@{}c@{}}Pair Type\\ (d-g, g-v,d-v)\end{tabular}} \\ \midrule
\textbf{Task 1} &
  \textbf{BERT} &
  0.883 &
  0.553 &
  0.471 \\
\textbf{} &
  \textbf{BioBERT} &
  0.940 &
  0.572 &
  0.478 \\
\textbf{} &
  \textbf{BioMegatron} &
  0.911 &
  0.548 &
  0.708 \\
\textbf{} &
  \textbf{FT BioBERT} &
  0.726 &
  0.638 &
  0.488 \\
\textbf{} &
  \textbf{FT BioMegatron} &
  0.758 &
  0.538 &
  0.474 \\ \midrule
 &
   &
  \textbf{gene, variant, drug, disease} &
  \multicolumn{2}{c}{\textbf{Sensitivity/Response or Resistance}} \\ \midrule
\textbf{Task 2} &
  \textbf{BERT} &
  \multicolumn{1}{r}{.996; .599 in \#5 (genes and variants)} &
  \multicolumn{2}{c}{0.695} \\
\textbf{} &
  \textbf{BioBERT} &
  \multicolumn{1}{r}{.998; ; .793 in \#5 (genes and variants)} &
  \multicolumn{2}{c}{0.679} \\
\textbf{} &
  \textbf{BioMegatron} &
  \multicolumn{1}{r}{1.0; .773 in \#5 (genes and variants)} &
  \multicolumn{2}{c}{0.656} \\
\textbf{} &
  \textbf{FT BioBERT} &
  \multicolumn{1}{r}{.990; ; .514 in \#5 (drugs, variants,genes)} &
  \multicolumn{2}{c}{0.680} \\
\textbf{} &
  \textbf{FT BioMegatron} &
  \multicolumn{1}{r}{.380 in large cluster \#2} &
  \multicolumn{2}{c}{0.691} \\ \bottomrule
\end{tabular}
}
\end{table}

In all the 5 models, the representations do not group according to target labels in Task 1 nor Task 2. Homogeneity of clusters regarding true/false labels equals on average .570, and regarding 'Sensitivity/Response'/'Resistance' .680. They are close to a random distribution of labels over clusters, because labels proportion are 0.50 and 0.65, respectively. 
 
Clustering analysis and homogeneity evaluation confirm that both BioBERT and BioMegatron encode fundamental semantic knowledge at entity level, in this case genes, variants, drugs and diseases. However, a significant part of the latent semantics is changed during fine-tuning, which is particularly apparent for a more complex Task 2 (\ref{itm:RQ1}).

\section{Discussion}

\subsection{Summary of main findings}

In this study we performed a detailed analysis on the embeddings of biological knowledge in transformer-based neuro-language models using a cancer genomics knowledge base. 

First, we compared the performance between biomedical fine-tuned transformers (BioBERT and BioMegatron) and a naive simple classifier (KNN) for two specific classification tasks. Specifically, these tasks aimed to determine whether each transformer model captures biological  knowledge about: pairwise associations between genes, variants, drugs and diseases (Task 1), and the clinical significance of relationships between gene variants, drugs and diseases (Task 2). 

The hypothesis under test was that transformers would show better performance compared with a naive classifier , eliciting the role of the pre-trained component of the model (\ref{itm:RQ4}). Results for both tasks support this hypothesis. For Task 1, both BioBERT and BioMegatron outperformed the naive classifier for distinguishing true versus false associations between pairs of biological entities. Similarly, for Task 2, both transformer models outperformed the naive classifier for predicting the clinical significance of quadruples of entities. For Task 2, the transformer models achieved an acceptable performance (AUC $>$ 0.8), although performance in Task 1 was lower (AUC approx. 0.6).  

We highlighted the need for  addressing the role of dataset imbalance within the assessment of embeddings (\ref{itm:RQ4}). Specifically, in our analysis, we found significant differences between AUCs for the imbalanced and balanced test sets. Furthermore, we found significant correlations between the classification error and imbalance for individual entities. Similarly, the error is associated with  a co-occurrence bias (within the corpus based on the biomedical literature): i.e. in Task 1: a true pair which occurs in the literature multiple times is more likely to be classified as true, compared to pairs that occur less frequently.

Second, we used probing methods to inspect the consistency of the representation for each type of biological entity, and we compared pre-trained versus fine-tuned models  (\ref{itm:RQ1},\ref{itm:RQ2}). More specifically, we determined the performance of each model in classifying the type (gene, variant, drug or disease) of entities based on their representation in the model via accuracy and selectivity. We quantified how much semantic structure is lost in fine-tuning, and how biologically meaningful is the remaining. For BioBERT, both accuracy and selectivity were lower for the fine-tuned models compared with the base models, including BERT-base which is not specific for the medical/biological domain. For BioMegatron, there was  only a slight difference in performance between the fine-tuned and non-fine-tuned models.  Probing experiments demonstrated that fine-tuned BioMegatron better preserves the pre-trained knowledge when compared to fine-tuned BioBERT (\ref{itm:RQ3}). 
 
Finally, we provide a qualitative and quantitative analysis of clustering patterns of the embeddings, using UMAP, HDBCAN and hierarchical agglomoerative clustering. We show that entities of the same type cluster together, and that this is more pronounced for the non-fine-tuned models compared with the fine-tuned models  (\ref{itm:RQ1},\ref{itm:RQ2}). A cluster analysis revealed  biological meaning. For instance, we found a cluster with vast majority of sentences related to resistant response to vemurafenib in melanoma treatment. Another example: a cluster specific to KIT gene, Gastrointestinal stromal tumor (GIST), Sunatinib and Imatinib. According to domain-expert knowledge the Imatinib, a KIT inhibitor, is a the standard first-line treatment for metastatic GIST, whereas sunatinib is the second option.

\subsection{Strengths and limitations}

Strengths: 

\begin{itemize}

\item We have used the CIViC database as the basis of our analysis. We consider this to be a high-quality dataset, because:  (i) it entails a set of relationships curated by domain experts;  (ii) most relationships include a confidence score;  (iii) it has been developed for a closely-related use case, namely to support clinicians in evaluation of clinical significance of variants. 

\item We employ state-of-the-art, bidirectional transformer models trained on a biomedical text corpus (PubMed abstracts) containing over 29M articles and 4.5B words. 


\item Patterns in representations are investigated using 2 methods (UMAP and HAC), instead of relying  on a signle method. Clusters are thoroughly described and quantified using homogeneity  metrics. 

\item We include input from domain experts in data preparation, evaluation and interpretation of results. It allows for: (i) the correct filtering of evidence; (ii) assessment of the relevance of investigated biomedical relations; (iii) granular analysis of clusters in search for biological meaning.  

\end{itemize}

Limitations: 
\begin{itemize}

\item The distribution of entities among the dataset has the potential to lead to overfitting. For example, if the EGFR gene is over-represented among true gene-drug pairs compared with other genes, a model could classify gene-drug pairs solely on the whether gene = EGFR and perform better than expected. Indeed, the distributions of entities in our dataset were highly right-skewed (Pareto distribution). This issue refers to the well-known imbalance problem, which leads to an incorrect performance evaluation.
Although we applied a balancing procedure, it is infeasible to create perfectly balanced dataset.


\item In CIViC, drug interaction types can be either \textit{combination}, \textit{sequential}, or \textit{substitutes}. In the generation of evidence sentences, we did not account for that variation, which for sentences with multiple drugs it may slightly alter the representation of clinical significance in the model. 

\item In CIViC, there are evidence items that claim contradicting clinical significance for the same relation. We excluded them from our dataset, however their future investigation would be of relevance.  

\end{itemize}


\subsection{Related work}
\subsubsection{Supporting somatic variant interpretation in cancer}

There is a critical need to evaluate the large amount of relevant variant data generated by tumor Next Generation Sequencing (NGS) analyses, which predominantly have unknown significance and complicates the interpretation of the variants \cite{Good2014}. One of the ways to streamline and standardise cancer curation data in electronic medical records is to use the web resources from the CIViC curatorial platform \cite{danos2018}. An open source and open access CIViC database, built on community input with peer-reviewed interpretations, already proved  to be useful for this purpose \cite{barnell2019}. The authors used the database to develop the Open-sourced CIViC Annotation Pipeline (OpenCAP), providing methods for capturing variants and subsequently providing tools for variant annotation. It supports scientists and clinicians who use precision oncology to guide patients' treatment. In addition, \cite{Danos2019StandardOP} described improvements at CIViC that include common data models and standard operating procedures for variant curation. These are to support a consistent and accurate interpretation of cancer variants.

Clinical interpretation of genomic cancer variants requires highly efficient interoperability tools. Evidence and clinical significance of the CIViC database was used in a novel genome variation annotation, analysis and interpretation platform TGex (the Translational Genomics expert) \cite{Dahary2019GenomeAA}. By providing access to a comprehensive knowledge base of genomic annotations, TGex tool simplifies and speeds up the interpretation of variants in clinical genetics processes. Furthermore, \cite{doi:10.1200/CCI.19.00127} provided CIViCpy, an open-source software for extracting and inspection of records from the CIViC database. The delivery of CIViCpy enables the creation of downstream applications and the integration of CIViC into clinical annotation pipelines.

\subsubsection{Text-mining approaches using CIViC}
The development of guidelines \cite{liStandardsGuidelinesInterpretation2017} for the interpretation of somatic variants, which include complexity of multiple dimensions of clinical relevance, allow for a better standardization of the assessment of cancer variants in the oncological community. In addition, they can enhance the rapidly growing use of genetic testing in cancer, the results of which are critical to accurate prognosis and treatment guidance. Based on the guidelines, \cite{heVariantInterpretationCancer2019} demonstrated computational approaches to take pre-annotated files and to apply criteria for the assessment of the clinical impact of somatic variants. In turn, \cite{Lever500686} proposed a text-mining approach to extract the data on thousands of clinically relevant biomarkers from the literature and using a supervised learning approach they constructed a publicly accessible knowledge base called CIViCmine. They extracted key parts of the evidence item, including: cancer type, gene, drug (where applicable), and the specific evidence type. The CIViCmine contains over 87K biomarkers associated with 8k genes, 337 drugs, and 572 cancer types, representing more than 25k abstracts and almost 40k full-text publications. This approach allowed counting the number of mentions of specific evidence items: cancer type, gene, drug (where applicable), and the specific evidence type in PubMed abstracts and PubMed Central Open Access full-text articles and comparing them with the CIViC knowledge base. A similar approach was previously proposed by \cite{2016PLSCB..12E5017S} who proposed a method to automate the extraction of disease-gene-variant triples from all abstracts in PubMed related to a set of ten important diseases.

\cite{seva-etal-2018-identifying} developed a  NLP pipeline for identifying the most informative key sentences in oncology abstracts by assessing the clinical relevance of sentences implicitly based on their similarity to the clinical evidence summaries in the CIViC database. They used two semi-supervised  methods: transductive learning from positive and unlabelled data (PU Learning) and Self-Training by using abstracts summarised in relevant sentences as unlabelled examples. \cite{DBLP:journals/corr/abs-1808-08485} developed deep probabilistic logic (DPL) as a general framework for indirect supervision, by combining probabilistic logic with deep learning. They used existing knowledge bases with hand-curated drug-gene-mutation facts: the Gene Drug Knowledge Database (GDKD) \cite{Dienstmann118} and CIViC, which together contained 231 drug-gene-mutation triples, with 76 drugs, 35 genes and 123 mutations. Recently, \cite{DBLP:journals/corr/abs-1904-02347} proposed a novel multiscale neural architecture for document- level n-ary relation extraction, which combines representations learned over various text spans throughout the document and across the subrelation hierarchy. For distant supervision, they used CIViC, GDKD \cite{Dienstmann118}, and OncoKB \cite{doi:10.1200/PO.17.00011} knowledge bases. 

This section summarized the usage of the CIViC database in the development of  NLP pipelines as well as approaches to using NLP with cancer related literature. However, we did not find any study which used cancer genomic databases (such as CIViC) to investigate the semantic characterisation of a biomedically trained neural language models. 

\subsection{ Model's bias caused by the unbalanced training set}
 Our findings regarding the bias in the models caused by the unbalanced dataset align with findings in the previous works. \cite{mccoyRightWrongReasons2019} show that NLI models rely on adopted heuristic from statistical regularities in training sets, which are valid for frequent cases, but invalid for less frequent ones. It results in low performance in HANS (Heuristic Analysis for NLI Systems), which is attributed to invalid heuristics rather than deeper understanding of language.
\cite{gehmanRealToxicityPromptsEvaluatingNeural2020} recommends a careful examination of the dataset due to possible toxic, biased, or otherwise degenerate behavior of language models.
Similarly, in \cite{nadeemStereoSetMeasuringStereotypical2021} a strong stereotypical bias was reported in pre-trained BERT, GPT2, ROBERTA, and XLNET. 
Distribution in the dataset affects the performance \cite{zhongFactualProbingMASK2021b}, leading to overestimation of model's inference and deeper understanding of language \cite{gururanganAnnotationArtifactsNatural2018, minSyntacticDataAugmentation2020}. In our study, we confirmed the importance of integrating a balancing strategy for embedding studies.

\subsection{ Evaluation of semantic knowledge in transformer-based models}
 Fine-tuning distorts the original distribution within pre-trained models: higher layers are more adjusted to the specific task and lower layers retain their representation \cite{durraniHowTransferLearning2021,merchantWhatHappensBERT2020}. Although fine-tuning affects top layers, it is interpreted to be a conservative process and there is no catastrophic forgetting of information in the entire model \cite{merchantWhatHappensBERT2020}. However, it has been reported that fine-tuned models can fail to leverage syntactic knowledge \cite{mccoyRightWrongReasons2019, minSyntacticDataAugmentation2020} and rely on pattern matching or annotation artifacts \cite{gururanganAnnotationArtifactsNatural2018, jiaAdversarialExamplesEvaluating2017}.
It is expected that fine-tuned representations will differ significantly from the pre-trained one \cite{rajaeeHowDoesFinetuning2021a} and architectures will deliver different representations of background and linguistic knowledge \cite{durraniHowTransferLearning2021}.

 Probing proved to be an effective method to investigate what information is encoded in the model and how it influences the output \cite{adiFinegrainedAnalysisSentence2017,belinkovProbingClassifiersPromises2021,hupkesVisualisationDiagnosticClassifiers2018}. In recent work, probing was used to verify the model's understanding of scale and magnitude \cite{zhangLanguageEmbeddingsCapture2020} or whether a model can reflect an underlying Foundational Ontology \cite{jullienTransformersEncodeFoundational2022}. In \cite{jinProbingBiomedicalEmbeddings2019} probing was used to determine what additional information is carried intrinsically by BioELMo and BioBERT.

 Recent work on applying language models to biomedical tasks are: MarkerGenie - identifies bioentity relations from texts and tables of publications in PubMed and PubMed Central \cite{guMarkerGenieNLPenabledTextmining2022}; ScispaCy model - relevant for drug discovery, aims to cover disease-gene interactions significant from pharmacological perspective \cite{qumsiyehBiomedicalInformationExtraction2021}; DisKnE - aims to evaluate pre-trained language models about the disease knowledge \cite{alghanmiProbingPreTrainedLanguage2021}; In \cite{vigBERTologyMeetsBiology2021} transformers are used for better understanding working mechanisms in proteins. Biomedical transformers has demonstrated to be highly effective in biomedical NLI task \cite{jinProbingBiomedicalEmbeddings2019}, but safety and validation of their usage is still an under-explored area. A promising direction of future research is to integrate structured knowledge into the models \cite{colon-hernandezCombiningPretrainedLanguage2021, yuanImprovingBiomedicalPretrained2021}.

\section{Conclusions}

In this work we performed a detailed analysis of fundamental knowledge representation properties of transformers, demonstrating that they are biased towards more frequent statements. We recommend to account for this bias in biomedical applications. In terms of the semantic structure of the model, BioMegatron shows more salient biomedical knowledge embedding than BioBERT, as the representations cluster into more interpretable groups and the model better retains the semantic structure after fine-tuning.

We also investigated the representation of entities both in base and fine-tuned models via probing \cite{ferreira-etal-2021-representation}. We found that the fine-tuned models loose the general structure acquired at the pre-training phase and degrade the models with regard to cross-task transferability.

We found biologically relevant clusters, such as genes and variants that are present in the same biological pathways. Considering the vectors used in probing, we found that the distances are associated with entity type (gene, variant, drug, disease). However, the fine-tuning renders the representations internally more inconsistent, which was quantified by the evaluation of clusters' homogeneity. 
We investigated whether the models can capture the quality of evidence and found that they did not perform significantly better for well-known relations. Even for eminent clinical quadruples /statements, the models misclassified the clinical significance (whether sensitive or resistant to treatment), highlighting the limitations of contemporary neural language models.

\section*{Acknowledgements}
This project has received funding from the European Union's Horizon 2020 research and innovation programme under grant agreement No 965397.
This project has also be supported by funding from the digital Experimental Cancer Medicine Team, Cancer Biomarker Centre, Cancer Research UK Manchester Institute P126273.

\printbibliography 
\newpage

\setcounter{secnumdepth}{0}
\section{Supplementary}
\setcounter{secnumdepth}{0}
\label{sec:SuppMat}
\setcounter{table}{0}
\renewcommand\thetable{S.\arabic{table}}
\setcounter{figure}{0}
\renewcommand\thefigure{S.\arabic{figure}}

\subsection{Supplementary Methods}

\subsubsection{Downloading the data}

\label{sec:download_data}
The data was downloaded via CIViC API using following queries:

- 'https://civicdb.org/api/variants/XYZ where XYZ is a 'variant id'

Variant id can be found in the list of all available variants:

- 'https://civicdb.org/api/variants?count=2634'

\subsubsection{Balancing the test set}

\label{sec:balancing_testset}
We excluded the imbalanced pairs/quadruples from the \textit{test set} in order to create a \textit{balanced test set} according to following procedure.

First, we give two definitions of imbalanced entity, followed by the definitions of imbalanced pair and imbalanced quadruple.
We define 2 types of imbalanced entity, true-imbalanced entity and false-imbalanced entity. An entity is considered as true-imbalanced entity if it meets following criteria:

\textbf{Over 70\% of training pairs/quadruples containing this entity are true.}

In reverse, the criteria for false-imbalanced entity is:

\textbf{Less than 30\% of training pairs/quadruples containing this entity are true.}

Based on the definition of true-imbalanced entity and false-imbalanced entity, we can define imbalanced pair as:

\textbf{Either one element of the pair is true-imbalanced entity and the other element is not false-imbalanced entity, or one element of the pair is false-imbalanced entity and the other element is not true-imbalanced entity.}

Similar to the imbalanced pair definition, the imbalanced quadruple can be defined as following:

\textbf{Either one element of the quadruple is true-imbalanced entity and no other element is false-imbalanced entity, or one element of the quadruple is false-imbalanced entity and no other element is true-imbalanced entity.}

Note, for quadruples \textit{true$\mid$false} should be replaced with \textit{sensitivity/response $\mid$ resistance}.

The key intuition of the balancing is to remove the bias that some pairs/quadruples containing specific entities are almost always true (or false). Removing the bias allows us to compare the test results more fairly.

Note, we apply the balancing only to the pairs that are in the test set due to following reasons. First, the training set after balancing would be too small. This is a common drawback when trying to balance the dataset without oversampling, and remains an open challenge for real world datasets. Second, in a ML pipeline the test set should be isolated at the very beginning, before any exploratory data analysis or feature engineering. As the balancing aims for better performance evaluation, we must consider ratios in the test set, but this information should not leak to any activity done on the training set. However, we do exclude pairs (from the test set) also looking at the occurrence in the training set, as we want to mitigate the possible impact of overfitting during training. Balancing the test set left us with 38-53\% of pairs in the balanced test set.

\subsubsection{Transformers}
\label{supp_transformers}

Because both BioBERT and BioMegatron models allow 512 tokens in the input sequences, which is far longer than the input sequences we defined, we do not consider the sentence truncation in this work.

Transformers models have multiple layers, with BioBERT having 12 layers and BioMegatron having 24 layers. One output is generated for each layer, but the output of last layer is generally used as the output of transformers models since we want to fully use the neural network connection architecture through multiple layers.
Multiple vectors are contained in the transformers model's last layer output, where each vector represents one input token in input sequence respectively. A total of 512 vectors are contained in last layer outputs of both BioBERT and BioMegatron because they both allow the same vector size in the input sequences.
Because we do a sentence-level classification task in this work and the first token of each input sequence is ``[CLS]'', we use the vector output of ``[CLS]'' token (first token) in the sequence as pooled output vector of transformers models.
Although there are two major output vector pooling methods, either obtaining the first token vector or averaging the vector of all tokens, we choose to use former, since it is used in most sentence level transformers pre-training tasks such as sentence classification and next sentence prediction.
BioBERT model uses 768-dimensions output vector, while BioMegatron uses 1024-dimensions.

 \begin{equation}
 	\begin{split}
 	V_r=f^{TRF}_\theta(seq)[0]
 	\end{split}
 	\label{eq:transformers_get_emb}
 \end{equation}

As shown in Eq.~\ref{eq:transformers_get_emb}, $f^{TRF}_\theta$ is last-layer output function of transformers model, $seq$ is input sequence of the tranformers model. We use first token's output vector, $f^{TRF}_\theta(seq)[0]$, as pooled output of the sequence, $V_r$.

For training purposes, we stack a classification layer on top of transformers models.
For the Task 1, we need to classify the true and false pairs. We stack a fully connected N-to-1 linear layer and use sigmoid activation to constrain the output value from 0 to 1. Binary cross entropy loss function is used for true/false classification.

For Task 2, we need to classify the multiple clinical significance categories for each input sentence. There are 2 clinical significance categories, ``Sensitivity/Response'' and ``Resistance'' while more categories could be added in further dataset. We use N-to-2 linear layer and softmax activation to get one probability score for each category, then cross entropy loss function is used for model parameter optimization.

\subsubsection{Clustering the probing input}
\label{supp_clustering_probing}
In total, 4,500 and 3,572 vectors were obtained from the pairs and quadruples test set, respectively (see Task 1 and Task 2). Vectors for pairs were aggregated from 3 fine-tuned models trained for each pair type. Each vector consists of 768 for BERT, BioBERT , and 1024 dimensions for BioMegatron. We used UMAP for dimensionality reduction and HDBSCAN clustering algorithm to identify patterns in an 
unsupervised manner.

\subsection{Supplementary Tables}

\setlength\LTleft{0pt}
\setlength\LTright{0pt}
\fontsize{6}{8}\selectfont
\begin{longtable}{@{\extracolsep{\fill}}ccccccc@{}}
\caption{Spearman correlations between the classification error and the number pairs in the training set where an entity occurs. E.g. For BioBERT, there is a significant negative correlation between number of drug-gene pairs that a drug entity occurs in the training set and the classification error. }
\label{tab:correlations_error_nb_pairs}\\
\toprule
Pair type & Model & Entity & True/false pair vs error & Spearman correlation & p-val & Significance \\* \midrule
\endfirsthead
\multicolumn{7}{c}%
{{\bfseries Table \thetable\ continued from previous page}} \\
\toprule
Pair type & Model & Entity & True/false pair vs error & Spearman correlation & p-val & Significance \\* \midrule
\endhead
\cmidrule(l){4-7}
\endfoot
\endlastfoot
\multirow{8}{*}{DRUG - VARIANT} & \multirow{4}{*}{BioBERT} & \multirow{2}{*}{DRUG} & True & -0.75 & 0.0000 & *** \\
 &  &  & False & 0.73 & 0.0000 & *** \\
 &  & \multirow{2}{*}{VARIANT} & True & 0.23 & 0.0010 & * \\
 &  &  & False & 0.06 & 0.3825 & ns \\
 & \multirow{4}{*}{BioMegatron} & \multirow{2}{*}{DRUG} & True & -0.69 & 0.0000 & *** \\
 &  &  & False & 0.68 & 0.0000 & *** \\
 &  & \multirow{2}{*}{VARIANT} & True & 0.15 & 0.0382 & * \\
 &  &  & False & 0.05 & 0.4591 & ns \\
\multirow{8}{*}{DRUG - GENE} & \multirow{4}{*}{BioBERT} & \multirow{2}{*}{DRUG} & True & -0.42 & 0.0000 & *** \\
 &  &  & False & 0.27 & 0.0000 & *** \\
 &  & \multirow{2}{*}{GENE} & True & -0.55 & 0.0000 & *** \\
 &  &  & False & 0.41 & 0.0000 & *** \\
 & \multirow{4}{*}{BioMegatron} & \multirow{2}{*}{DRUG} & True & -0.51 & 0.0000 & *** \\
 &  &  & False & 0.31 & 0.0000 & *** \\
 &  & \multirow{2}{*}{GENE} & True & -0.48 & 0.0000 & *** \\
 &  &  & False & 0.45 & 0.0000 & *** \\
\multirow{8}{*}{VARIANT - GENE} & \multirow{4}{*}{BioBERT} & \multirow{2}{*}{VARIANT} & True & -0.30 & 0.0004 & *** \\
 &  &  & False & 0.05 & 0.5646 & ns \\
 &  & \multirow{2}{*}{GENE} & True & -0.47 & 0.0000 & *** \\
 &  &  & False & 0.61 & 0.0000 & *** \\
 & \multirow{4}{*}{BioMegatron} & \multirow{2}{*}{VARIANT} & True & -0.29 & 0.0007 & *** \\
 &  &  & False & 0.07 & 0.4023 & ns \\
 &  & \multirow{2}{*}{GENE} & True & -0.47 & 0.0000 & *** \\
 &  &  & False & 0.63 & 0.0000 & *** \\* \cmidrule(l){1-7} 
\end{longtable}

\begin{table}
\centering
\caption{Examples of variant entities whose representations appear in the same cluster (5) as gene representations for all 3 base models (BERT, BioBERT and BioMegatron) according to UMAP transformation. Variant representations stem from sentences where the variant entity contains gene name.}
\label{tab:variants_in_umap_cluster}
\resizebox{\textwidth}{!}{%
\begin{tabular}{|l|l|}
\hline
\textbf{variant entry} & \textbf{sentence constructed from quadruple}                                                                                                \\ \hline
IGH-CRLF2 &
  \begin{tabular}[c]{@{}l@{}}IGH-CRLF2 of CRLF2 identified in B-lymphoblastic Leukemia/lymphoma,\\  BCR-ABL1–like is associated with Ruxolitinib\end{tabular} \\ \hline
ZNF198-FGFR1           & \begin{tabular}[c]{@{}l@{}}ZNF198-FGFR1 of FGFR1 identified in Myeloproliferative \\ Neoplasm is associated with Midostaurin\end{tabular}   \\ \hline
SQSTM1-NTRK1           & \begin{tabular}[c]{@{}l@{}}SQSTM1-NTRK1 of NTRK1 identified in Lung Non-small Cell \\ Carcinoma is associated with Entrectinib\end{tabular} \\ \hline
CD74-ROS1 G2032R       & \begin{tabular}[c]{@{}l@{}}CD74-ROS1 G2032R of ROS1 identified in Lung Adenocarcinoma \\ is associated with DS-6501b\end{tabular}           \\ \hline
BRD4-NUTM1             & \begin{tabular}[c]{@{}l@{}}BRD4-NUTM1 of BRD4 identified in NUT Midline Carcinoma\\  is associated with JQ1\end{tabular}                    \\ \hline
KIAA1549-BRAF &
  \begin{tabular}[c]{@{}l@{}}KIAA1549-BRAF of BRAF identified in Childhood Pilocytic \\ Astrocytoma is associated with Trametinib\end{tabular} \\ \hline
TPM3-NTRK1             & \begin{tabular}[c]{@{}l@{}}TPM3-NTRK1 of NTRK1 identified in Spindle Cell Sarcoma \\ is associated with Larotrectinib\end{tabular}          \\ \hline
KIAA1549-BRAF &
  \begin{tabular}[c]{@{}l@{}}KIAA1549-BRAF of BRAF identified in Childhood Pilocytic\\  Astrocytoma is associated with Vemurafenib and Sorafenib\end{tabular} \\ \hline
CD74-NRG1              & \begin{tabular}[c]{@{}l@{}}CD74-NRG1 of NRG1 identified in Mucinous Adenocarcinoma\\  is associated with Afatinib\end{tabular}              \\ \hline
EWSR1-ATF1             & \begin{tabular}[c]{@{}l@{}}EWSR1-ATF1 of EWSR1 identified in Clear Cell Sarcoma\\  is associated with Crizotinib\end{tabular}               \\ \hline
\end{tabular}%
}
\end{table}

\begin{table}
\centering
\caption{AUCs and Brier scores for the balanced test set stratified by evidence level. KNN performs significantly worse for evidence level D compared to the Transformers (bold).}
\label{tab:auc_brier_by_evidence_level}
\begin{tabular}{@{}lllllll@{}}
\toprule
\multicolumn{1}{c}{\multirow{2}{*}{\textbf{Evidence level}}} &
  \multicolumn{3}{c}{\textbf{AUC}} &
  \multicolumn{3}{c}{\textbf{Brier score loss}} \\
\multicolumn{1}{c}{} &
  \multicolumn{1}{c}{\textbf{B}} &
  \multicolumn{1}{c}{\textbf{C}} &
  \multicolumn{1}{c}{\textbf{D}} &
  \multicolumn{1}{c}{\textbf{B}} &
  \multicolumn{1}{c}{\textbf{C}} &
  \multicolumn{1}{c}{\textbf{D}} \\ \midrule
BioBERT     & 0.683 & 0.900 & 0.812 & 0.254 & 0.148 & 0.202 \\
BioMegatron & 0.703 & 0.939 & 0.816 & 0.274 & 0.103 & 0.178 \\
KNN         & 0.682 & 0.910 & \textbf{0.705} & 0.231 & 0.122 & \textbf{0.228} \\ \bottomrule
\end{tabular}
\end{table}

\begin{table}
\centering
\caption{Pairs in cluster 5 in BioBERT representations containing only PIK3CA and ERBB3 genes.}
\label{tab:biobert_cluster5}
\begin{tabular}{@{}lllll@{}}
\toprule
\textbf{Cluster \#5 (brown)} & \textbf{Variant} & \textbf{Gene} & \textbf{TRUE} & \textbf{Predicted probability} \\ \midrule
1  & R93W   & PIK3CA & 1 & 0.678 \\
2  & H1047R & PIK3CA & 1 & 0.664 \\
3  & D350G  & PIK3CA & 1 & 0.666 \\
4  & G1049R & PIK3CA & 1 & 0.624 \\
5  & H1047L & PIK3CA & 1 & 0.673 \\
6  & R103G  & ERBB3  & 1 & 0.773 \\
7  & E545G  & PIK3CA & 1 & 0.657 \\
8  & E281K  & ERBB3  & 0 & 0.756 \\
9  & C475V  & ERBB3  & 0 & 0.780 \\
10 & F386L  & PIK3CA & 0 & 0.680 \\
11 & D816E  & ERBB3  & 0 & 0.776 \\ \bottomrule
\end{tabular}
\end{table}

\begin{table}[]
\centering
\caption{Cluster 2 for BioBERT and cluster 1 for BioMegatron, where the Variant entities contain gene names.}
\label{tab:gene_variant_pairs_cluster21}
\resizebox{\textwidth}{!}{%
\begin{tabular}{@{}llllll@{}}
\toprule
\textbf{\#} & \textbf{Variant} & \textbf{Gene} & \textbf{True/false} & \textbf{cluster \# in BioBERT HAC} & \textbf{cluster \# in BioMegatron HAC} \\ \midrule
1  & D1930V       & ATM    & 1 & 2 & other \\
2  & M2327I       & ATM    & 0 & 2 & other \\
3  & R777FS       & ATM    & 1 & 2 & other\\
4  & ZKSCAN1-BRAF & BRAF   & 1 & 2 & 1 \\
2  & IGH-CRLF2    & CRLF2  & 1 & 2 & 1 \\
6  & DEK-AFF2     & DEK    & 1 & 2 & 1 \\
7  & EWSR1-ATF1   & EWSR1  & 1 & 2 & 1 \\
8  & FGFR2-BICC1  & FGFR2  & 1 & 2 & 1 \\
9  & ATP1B1-NRG1  & NRG1   & 1 & 2 & 1 \\
10 & CD74-NRG1    & NRG1   & 1 & 2 & 1 \\
11 & NRG1         & NRG1   & 1 & 2 & 1 \\
12 & ETV6-NTRK2   & NTRK1  & 0 & 2 & 1 \\
13 & LMNA-NTRK1   & NTRK1  & 1 & 2 & 1 \\
14 & SQSTM1-NTRK1 & NTRK1  & 1 & 2 & 1 \\
12 & ETV6-NTRK2   & NTRK2  & 1 & 2 & 1 \\
16 & NTRK1-TRIM63 & NTRK2  & 0 & 2 & 1 \\
17 & RCSD1-ABL1   & RCSD1  & 1 & 2 & 1 \\
18 & TFG-ROS1     & ROS1   & 1 & 2 & 1 \\
19 & UGT1A1*60    & UGT1A1 & 1 & 2 & 1 \\ \bottomrule
\end{tabular}
}
\end{table}

\begin{table}[]
\centering
\caption{BioBERT quadruples from clusters \#5 and \#6. No obvious patterns. R stands for Resistance and S/R is for Sensitivity/Response.}
\label{tab:biobert_cluster56}
\resizebox{\textwidth}{!}{%
\begin{tabular}{|l|l|l|l|l|l|}
\hline
E17K               & AKT3   & Melanoma                      & Vemurafenib  & R   & 5 \\ \hline
ALK FUSION G1202R  & ALK    & Cancer                        & Alectinib    & R   & 5 \\ \hline
D835H              & FLT3   & Acute Myeloid Leukemia        & Sorafenib    & R   & 5 \\ \hline
G12D               & KRAS   & Colorectal Cancer             & Panitumumab  & R   & 5 \\ \hline
G12R               & KRAS   & Colorectal Cancer             & Panitumumab  & R   & 5 \\ \hline
K117N              & KRAS   & Clear Cell Sarcoma            & Vemurafenib  & R   & 5 \\ \hline
OVEREXPRESSION     & PIK3CA & Melanoma                      & Vemurafenib  & R   & 5 \\ \hline
LOSS               & PTEN   & Melanoma                      & Vemurafenib  & R   & 5 \\ \hline
M237I              & TP53   & Glioblastoma                  & AMGMDS3      & R   & 5 \\ \hline
L3 DOMAIN MUTATION & TP53   & Breast Cancer                 & Tamoxifen    & R   & 5 \\ \hline
T790M       & EGFR & Lung Non-small Cell Carcinoma & Cetuximab and Panitumumab and Brigatinib & S/R & 6 \\ \hline
Y842C              & FLT3   & Acute Myeloid Leukemia        & Lestaurtinib & S/R & 6 \\ \hline
ITD \ D839G & FLT3 & Acute Myeloid Leukemia        & Pexidartinib                             & R   & 6 \\ \hline
ITD I687F          & FLT3   & Acute Myeloid Leukemia        & Sorafenib    & R   & 6 \\ \hline
D839N              & FLT3   & Acute Myeloid Leukemia        & Pexidartinib & R   & 6 \\ \hline
ITD \ Y842C & FLT3 & Acute Myeloid Leukemia        & Sorafenib and Selinexor                  & R   & 6 \\ \hline
G12D               & KRAS   & Melanoma                      & Vemurafenib  & R   & 6 \\ \hline
G12S               & KRAS   & Lung Non-small Cell Carcinoma & Erlotinib    & R   & 6 \\ \hline
G12V               & KRAS   & Colon Cancer                  & Regorafenib  & S/R & 6 \\ \hline
G12V               & KRAS   & Lung Cancer                   & Gefitinib    & R   & 6 \\ \hline
E545G              & PIK3CA & Melanoma                      & Vemurafenib  & R   & 6 \\ \hline
\end{tabular}%
}
\end{table}

\begin{table}[]
\centering
\caption{Homogeneity in clusters obtained from 2 dimensional UMAP representation using HDBSCAN algorithm. }
\label{tab:homogeneity_clusters_umap}
\resizebox{\textwidth}{!}{%
\begin{tabular}{@{}llll@{}}
\toprule
\multicolumn{1}{c}{\textbf{\# cluster}} & \multicolumn{1}{c}{\textbf{BERT}} & \multicolumn{1}{c}{\textbf{BioBERT}} & \multicolumn{1}{c}{\textbf{BioMegatron}} \\ \midrule
1 & 99.7 \% variant & 99.6 \% variant & 100 \% disease \\
2 & 100 \% drug     & 100 \% disease  & 100 \% drug    \\
3 & 100 \% disease  & 99.7 \% variant & 100 \% variant \\
4 & 98.8 \% disease & 99.7 \% drug    & 100 \% disease \\
5                                      & 59.9 \% gene, 40.1 \% variant     & 79.3 \% gene, 20.7 \% variant        & 77.3 \% gene, 20.0\% variant             \\
6 &                 &                 & 100 \% variant \\ \bottomrule
\end{tabular}
}
\end{table}


\pagebreak
\newpage

\subsection{Supplementary Figures}


\begin{figure}[htp!]
\centering
\begin{subfigure}{.50\textwidth}
  \centering
  \includegraphics[width= .9\textwidth]{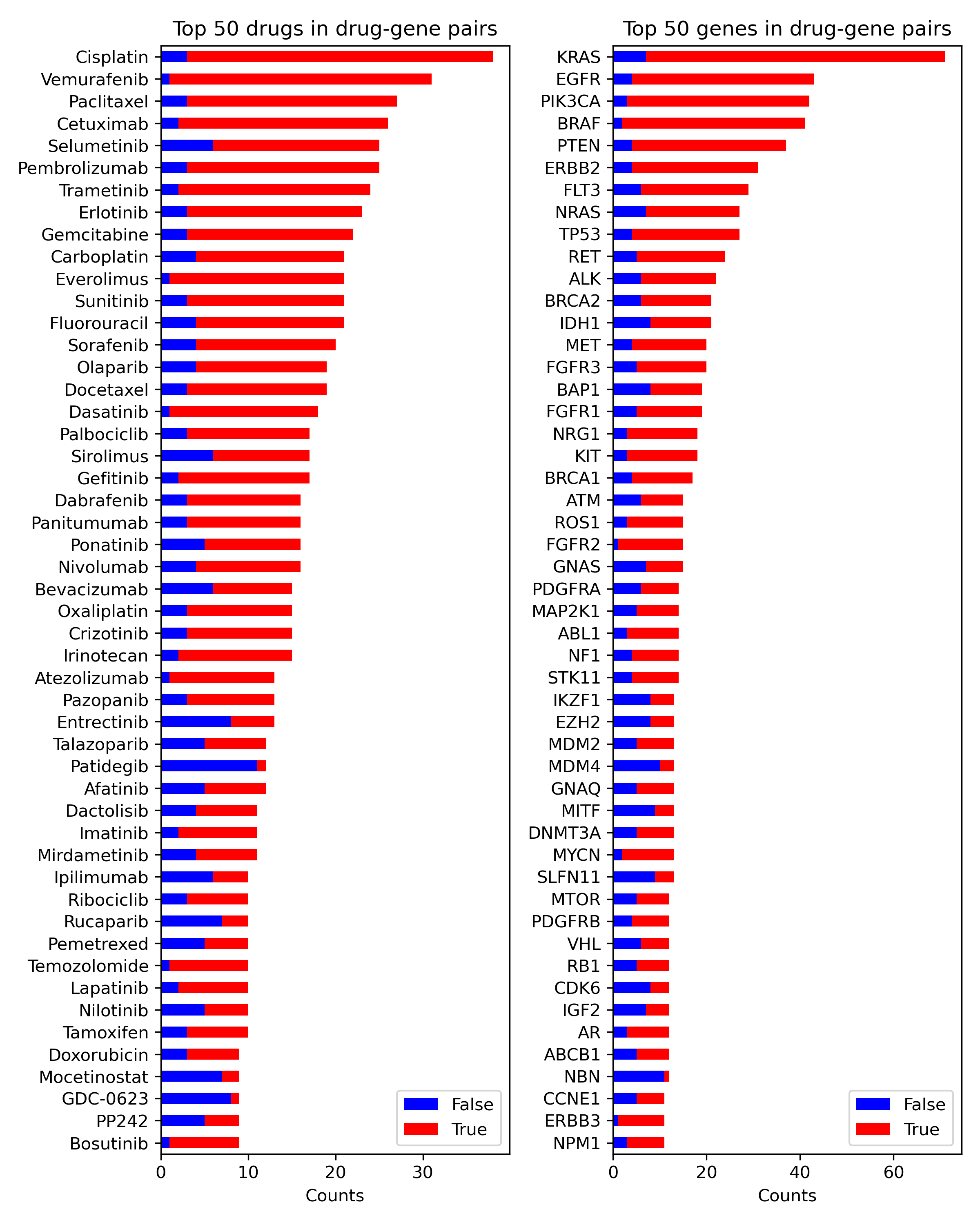}
\caption{}
\label{fig:top_50_drug_gene}
\end{subfigure}%
\begin{subfigure}{.50\textwidth}
  \centering
  \includegraphics[width= .9\textwidth]{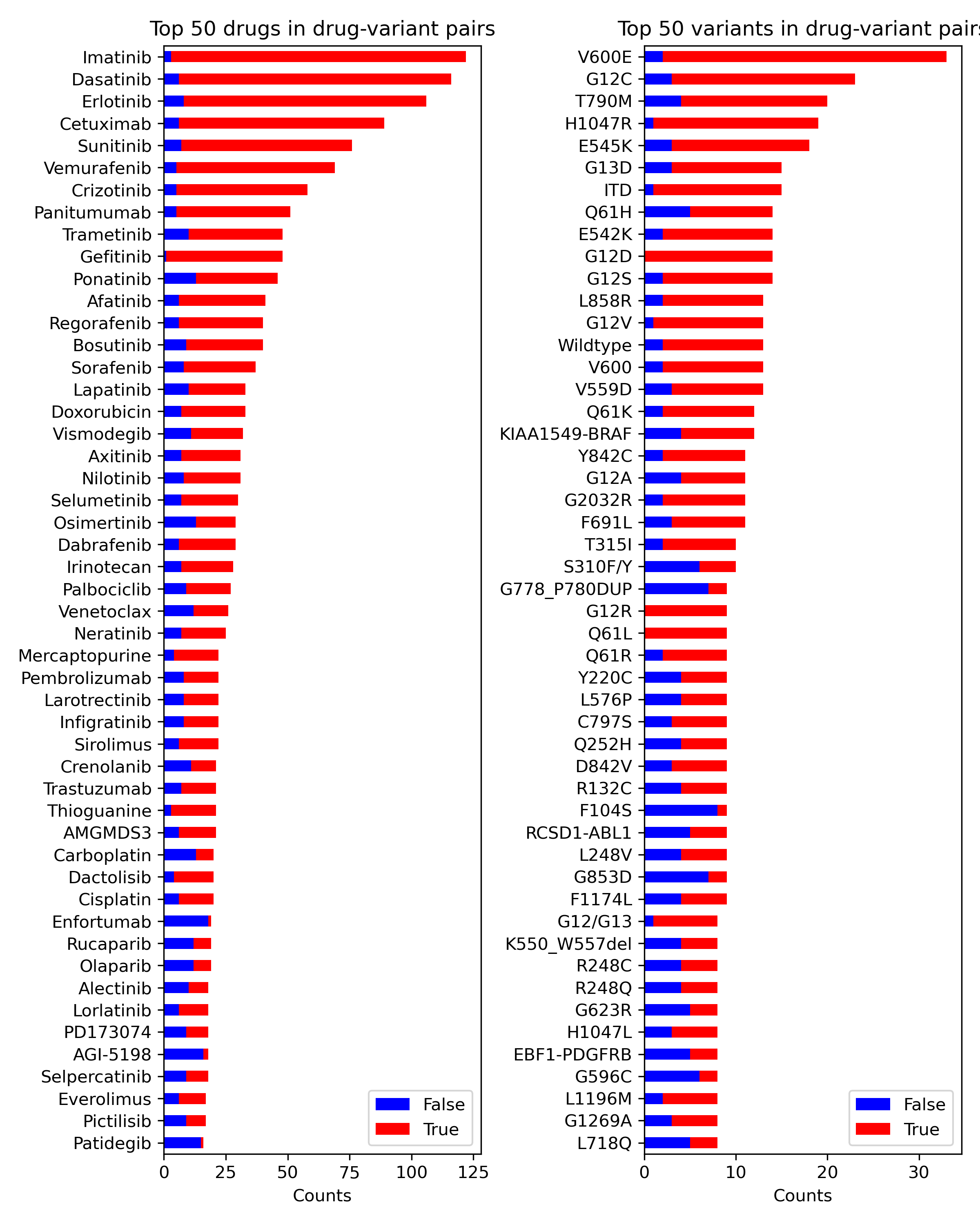}
\caption{}
\label{fig:top_50_drug_variant}
\end{subfigure}%
\\
\begin{subfigure}{.50\textwidth}
  \centering
  \includegraphics[width= .9\textwidth]{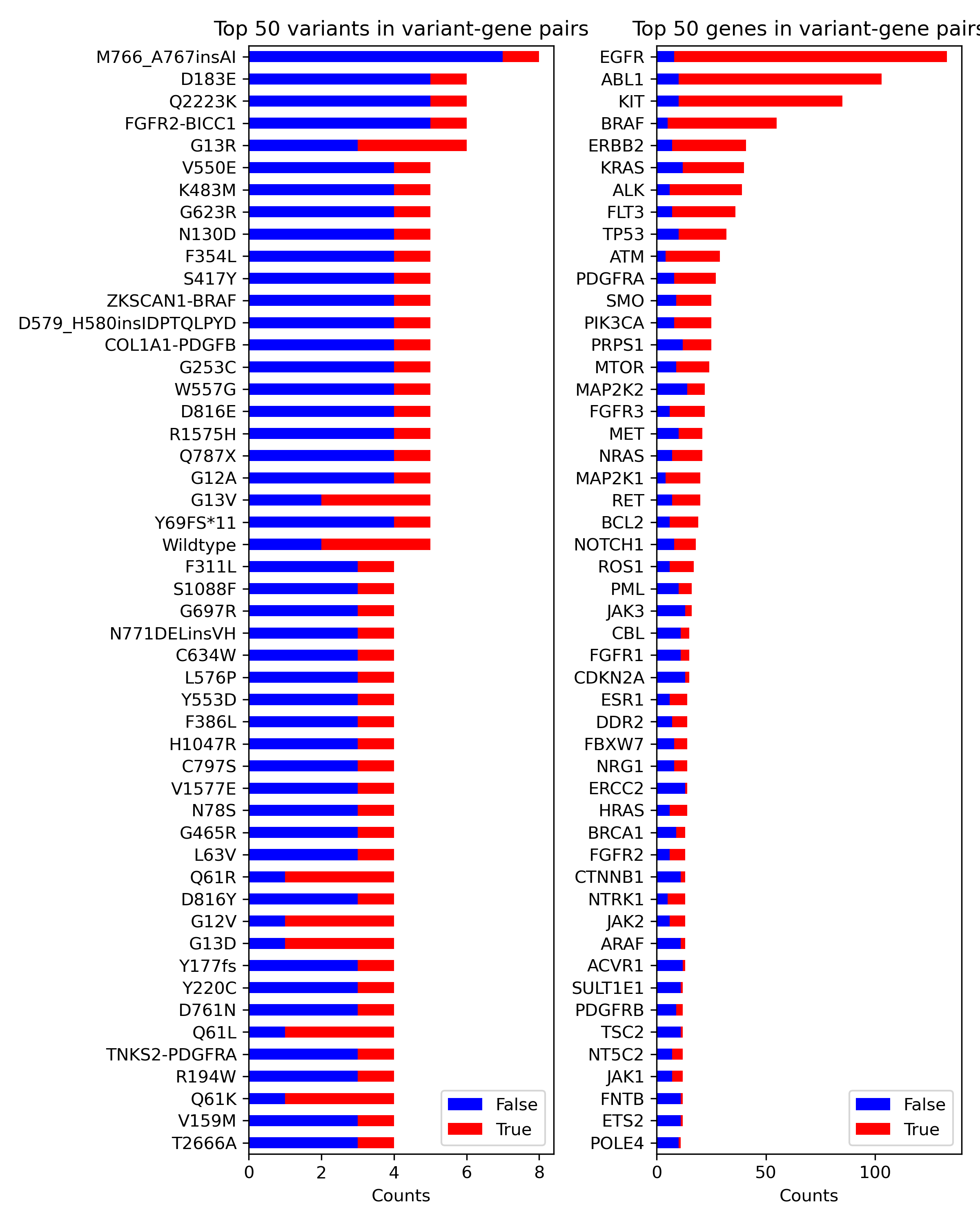}
\caption{}
\label{fig:top_50_variant_gene}
\end{subfigure}%

\caption{Top 50 pairs in the dataset from Task 1. Most frequent entities occur mostly in true pairs, except for variants in variant-gene pairs.}
\label{fig:top_50_pairs}
\end{figure}


\begin{figure}[hp]
\centering
\begin{subfigure}{.50\textwidth}
  \centering
  \includegraphics[width= .9\textwidth]{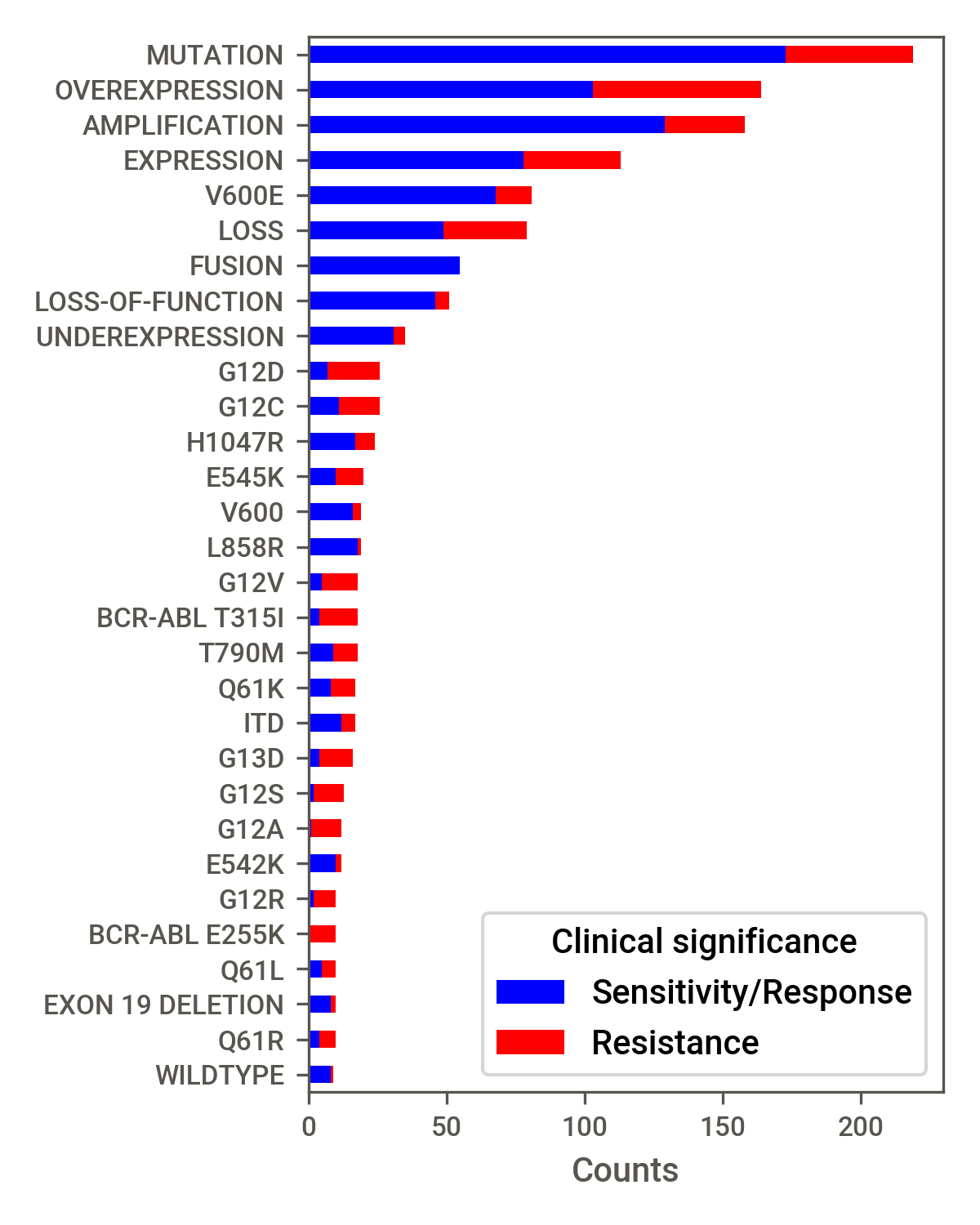}
\caption{}

\end{subfigure}%
\begin{subfigure}{.50\textwidth}
  \centering
  \includegraphics[width= .9\textwidth]{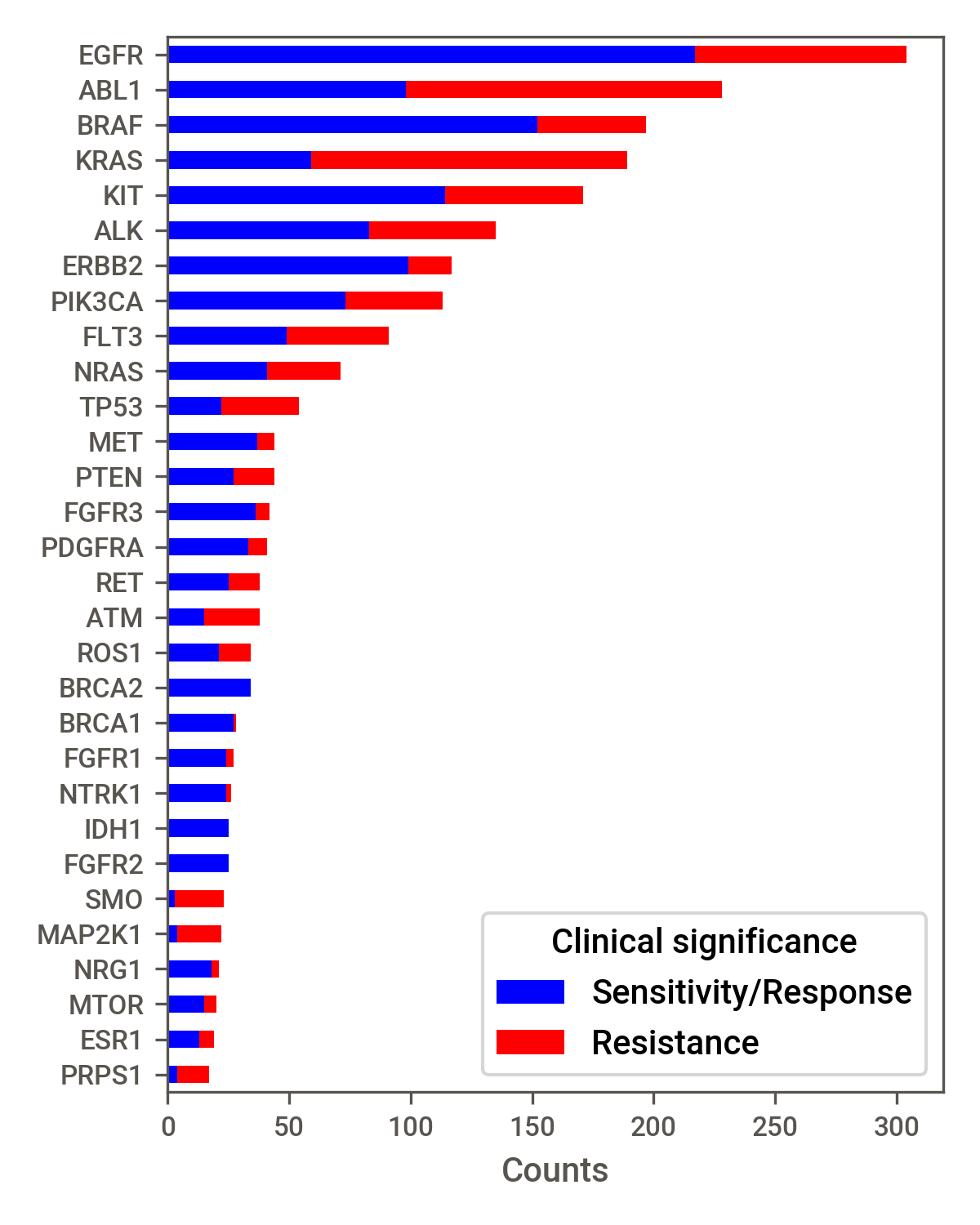}
\caption{}

\end{subfigure}%
\\
\begin{subfigure}{.49\textwidth}
  \centering
  \includegraphics[width= .9\textwidth]{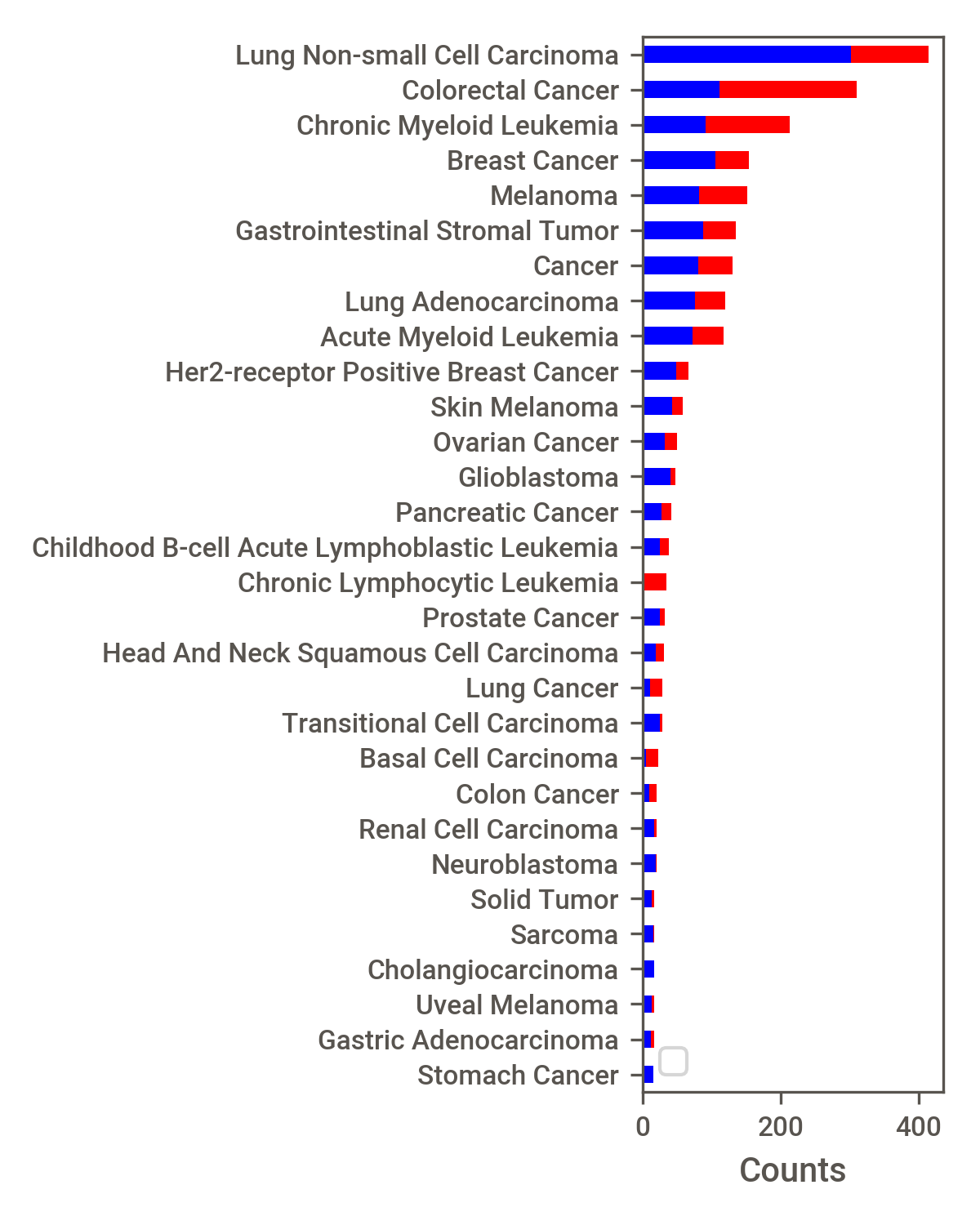}
\caption{}
\end{subfigure}%
\begin{subfigure}{.49\textwidth}
  \centering
  \includegraphics[width= .9\textwidth]{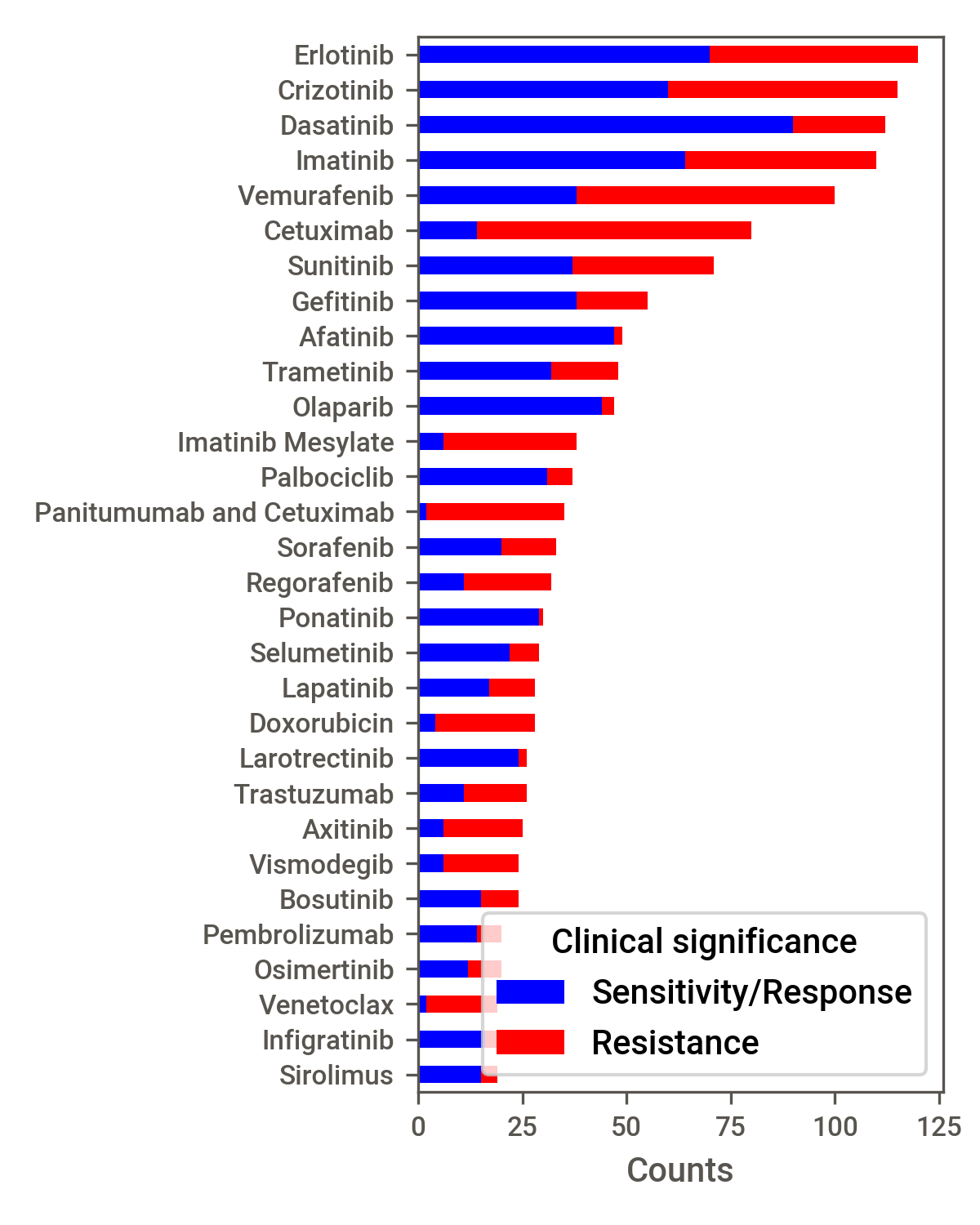}
\caption{}

\end{subfigure}%

\caption{Top 30 entities of each type in the dataset from Task 2: a) variants; b) genes; c) diseases; d) drugs.}
\label{fig:top_30_quads}
\end{figure}

\end{document}